%% file: main.tex
\documentclass[12pt,a4paper, twoside, openright,titlepage]{book}
\usepackage{mathdots}
\usepackage{thesis}
\usepackage{amsmath}
\usepackage{tabularx}
\usepackage{comment}
\usepackage{float} 
\usepackage{hyperref}
\usepackage{rotating}
\usepackage{amssymb}
\usepackage{latexsym}
\usepackage{epsfig}
\usepackage{graphicx}
\usepackage{url}
\usepackage{fancyhdr}

\usepackage{rotating}
\usepackage{algorithm}
\usepackage{algpseudocode}

\algnewcommand\algorithmicinput{\textbf{Input:}}
\algnewcommand\INPUT{\item[\algorithmicinput]}
\algnewcommand\algorithmicoutput{\textbf{Output:}}
\algnewcommand\OUTPUT{\item[\algorithmicoutput]}
\usepackage{array}
\usepackage{ctable}
\usepackage{multirow}
\usepackage{graphicx}
\usepackage{epigraph}
\usepackage{enumerate}
\usepackage{xcolor}
\usepackage{mdframed}
\usepackage{pdflscape}

\definecolor{grey}{gray}{0.6}

\usepackage{setspace}
\begin{document}
    \setcounter{page}{1}
    \pagenumbering{roman}
  \include{HeadTail/coverpage}
  \cleardoublepage
  \setcounter{page}{1}
  \pagenumbering{roman}
\addcontentsline{toc}{chapter}{Title Page}
   \include{HeadTail/title}      \cleardoublepage
   \cfoot{\thepage}
\begin{spacing}{1.1}
   \addcontentsline{toc}{chapter}{Approval of the Viva-Voce board}
   \include{HeadTail/approval}       \cleardoublepage
   
   \addcontentsline{toc}{chapter}{Certificate by the Supervisor}
   \include{HeadTail/cert}       \cleardoublepage
   \include{HeadTail/dedication}    \cleardoublepage
   
   \include{HeadTail/quote}    \cleardoublepage   

    \addcontentsline{toc}{chapter}{Declaration}
    \include{HeadTail/declaration}     \cleardoublepage
 \addcontentsline{toc}{chapter}{Acknowledgement}
 \include{HeadTail/ack}            \cleardoublepage
\addcontentsline{toc}{chapter}{Abstract}
\include{HeadTail/Abstract}
\cleardoublepage
\tableofcontents
\cleardoublepage
   \addcontentsline{toc}{chapter}{List of Abbreviations} 
   \markboth{List of Abbreviations}{List of Abbreviations}  
   \include{Symbol/abbreviation_v2} 
       \clearpage{\pagestyle{empty}\cleardoublepage} 



%


\thispagestyle{empty}
 
\listoffigures
 
\listoftables
 \cleardoublepage
\newpage

    \setcounter{page}{1}
    \pagenumbering{arabic}

\include{Chapter1/chapter_intro_v12}

\include{Chapter3/chapter_iet_v11}

\include{Chapter4/chapter_gazeir_v9}
\include{Chapter5/chapter_eac_v6}
\include{Chapter6/chapter_prl_v7}

\include{Chapter7/chapter_gazeactivity_v8}

\include{Conclusion/conclusion_v10}

\end{spacing}

%

\begin{appendix}
\include{Appendix/AppendixI_v3}

\clearpage{\pagestyle{empty}\cleardoublepage} 
\end{appendix}
    \cleardoublepage
    \lhead{} \rhead{}

\clearpage
\pagestyle{plain}
    \addcontentsline{toc}{chapter}{References}
    \markboth{References}{References}
\bibliographystyle{myIEEEtran}
\bibliography{reference_v7}

  \include{HeadTail/publication_v2}
  \begin{spacing}{1.25}
  \end{spacing}

\clearpage{\pagestyle{empty}\cleardoublepage}
  \include{HeadTail/biodata_v6}

    \begin{spacing}{1.25}
  \end{spacing}
\end{document}

%% file: HeadTail/coverpage.tex
\thispagestyle{plain}
\begin{center}
\vspace*{1cm} \begin{spacing} {1.25}
{\LARGE \bf IMAGE BASED EYE GAZE TRACKING AND ITS APPLICATIONS\\}
\end{spacing}
\end{center}

\vspace{12cm}
\begin{flushright}
\begin{Large}
{\bf {\em  Anjith George ~~}\noindent \vskip
-.8\baselineskip \noindent \rule[-2.5 mm]{\textwidth}{3 pt}\\}
\end{Large}
\end{flushright}

%% file: HeadTail/title.tex
\thispagestyle{plain}
\begin{center}
\begin{spacing} {1}
{\Large \bf IMAGE BASED EYE GAZE TRACKING AND ITS APPLICATIONS\\}
\end{spacing}

\vspace*{1.0cm} {\em Thesis submitted to\\ Indian Institute of
Technology, Kharagpur \\ for the award of the degree}\\
\vspace*{0.4cm} {\em of} \\
\vspace*{0.4cm}
{\Large {\bf Doctor of Philosophy}}\\

\vspace*{0.4cm}
{\em by}\\
\vspace*{0.4cm}
{\Large{\bf Anjith George}}\\

\vspace*{2cm}
{\em under the guidance of }\\
\vspace*{0.3cm}
{\bf Prof. Aurobinda Routray}\\
{(Electrical Engineering Department)}

\vspace*{1.6cm}

\begin{figure}[htbp]
\centerline{
\includegraphics[scale=.3]{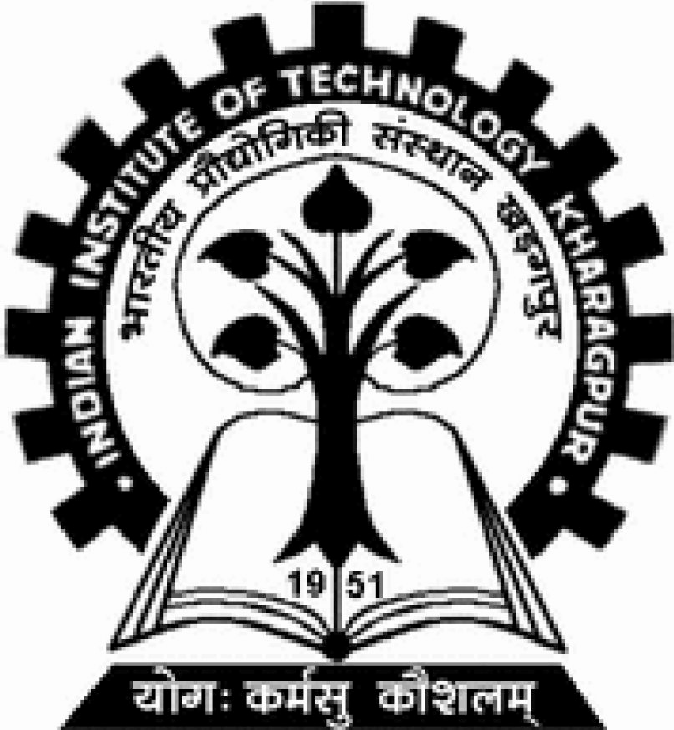}}
\end{figure}
{{\bf ELECTRICAL ENGINEERING DEPARTMENT\\
INDIAN INSTITUTE OF TECHNOLOGY KHARAGPUR\\
AUGUST 2017}} \\
\vspace*{0.25cm} \copyright 2017, Anjith George. All rights
reserved.
\end{center}

%% file: HeadTail/approval.tex
\thispagestyle{empty}
\vspace*{1cm}
\vskip 2.0\baselineskip \centerline{\underline{\large{\bf
APPROVAL OF THE VIVA-VOCE BOARD}}} 
\vspace*{1cm}
\vskip 1.0\baselineskip \noindent 
\begin{flushright}
---/---/2017
\end{flushright}
\vspace*{1cm}
Certified that the thesis entitled IMAGE BASED EYE GAZE TRACKING AND ITS APPLICATIONS submitted by ANJITH GEORGE to the Indian Institute of Technology, Kharagpur, for the award of the degree Doctor of Philosophy has been accepted by the external examiners and that the student has successfully defended the thesis in the viva-voce examination held today.\\

\vspace*{2cm}
\newcommand\textbox[2]{%
  \parbox{.333\textwidth}{#1}%
}
\noindent \textbox{(Member of the DSC)\hfill}\textbox{\hfil (Member of the DSC) \hfil}\textbox{\hfill (Member of the DSC)}\\

\vspace*{2cm}
\noindent \textbox{\hfil \textcolor{white}{(Member of the DSC)}\hfil}\textbox{\hfill (Supervisor)}\\

\vspace*{2cm}
\noindent \textbox{(External Examiner)\hfill}\textbox{\hfil \textcolor{white}{(Member of the DSC)}\hfil}\textbox{\hfill (Chairman)}

%
%
%
%
%
%

%% file: HeadTail/cert.tex
\thispagestyle{empty}

\begin{figure}[htbp]
\includegraphics[scale=.3]{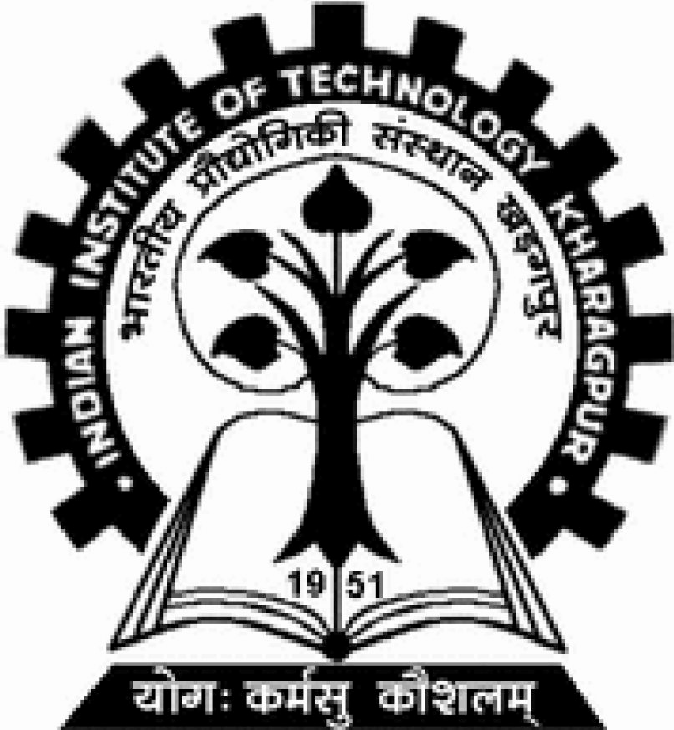}
\end{figure}

\vspace*{-4.0\baselineskip}

\hspace*{2cm} {\large{\bf{\sf Electrical Engineering Department}}} \\
\hspace*{2.6cm}   {\large{\bf{\sf Indian Institute of Technology, Kharagpur}}}\\
\hspace*{2.6cm}   {\large{\sf Kharagpur, India-721302}}\\
\vspace*{0.5cm}
\rule{\textwidth}{0.75pt}

\vskip 2.0\baselineskip \centerline{\underline{\Large{\bf
Certificate}}} \vskip 2.0\baselineskip \noindent This is to
certify that this thesis entitled {\bf Image based eye gaze tracking and its applications} submitted by {\bf {Anjith George}}, to the Indian
Institute of Technology, Kharagpur, is a record of bona fide research work carried out under my supervision and is worthy of consideration for award of Ph.D of the Institute.
\vspace*{1cm}
\begin{flushright}
\raisebox{0.4cm}[4cm][0cm]{
\begin{minipage}{7.8cm}
{\bf Prof. Aurobinda Routray}\\
Electrical Engineering Department \\
Indian Institute of Technology, Kharagpur \\
 {India}- 721302.\\
\end{minipage}
}
\end{flushright}
\begin{flushleft}

I.I.T. Kharagpur\\
August, 2017\\
\end{flushleft}

%% file: HeadTail/dedication.tex
\thispagestyle{empty}%
\vspace*{0.25\textheight}
\begin{center}
{\AlgoFont{14} {\it - To my parents -}}\\
\vspace*{1.0\baselineskip}
{\AlgoFont{14} {\it            - To my loving wife Neethu -}}\\
\vspace*{0.5\baselineskip}
\end{center}

%% file: HeadTail/quote.tex
\thispagestyle{empty}%
\vspace*{0.25\textheight}
\begin{center}
{\AlgoFont{14} {\it ``Research is formalized curiosity. It is poking and prying with a purpose.'' }}\\
\vspace*{1.0\baselineskip}
{\AlgoFont{14} {\it            - Zora Neale Hurston }}\\
\vspace*{0.5\baselineskip}
\end{center}

%% file: HeadTail/declaration.tex
\thispagestyle{empty}


\vskip 2.0\baselineskip \centerline{\underline{\Large{\bf
Declaration}}} \vskip 2.0\baselineskip \noindent I certify that

\begin{itemize}

\item [a.] the work contained in the thesis is original and has
been done by me under the guidance of my supervisor;

\item [b.] the work has not been submitted to any other institute
for any other degree or diploma;

\item [c.] I have followed the guidelines provided by the Institute
in preparing the thesis;

\item [d.] I have conformed to ethical norms and guidelines while
writing the thesis;

\item [e.] whenever I have used materials (data, models, figures and text) from other sources, I have given due credit to them by citing them in the text
of the thesis, and giving their details in the references, and taken
permission from the copyright owners of the sources, whenever
necessary.

\end{itemize}

\vspace{2cm} Anjith George

%% file: HeadTail/ack.tex
\chapter*{Acknowledgment}

I am grateful to my supervisor Prof. Aurobinda Routray for his constant support and encouragement throughout the research work.  This thesis would not have taken its current shape without his constant motivation and support. I am thankful to him for his insightful comments and willingness to help whenever needed.

I would like to thank the members of the Doctoral Scrutiny Committee, Prof. Pallab Dasgupta, Prof. P. K. Dutta, and Prof. A.K. Deb for their valuable suggestions and interactions throughout the tenure. I am thankful to Head, Department of Electrical Engineering, IIT Kharagpur, for providing the facilities to carry out my research work.  I am also
grateful to all other faculty members of the department for the
help received during my research work.

I am grateful to Prof. P. Patnaik, Department of Humanities and Social Science, IIT Kharagpur for the help with experiment designs and valuable discussions.  I also appreciate the support from Prof. Rajlakshmi Guha for the experiment design and analysis. I am also grateful to Department of Information Technology, Govt. of India and Samsung Electronics for funding various phases of the project.

I would like to acknowledge the help and support of all my colleagues and labmates from NEX and RTES Lab, Department of Electrical Engineering. Special thanks to Anand Sahadev, Anirban Dasgupta, SL Happy and Ramnarayan Mohanty for the support and insightful discussions.

Last but not the least, I would like to thank my parents for their support throughout these years. I would like to thank my wife Neethu for her wholehearted support and encouragements through the difficult times.  This thesis would not have been possible without her sacrifice, love and support.

\vspace{0.25in}
\begin{flushright}
{\em Anjith George}
\end{flushright}

%% file: HeadTail/Abstract.tex
\chapter*{Abstract}
\indent Eye movements play a vital role in perceiving the world. Eye gaze can give a direct indication of the user’s point of attention, which can be useful in improving human-computer interaction. Gaze estimation in a non-intrusive manner can make human-computer interaction more natural. Eye tracking can be used for several applications such as fatigue detection, biometric authentication, disease diagnosis, activity recognition, alertness level estimation, gaze contingent display, human-computer interaction, etc. Even though eye tracking technology has been around for many decades, it has not found much use in consumer applications. The main reasons are the high cost of eye tracking hardware and lack of consumer level applications. In this work, we attempt to address these two issues. In the first part of this work, image-based algorithms are developed for gaze tracking which includes a new two-stage iris center localization algorithm. The iris center location along with eye corners are used to develop a gaze tracking framework which can operate even with off the shelf webcams. We have further developed an algorithm for head-mounted eye trackers. Most of the available algorithms perform well only in controlled environments. In order to make eye tracking ubiquitous, they should work in outdoor conditions as well. To this end, we have developed a new algorithm which works in challenging conditions such as motion blur, glint and varying illumination levels. A person independent gaze direction classification framework is also developed which eliminates the requirement of user specific calibration.  A convolutional neural network based classifier is proposed for real-time gaze direction classification.

In the second part of this work, we have developed two applications which can benefit from eye tracking data. A new framework for biometric identification based on eye movement parameters is developed. A score fusion methodology with a new set of features extracted from fixations and saccades is adopted for biometric identification. The addition of eye movement features along with the conventional iris recognition systems can lead to a counterfeit-proof biometric modality with inbuilt liveliness detection capability. A framework for activity recognition, using gaze data from a head mounted eye tracker is also developed. The information from gaze data, ego-motion, and visual features are integrated to classify the activities. This approach improves the classification accuracy in indoor conditions where conventional activity detection modalities fail.

\textbf{Keywords}:  Eye detection, Eye tracking, Gaze tracking, Eye movement biometrics, Egocentric activity recognition.

%% file: Symbol/abbreviation_v2.tex
\chapter*{List of Abbreviations}\label{LstAbb}
\begin{flushleft}
\begin{longtable}{lp{9.3cm}}
AAM	& Active Appearance Model\\
AR	& Augmented Reality\\
BoW	& Bag of Words\\
BP-DP & Bright Pupil Dark Pupil method\\
CHT	& Circular Hough Transform\\
CLM	& Constrained Local Model\\
CMC	& Cumulative Match Characteristics\\
CNN	& Convolutional Neural Network\\
CPU	& Central Processing Unit\\
DCT	& Discrete Cosine Transform\\
DET	& Detection Error Trade-off\\
EAC	& Eye Accessing Cues\\
EC-IC & Eye Corner-Iris Centre\\
EER	& Equal Error Rate\\
EGT & Eye Gaze Tracking\\
EOG	& Electro Oculogram\\
ERT	& Ensemble of Randomized Tree\\
FFT	& Fast Fourier Transform\\
GOF & Goodness of Fit\\
GPF	& Generalized Projection Function\\
GPU	& Graphical Processing Unit\\
HCI	& Human Computer Interaction\\
HOG	& Histogram of Oriented Gradients\\
I-VT	 & Velocity Threshold\\
IC	& Iris Center\\
IR	& Infrared\\
KF	& Kalman Filter\\
KNN	& K Nearest Neighbours\\
LBG	& Linde-Buzo-Gray Algorithm\\
LBP	& Local Binary Pattern\\
MH & Motion Histogram\\
MSER	 & Maximally Stable Extremal Regions\\
NCC	& Normalized Cross Correlation\\
NIR	& Near Infrared\\
NLP	& Neuro-Linguistic Programming\\
OPMM	 & Oculomotor Plant Mathematical Model\\
OS	& Operating System\\
PC	& Pupil Centre\\
PCA	& Principal Component Analysis\\
PoG	& Point of Gaze\\
PoR	& Point of Regard\\
RAM	& Random Access Memory\\
RAN	& Random dot stimulus\\
RANSAC & Random Sample Consensus\\
RBF	& Radial Basis Function\\
RBFN	 & Radial Basis Function Network\\
ReLU	 & Rectifier Linear Unit\\
RF	& Random Forest\\
ROI	& Region of Interest\\
SDSE	 & Synchronized Delaunay Sub-manifold Embedding\\
SGD	& Stochastic Gradient Descent\\
SIFT	 & Scale Invariant Feature Transform\\
SVM	& Support Vector Machine\\
TEX	& Text stimulus\\
VOG	& Video Oculography\\
VR	& Virtual Reality\\
WEC	& Worst Eye Characteristics\\

\end{longtable}
\end{flushleft} 

%% file: Chapter1/chapter_intro_v12.tex
\chapter{Introduction}{Introduction}
\graphicspath{{Chapter1/pics/}}
\begin{onehalfspacing}

\section{Introduction}
Human eyes are powerful means of nonverbal communication. They are good indicators of a person’s attention and interest. Eye movements are a natural part of our interaction with the world. Gaze direction can give a direct indication of the user’s point of focus. Eye movement involves an attention shift and tracking one’s gaze can reveal a lot about his actions. This can be vital in improving human-computer interaction (HCI), developing intelligent interfaces where machines can identify and interact with humans in a more natural way \cite{george2015design}. Seamless and personalized interaction is possible when the machine can recognize the identity, interest, intentions, context, and actions of the user.

In this context, non-intrusive estimation of gaze location can be useful in many practical applications. Eye gaze tracking (EGT) refers to the process of estimating the point where the user is looking, and the instrument used for this is known as eye tracker \cite{duchowski2007eye}.

\section{Motivation }

Even though eye tracking technology has been around for many decades, it has not found widespread adoption at the consumer level. Several challenges need to be addressed to make eye tracking a ubiquitous tool. The high cost of commercially available eye trackers is one of the prime factors limiting its utility. Reduced accuracy in real world scenarios is another factor which limits the applicability. Lack of consumer level use cases is another issue.  

In this work, an attempt is made to address the above mentioned issues. The contents of this thesis can be divided into two parts. The first part focuses on improving the image based gaze tracking systems. To attain this, we have developed applications for two different settings, specifically for the desktop environments and outdoor conditions. We have developed algorithms for gaze tracking in real world conditions, keeping low cost in mind. Most of the existing eye trackers require special hardware cameras and illumination systems making the system costlier. However, low accuracy eye tracking systems can be developed using off the shelf webcams with no additional hardware. It is worth noting that the accuracy requirement for each application is different. A comparatively low accuracy eye tracker would suffice for localizing the approximate region of gaze for gaming applications. High temporal, as well as spatial resolution, may be required for applications like eye movement based biometric authentication. The cost of eye tracking should be minimized to make the technology ubiquitous. To this end, we propose to develop a webcam-based eye tracker which can be implemented using the software without the requirement of any additional hardware. Further, pervasive eye tracking is possible when eye tracking systems are wearable and robust against real world conditions. We have developed robust algorithms for head mounted eye trackers to tackle this issue.

In the second part, two applications of eye tracking data are developed, namely biometric authentication and activity recognition. Eye movements exhibit signature patterns with a possible use case in biometric authentication. Eye movements are generated by a complex oculomotor plant which is very hard to spoof by mechanical replicas.Hence, use of eye movement dynamics along with iris recognition technology could lead to a robust counterfeit-resistant person identification system. Information obtained from eye movements might be useful in defining the user context which can be useful for designing cognitive-aware interfaces.  Patterns in eye movements can also be useful in classifying the activities of the user.

A brief introduction to eye gaze tracking along with a detailed review of current applications and state of the art in eye tracking technology are included below. 

\section{Anatomy of eye}

\subsection{External anatomy}

\begin{figure}[!htb]
\centering
\includegraphics[width=0.85\linewidth]{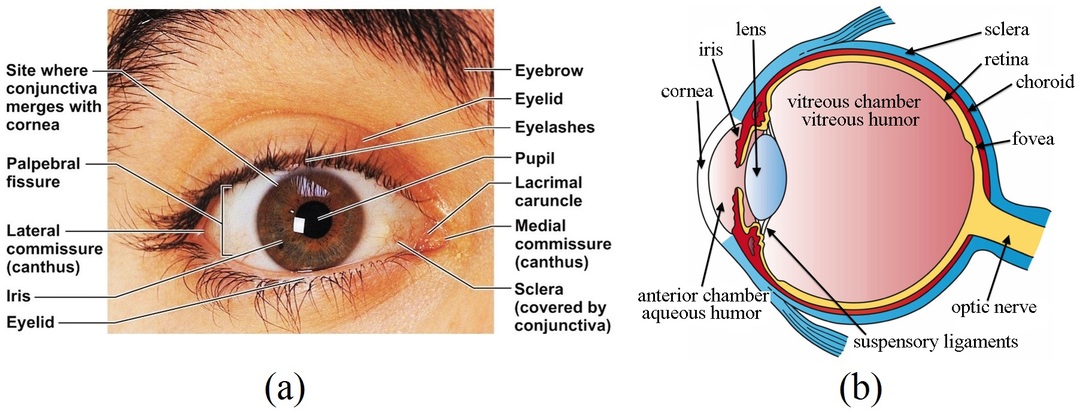}
\caption{External anatomy of eye, a) Frontal view, b)Side view.}
\label{fig:anatomy}
\end{figure}

Figure \ref{fig:anatomy} shows the external anatomy of the eye. The human eye contains several parts which help in the formation of the image in the fovea.  The focusing is primarily done by the cornea, which is the front surface of the eye.  Iris regulates the amount of light reaching the back of the eye. Iris acts like a diaphragm adjusting the size of the pupil to control the amount of light.  Focusing is done by the lens located behind pupil through a process known as accommodation. This helps in forming a clear image even if the object in focus is near or far.  The focused light fall on the fovea and the photoreceptors in fovea convert light into an electrical signal which is then transmitted to visual cortex via optical nerve. There are two types of photoreceptors namely, rods and cones. The greatest concentration of rods and cones is in an area of the retina called the fovea. The center of the fovea known as foveola is entirely composed of cones.

\subsection{Imaging of eye region}
\begin{figure}[!htb]
\centering
\includegraphics[width=0.85\linewidth]{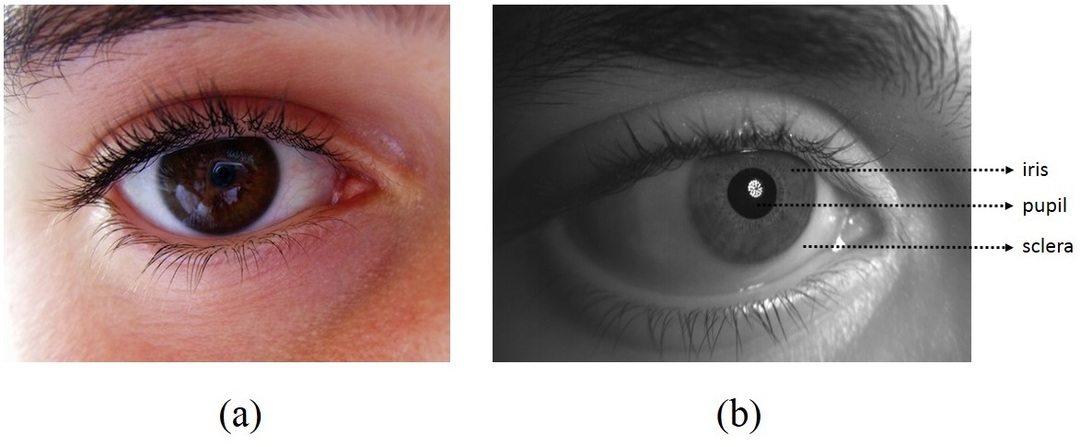}
\caption{Eye image captured under, a) Visible light, b) NIR lighting.}
\label{fig:visible_nir}
\end{figure}

Figure \ref{fig:visible_nir} \cite{happy2012video} shows the eye images captured in visible as well as near infrared (NIR) lighting conditions. In visible images, the boundary between iris and sclera is more prominent than the pupil iris boundary (also known as limbus boundary). Most of the visible spectrum eye trackers make use of these edges for gaze tracking. In contrast, in NIR images, pupil iris boundary is much more prominent. Most of the commercial eye trackers leverage pupil boundary (using dark pupil method) and glints on the eyes to estimate the gaze position.  It is worth noting that accuracy in estimating the pupil iris boundary in NIR images is much more than determining the sclera iris boundary in visible spectrum images. However, visible spectrum imaging has the advantage that it does not require specific hardware or lighting arrangements and most of the smart devices are readily having a front facing camera.

 \section{Methods for measuring eye movements}

 In literature, primarily four methods are used for eye tracking: scleral coil, Electro-Oculography (EOG), Photo-Oculography (POG) or Video-Oculography (VOG), and image-based Corneal/ Pupil reflection based method \cite{duchowski2007eye}.

 In scleral coil method, a contact lens with a mechanical or optical reference object is worn directly on the eye. A coil is attached to the contact lens, and the position of eyes can be found from the electric potential induced when placed in a magnetic field. The scleral coil method is the most accurate method for estimating the eye position. However, the high level of discomfort due to its invasive nature prevents its use in practical applications. Moreover, this method determines the position of the eye with respect to the head.  Gaze estimation requires the eye position as well as the head pose. Lack of head pose information limits its use for point gaze estimation \cite{young1975survey}.

 EOG method uses the electric potential differences of skin measured through the electrodes placed on regions around the eye. The eye movements recorded through EOG do not give any information about the head pose. Hence, this modality is not suitable for point of regard estimation unless the head pose is also available (which requires another head tracker) \cite{westheimer1954mechanism}.

 POG method tries to measure the features of the eye such as the shape of pupil, limbus sclera boundary, and corneal reflections from the light source. The variations in the appearance with eye and head movements are used to estimate the point of regard \cite{duchowski2007eye}.

 Video or image based trackers use cameras and image processing algorithms to find the gaze point in real-time. There are different types of video-based eye trackers such as head mount, table mount and tower mount trackers. 

Literature suggests that image-based eye tracking methods are suitable for practical applications due to their non-intrusive and non-contact nature. The steady increase in computational power and camera quality improve the performance of image-based eye trackers.  A brief description of different types of image-based eye trackers is provided in the next section.

 \section{Image based eye gaze tracking}

 There are mainly two types of image-based eye trackers based on the illumination source, 1) Eye trackers which use active Infrared illumination and, 2) eye trackers using visible spectrum illumination.

 Active IR illumination methods utilize bright pupil and dark pupil effects (BP-DP) \cite{morimoto2005eye}. These methods are efficient and accurate methods for detection of pupil center at low frame rates and controlled conditions. The method works on a differential infrared scheme. In this approach \cite{liu2002real}, Infrared sources of two frequencies are used. The illumination sources are synchronized with the image capturing system. The first image is captured with infrared lighting at 850 nm which produces a distinct glow in the pupils (the red-eye effect). The second image uses a 950 nm infrared source for illumination that results in an image with dark pupils. These two images differ only by the brightness of pupil region. Now the difference between the two images is found in which the pupil region will be highlighted. After post-processing, the pupil blobs are identified and used for computing the pupil center \cite{guang2008real}. The main disadvantage of this method is that the detection rate changes with several factors such as brightness changes, the size of pupils, face orientation, and external light interference. The intensity of external light should be limited. The reflection and glints from spectacles pose another problem. Recently many developments were made in tuning the irradiation of IR illuminators. IR illuminators have to be tuned in order to operate in different natural light conditions, multiple reflections of glasses, and variable gaze directions. Some researchers tried to implement systems which combine the active IR methods with appearance-based methods.  These combined models can robustly track eyes even when the pupils are not very bright due to significant external illumination interferences \cite{liu2002real}. However, active infrared based systems require special cameras and lighting arrangements, which makes them costly.  Most of the commercially available eye trackers use active illumination with different methods such as BP-DP, dark pupil, and bright pupil method.

 Recently some researchers have proposed methods \cite{ferhat2016low} to use the standard off the shelf webcams for gaze estimation without the need for any additional hardware. The accuracy of visible spectrum trackers is less than that of the IR-based trackers \cite{ferhat2016low}. However, with the increase in camera quality available in smart devices like mobiles and tablets, the accuracy of gaze estimation could increase. Developing algorithms which can work in standard cameras could enhance the adoption of the technology.

\begin{figure}[!htb]
\centering
\includegraphics[width=0.8\linewidth]{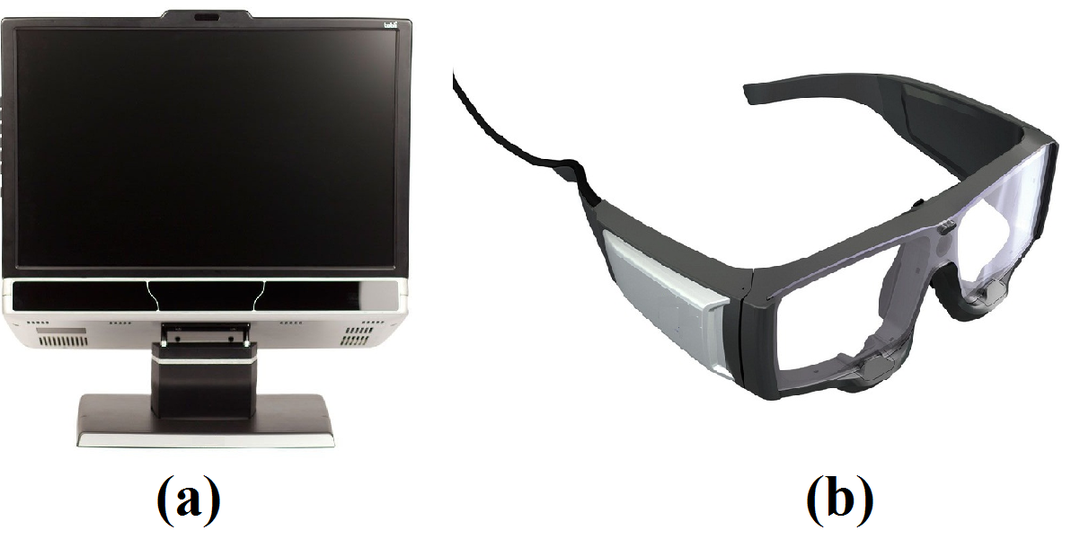}
\caption{Different types of eye trackers, a) Remote eye tracker, b) Head-mounted eye tracker.}
\label{fig:remote_head}
\end{figure}

 There are different configurations of eye trackers as well. The type of eye tracker to be used varies for different applications. They can be broadly categorized into two, 1) Remote eye trackers and 2) head-mounted eye trackers. Figure \ref{fig:remote_head} shows example images of these two types.

 \subsection{Remote eye trackers}

 Remote eye trackers are placed at a distance from the user, usually in controlled desktop environments. They can be monocular or binocular. Most of the eye trackers use bright pupil and dark pupil effects along with the corneal glints. The reflections and the deformations can be used to make eye tracking head pose invariant. The main disadvantage of such trackers is that they are limited to desktop environments.


 \subsection{Head mounted eye trackers}

 Head mounted trackers usually contain two cameras; one camera focuses on the user's eye to find the pupil center and the other camera is looking outwards capturing the user's field of view. Head mounted trackers are more useful for collecting eye-tracking data in natural interaction environments. Typically, the eye tracking data as well as the video from the camera are stored in the memory associated with the eye tracker for offline analysis. Recently, with the emergence of virtual reality (VR) headsets, there has been attempts to use eye tracking data in real-time as an interaction channel.


 \section{Taxonomy of eye movements}

 Eye movements help in orienting the fovea towards the area of interest \cite{duchowski2007eye}. This is achieved by several types of eye movements such as saccades, smooth pursuit, vergence, vestibular, and nystagmus. Eye fixations also contain small movements of eyes. A brief description of eye movement types is given below.

 \subsection{Fixations}
 Fixation refers to maintaining of the visual gaze on a single location. Fixations help in gathering visual information. The typical duration of a fixation lies in the range of 200-300 ms. A fixation contains several other miniature eye movements like tremor, drift, and microsaccades.

 \subsection{Saccades}

 The brain perceives the visual field when an area is focused in fovea during fixations. Changing the focus of fixation from one region to another is achieved by a rapid movement of the eyeball known as saccades. Saccade refers to the fast ballistic movements of eyes which help in changing the focus of visual attention. The duration of saccades can range from 10 to 100 milliseconds \cite{duchowski2007eye}. Saccades are the fastest movement any human organ can make with a peak velocity upto 900 degrees/second.

 \subsection{Smooth pursuit}

 Pursuit movements are generated when the eyes are following a moving object. The velocity of eye movement is adjusted to keep the moving stimulus in the fovea. Tracking is carried out with catch-up saccades if the velocity of the moving object is high.

 \subsection{Vergence}
 Vergence is the movements of eyes in the opposite directions. Vergence movements help in focusing the eyes on distant objects and depth perception. There are two subcategories in vergence movements they are\\
 Divergence -- The simultaneous movement of both the eyes away from each other\\ , and 
 Convergence -- Movement of both eyes together in the inward direction.

 \subsection{Nystagmus}

 These are conjugate type movements \cite{duchowski2007eye}. There are two types of nystagmus movements: Vestibular nystagmus which compensates for the head movements and Optokinetic nystagmus which compensate for the retinal movement of the target.


\section{Applications of eye gaze tracking}

Earlier, the uses of eye gaze tracking (EGT) were limited to scientific studies in controlled conditions. Typical applications included the study of psychology, ophthalmology, neurology,  oculomotor characteristics and abnormalities \cite{sengupta2017multimodal}, \cite{sengupta2016alertness}, \cite{dasgupta2015drowsiness}. Recently there has been a surge in applications of eye gaze tracking including human-computer interaction \cite{morimoto2005eye}, usability studies, psychology studies, biometrics \cite{kasprowski2004eye}, virtual reality \cite{pai2016gazesim}, gaze-contingent displays \cite{duchowski2004gaze}, usability research, disease diagnosis and gaming \cite{duchowski2007eye}.

The applications of EGT can be broadly classified into two categories \cite{duchowski2007eye},  diagnostic and interactive. In diagnostic applications, eye gaze data is used to estimate users visual and attention processes.  Interactive applications treat eye gaze data as an input modality to
interact with machines.

Duchowski \cite{duchowski2002breadth} divides the interactive applications into two, selective and gaze-contingent.  In selective paradigm, the eye gaze is used as a pointing device similar to a computer mouse. Gaze-contingent paradigm describes a display system which depends on the foveated region of the eye gaze.

Inherently, human eyes function as an input device, intended to perceive visual stimuli. However, eye movements have certain advantages in human-computer interaction. Eye movements are much faster than hand movements (with saccadic peak velocities up to 900 degrees/second). The user usually looks towards the target location before making the mechanical movement using the hands \cite{jacob2003eye}. However, using eye gaze directly as a replacement for the mouse has several issues like lack of a click mechanism (Midas touch problem). Owing to these problems researchers have found efficient methods for incorporating eye gaze along with traditional mouse based interaction. Zhai \textit{et al.} \cite{zhai1999manual} introduced MAGIC, an approach which uses both eye gaze and mouse input for a more efficient cursor movement control. The point of gaze also gives an explicit indication of user's point of attention. This information can be used in HCI to identify users context for different actions and to respond accordingly.  A detailed description of mouse warping using eye gaze can be found in \cite{zhai1999manual}.

The recent advancements in the areas of virtual and augmented reality can also be benefited from eye tracking technology \cite{pai2016gazesim}. Humans have foveated vision where maximum visual information is obtained from the region which is focused on the fovea. Virtual reality headsets can use this information to render high resolution at the locations where the user is looking \cite{patney2016perceptually}. This improves the perception of the visual scene at a reduced computational load for rendering the image. Eye movements can also be used as an interaction modality, either as a pointing device or for eye based typing \cite{majaranta2014eye},\cite{majaranta2002twenty}.

The peak velocity-duration (PV-D) ratio \cite{ueno2002dynamics} has been reported as a good indicator of fatigue level. Certain patterns of eye movements known as eye accessing cues (EAC) \cite{diamantopoulos2009critical} has been known to be related to cognitive processes.  Eye movements are also helpful in identifying certain kinds of diseases such as nystagmus, schizophrenia and autism \cite{leigh2015neurology},\cite{levy2010eye}.

Assistive technology is another area where eye tracking technology can be of much use. In motor neuron diseases like amyotrophic lateral sclerosis (ALS), eye tracking opens the possibility to use eye movements as an interaction channel for persons who are paralyzed \cite{majaranta2011gaze}.

\section{Objectives and scope of the thesis}

 From the above discussions, it is evident that eye tracking technology can prove to be very useful in various domains. The focus of this work is the development of algorithms for gaze tracking and its applications. The research issues and the objectives are outlined here.

Pupil localization is one of the most crucial stage in gaze tracking. Most of the algorithms available in literature fail in challenging situations such as motion blur, head movement, movement of iris towards corners, illumination variations, and partial occlusions. Robust algorithms need to be developed for pupil localization for these conditions. Methods available for gaze tracking typically use a calibration stage. However, this cumbersome calibration stage can be avoided for many applications where only the direction of gaze relative to the head is required. Appearance-based methods can be developed for gaze direction classification without the need of explicit calibration. There have been several attempts to use eye movements as a biometric modality. However, the accuracy of most of the methods has not been satisfactory \cite{kasprowski2004eye}. Information from saccades and fixations can be used efficiently to improve the accuracy of these systems. Eye tracking data from head-mounted eye trackers can be used for activity classification. Combining eye movement patterns along with the visual features and head motion could be more accurate in classifying activities.

The objectives of this work are listed below.


\begin{itemize}
\setlength\itemsep{0em}

\item \textbf{Objective 1.} To develop image-based algorithms for gaze tracking 
\begin{itemize}
\item for desktop environments 
\item for head-mounted eye trackers with  NIR (Near Infrared) illumination
\end{itemize}

\item \textbf{Objective 2.} To develop a person-independent system for gaze direction classification which can be used for cognitive state identification with eye accessing cues. 

\item \textbf{Objective 3.} To develop an efficient framework for eye movement based biometrics

\item \textbf{Objective 4.} To develop a new framework for activity recognition using eye tracking
data and image based features

\end{itemize}

\section{Contributions of the thesis}

The contributions of this work can be outlined as:

\begin{itemize}
 \setlength\itemsep{0em}

\item A fast and accurate two-stage iris center localization algorithm for gaze tracking in low-resolution video.

\item A robust pupil localization algorithm for head-mounted eye trackers.

\item A person independent method for gaze direction classification.

\item A new framework for biometric identification using eye movements.

\item A framework for egocentric activity recognition using eye movements, ego-motion, and visual features.
\end{itemize}

\section{Thesis organization}

The organization of the thesis is given as:
\begin{itemize}
 \setlength\itemsep{0em}

\item \textbf{Chapter 1. Introduction}

  This chapter gives a brief introduction to eye tracking and its applications. It also discusses the motivations behind the work along with objectives and contributions.

\item \textbf{Chapter 2. Eye localization for gaze tracking in low-resolution images}

 A framework has been developed for image based gaze tracking in desktop environments using an efficient iris center localization algorithm in this chapter. 
 
\item \textbf{Chapter 3. Pupil center localization algorithm for NIR images}

  This chapter describes the development of a robust algorithm for pupil localization in NIR images in uncontrolled conditions.
\item \textbf{Chapter 4. Eye gaze direction classification using Convolutional Neural Network}
A convolutional neural network based approach is developed for classification of eye gaze direction which in turn helps in finding eye accessing cues.

\item \textbf{Chapter 5. Eye movement-based biometric authentication}
 A score level fusion approach using a large set of features extracted from fixations and saccades for biometric authentication from eye tracking data is presented in this chapter.
   
\item \textbf{Chapter 6. Activity recognition from head mounted eye tracker}

  A framework for recognition of human activities from egocentric video and eye tracking is presented. 
\item \textbf{Chapter 7. Conclusions and future scopes}

This chapter concludes the work and discusses the future scope of the work presented in this thesis.

\end{itemize}

\end{onehalfspacing}

%% file: Chapter3/chapter_iet_v11.tex
\chapter{Eye Localization for Gaze Tracking in
Low-Resolution Images}{Eye Localization for Gaze Tracking in
Low-Resolution Images}
\graphicspath{{Chapter3/pics/}}
\begin{mdframed}[linecolor=grey!3,backgroundcolor=grey!3] 
 
This chapter presents an efficient framework developed for image based gaze tracking in desktop environments. The challenging problem of iris center localization is solved using a  novel two-stage algorithm. With the proposed framework, even low-cost consumer-grade webcams can be used for gaze tracking without any additional hardware. 
\end{mdframed}
\vspace{5mm}

\begin{onehalfspacing}

\section{Introduction}

Localization and tracking of the eye can be useful in face alignment, gaze tracking and human-computer interaction \cite{hansen2010eye}. The majority of the commercially available eye trackers use active IR illumination. However, IR-based methods need extra hardware and specifically zoomed cameras that limit the movement of the head. Further, the accuracy of IR-based method falls drastically in uncontrolled illumination conditions. An image-based algorithm for localizing and tracking the eye in the visible spectrum is proposed in this chapter. The main advantage of such a method is that it does not require any additional hardware and can work with regular low-cost webcams. 

Several approaches have been reported in the literature for the detection of iris center in low-resolution images. These methods can be broadly classified into four categories 1) Model-based methods, 2) Feature-based methods, 3) Hybrid methods, and 4) Learning-based methods.
Model-based approaches generally approximate iris as a circle. The accuracy of such methods may deteriorate when model assumptions are violated. In feature-based methods \cite{hansen2010eye}, local features like gradient information, pixel values, corners, isophote properties, etc. are used for the localization of iris center (IC). Hybrid methods combine both local and global information for higher accuracy than one particular method alone. Learning-based methods \cite{zhang2015appearance} try to learn representations from labeled data rather than using heuristic assumptions. 

Typically, the resolution of the front facing camera is limited in smart devices like laptops, desktops and mobile devices. For laptops and commercially available webcams, VGA resolution ($640 \times 480$) is a very common resolution. The size of eye patches with this resolution is in the range of $50 \times 40$. We develop the algorithms such that the performance is more than the acceptable levels for VGA resolution and above. However, the proposed approach works for even lower resolution images. A hybrid approach for the accurate detection and tracking of iris center in low-resolution images is presented here. A two-stage algorithm is proposed for localizing the IC. A novel convolution operator is derived from Circular Hough Transform (CHT) for IC localization. The new operator is efficient in the detection of IC even in partially occluded conditions and at extreme corner positions. Additionally, an edge-based refinement and ellipse fitting are carried out to estimate the IC parameters accurately. IC and eye corners are used in a regression framework to determine the point of gaze (PoG). 
 
The important contributions from this chapter are:
\begin{itemize}
\item A novel hybrid convolution operator for the fast localization of iris center
\item An efficient algorithm that can estimate the iris boundary in low-resolution grayscale images
\item A framework for the eye gaze tracking in low-resolution image sequences.
\end{itemize}


\section{Related works}

The localization of iris or pupil is an important stage in gaze tracking. Once the iris center has been successfully localized, regression-based methods can be used for finding the corresponding gaze points on the screen. Most of the passive image-based methods treat iris localization as a circle detection problem. Circular Hough Transform (CHT) is a standard method used for detection of circles \cite{illingworth1988survey}. Young \textit{et al.} \cite{young1995specialised} reported a method for the detection of iris using specialized Hough transform and tracking using active contour algorithm. However, this method requires high-quality images obtained from a head mounted camera.

Smereka \textit{et al.} \cite{smereka2008circular} presented a modified method for the detection of circular objects. They used the votes from each sector along with the gradient direction to detect circle locations. Atherton \textit{et al.} \cite{atherton1999size} proposed phase combined orientation annulus (PCOA) method for the detection of circles with convolutional operators. The annulus is convolved with the edge image to detect the peaks. Peng Yang \textit{et al.} \cite{yang2004novel} presented an algorithm for first localizing the eye region with Gabor filters and then localizing the pupil with a radial symmetry measure. However, the accuracy of the method falls when the iris moves to corners. Valenti \textit{et al.} proposed \cite{valenti2012accurate} an isophote property based iris centre localisation algorithm. The illumination invariance of isophote curves along with gradient voting is used for the accurate detection of iris centers. This method is further extended in \cite{valenti2012combining} for scale invariance using scale-space pyramids. The face pose and iris center obtained are combined to determine the point of gaze (PoG) achieving an average accuracy of 2-5 degrees in unconstrained situations. The accuracy of the method deteriorates when iris moves towards the corners resulting in false detection of eyebrows and eye corners as iris centers. Timm \textit{et al.} \cite{timm2011accurate} proposed a method using gradients of the eye region. An extensive search is carried out in all pixels maximizing the inner product of the normalized gradient and normalized distance vector. IC is obtained as the maximum of weighted function in the region of interest. The time taken for search increases with increase in the search area. The performance of the algorithm degrades in noisy and low-resolution images where the edge detection method fails.

D'Orazio \textit{et al.} \cite{d2002ball} have reported a method for detection of iris centre using convolution kernels. The kernels are convolved with the gradient of images and peak points are selected as candidates. The mean absolute error (MAE) similarity measure is used to reject false positive cases. Daugman \cite{daugman2004iris} proposed an integro-differential operator (IDO) for the accurate localization of iris in IR images. Curve integral of gradient magnitudes is computed to extract the iris boundary. Recently Baek \textit{et al.} \cite{baek2013eyeball} presented an eyeball model based method for gaze tracking. Elliptical shapes for eye model is saved in the database and used at the time of detection for finding the iris centers. A combined IDO and a weighted combination of features are used for the localization of iris center. Polynomial regression methods were used for training the system. They obtained average accuracy of 2.42 degrees visual angles. Sewell and Komogortsev \cite{sewell2010real} developed an artificial neural network based method for gaze estimation from low-resolution webcam images. They trained the neural network directly with the pixel values of the detected eye region. They obtained an average accuracy of 3.68 degrees. Zhou \textit{et al.} \cite{zhou2004projection} proposed a generalized projection function (GPF) that uses various projection functions and a special case hybrid projection function for localizing the iris center. The peak positions of vertical and horizontal GPF were used to localize the eye. Bhaskar \textit{et al.} \cite{bhaskar2003blink} proposed a method for identifying and tracking blinks in video sequences. Candidate eye regions are identified using frame differencing and are subsequently tracked using optical flow.  The direction and magnitude of the flow are used to determine the presence of blinks. They obtained an accuracy of 97\% in blink detection. Wang \textit{et al.} \cite{wang2003eye} proposed one-circle method where the detected iris boundary contours are fitted with an ellipse and back projected to find the gaze points. Recently many learning based methods have been proposed for iris center localization and gaze tracking. Markuš \textit{et al.} \cite{markuvs2014eye} proposed a method for localising pupil in images using an ensemble of randomized trees. They used a standard face detector to localize face and eye regions. Ensemble of randomized trees model was trained using the eye regions and ground truth locations. Their method obtained good accuracy in BioID database. However, the accuracy of gaze estimation was not discussed in their work. Zhang \textit{et al.} \cite{zhang2015appearance} proposed an appearance based gaze estimation framework based on Convolutional Neural Network (CNN). They trained a CNN model with a large amount of data collected in real-world conditions. Normalized face images and the head poses obtained from a face detector were used as the input to the CNN to estimate the gaze direction. They obtained good accuracy in person and pose independent scenarios. However, the accuracy for person dependent case is lower than current geometric model based methods. The accuracy might increase with larger amount of training data, but the time taken for on-line data collection and training becomes prohibitive. Schneider \textit{et al.} \cite{schneider2014manifold} proposed a manifold alignment based method for appearance based person independent gaze estimation.  From the registered eye images, a wide variety of feature extraction methods like LBP histogram, HoG, mHoG and DCT coefficients were extracted. A combination of LBP and mHOG based features obtained the best performance. Several regression methods were used for appearance based gaze estimation. Sub-manifolds for each individual were obtained using the ground truth gaze locations. Synchronized delaunay sub-manifold embedding (SDSE) method was used to align the manifolds of different persons. Even though their method achieved better performance compared to other appearance-based regression methods, the effect of head pose variations on the accuracy was not discussed.

Sugano \textit{et al.} \cite{sugano2014learning} proposed a person and head pose independent method for appearance based gaze estimation. They captured the images of different persons using a calibrated camera, and images corresponding to various head poses were synthesized.  An extension of random forest algorithm was used for training. The appearance of eye region and the head pose was used as the input to the algorithm which learns a mapping to the 3D gaze direction. 

  Most of the methods proposed in literature fail when iris moves towards the corners. Another problem is regarding eye blinks, most of the algorithms return false positives when the eyes are closed. A stable reference point is required along with the IC location for PoG estimation. Learning-based methods require large amounts of labeled data for satisfactory performance. The performance of such methods also deteriorates when imaging conditions are different. Training for person dependent models require huge amount of data and often require a considerable amount of time. This limits the deployment of such methods in mobiles, tablets, etc.
  
In the proposed method, IC can be accurately localized even in extreme corner locations using the ellipse approximation. The computatitional load is reduced using the two-stage scheme. Further, an eye closure detection stage is added to prevent false positives. The localization error can be minimized by tracking the IC in the subsequent frames. The estimated IC is used in a regression framework to estimate the PoG.

\section{Proposed algorithm}

Different stages of the proposed framework are described here. 
\subsection{ Face detection and eye region localization}
Knowledge of the position and pose of the face is an essential factor in determining the point of gaze. Detection and tracking of the face help in obtaining candidate regions for eye detection. This reduces the false positive rate as well as computation time. Haar-like feature based method \cite{viola2001rapid} is used for face detection because of its higher accuracy and faster execution. An improved implementation of face detection and tracking has been proposed in our earlier work \cite{dasgupta2013vision} (shown in Appendix A). The modified algorithm can detect in-plane rotated images using affine transform based algorithm. The computation is carried out in the down sampled images to make the detection faster. The search space of detection algorithm is dynamically constrained based on the temporal information, which further increases the speed of face detection. Kalman filter-based tracking is used to predict the location of the face when it is not detected.    This also helps in minimizing false detections. The de-rotated eye region obtained is used in subsequent stages. This makes the performance of the algorithm invariant to in-plane rotations. The purpose of the de-rotation stage is only to provide a de-rotated region of interest (ROI) for the further processing stages. The accuracy of face rotation estimation in the pre-processing stage is only up to ±15 degrees. More accurate in-plane face rotation is obtained in the later stage using the angle of the line connecting the inner eye corners. With the improved face-tracking scheme, the frame rates of processing increase greatly (up to 200 frames per second). The analysis and tradeoffs of the algorithm has been presented in \cite{dasgupta2013vision}.

\subsection{ Iris center localization}
\begin{figure}[htb]
\centering
\includegraphics[width=0.9\linewidth]{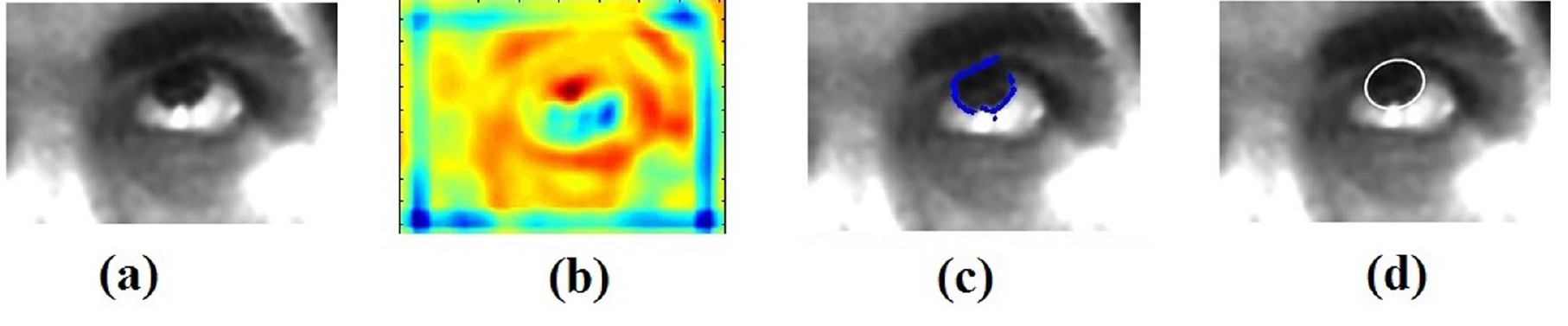}
\caption{Stages in ellipse fitting:  (a) Cropped eye region, 
 (b) Correlation surface from the proposed operator,
 (c) Selected candidate boundary points,
 (d) Fitted ellipse.}
\label{fig:1}
\end{figure}

The method proposed here use a coarse to fine approach for detecting the accurate center of the iris. The two-stage approach reduces the computational requirement as well as false detection rate. The outputs of various stages in IC localization are shown in Fig. \ref{fig:1}.
\subsubsection{Coarse iris center detection} In this stage, iris detection is formulated as circular disc detection. An average ratio between the width of face and iris radius was obtained empirically. For a particular image, the range of the radius is computed using this ratio and width of the detected face. The image gradient of iris boundary points will always be pointing outwards. The gradient directions and intensity information is used for the detection of eyes. The gradients of the image are invariant to uniform illumination changes. 

A novel convolution operator is proposed to detect peak location corresponding to the center of the circle. A class of convolution kernels, known as Hough Transform Filters \cite{atherton1999size} are used for this purpose. In CHT filter, the 3D accumulator is collapsed to a 2D surface by selecting a range of the radii.  

The 2D accumulator can be calculated efficiently using a convolution operator. Thus a CHT filter is derived, which acts directly upon the image without any requirement of edge detection. A vector convolution kernel is designed for correlating with the gradient image, which gives a peak at the center of the iris. 

The convolution operator is designed as a complex operator with magnitude unity. The operator detects a range of circles by taking dot products with orientations inside the radius range. The equation is similar to orientation annulus proposed by Atherton \textit{et al.} \cite{atherton1999size}. The equation of convolution kernel is given as

\begin{equation}
\begin{array}{*{20}{c}}
{}\\
{{O_{COA}}\left( {m,n} \right)}\\
{}
\end{array} = \left\{ {\begin{array}{*{20}{c}}
{\frac{{\cos {\theta _{mn}} + i\sin {\theta _{mn}}}}{{\sqrt {{m^2} + {n^2}} }},iff,R_{\min }^2<{m^2}+{n^2}< R_{\max }^2}\\
{0, otherwise}
\end{array}} \right.
\end{equation}

where,

\begin{equation}
{\theta _{mn}} = {\tan ^{ - 1}}\left( {{\raise0.7ex\hbox{$n$} \!\mathord{\left/
 {\vphantom {n m}}\right.\kern-\nulldelimiterspace}
\!\lower0.7ex\hbox{$m$}}} \right)
\end{equation}


Where $m$ and $n$  denote the coordinates of the kernel matrix with respect to the origin. The operator is scaled for equal contributions of circles in the radius range. A weighting matrix kernel ($W_{A}$) is also used for finding regions with maximum dark values

\begin{equation}
{W_A}\left( {m,n} \right) = \left\{ {\begin{array}{*{20}{c}}
{\frac{1}{{\sqrt {{m^2} + {n^2}} }},iff,{m^2} + {n^2} < R_{\max }^2}\\
{0,otherwise}
\end{array}} \right.
\end{equation}

The gradient complex orientation annulus is given as,

\begin{equation}
{C_{GCOA}} = {\mathop{\rm Re}\nolimits} \left( {{O_{COA}}} \right) \otimes {S_x} + i{\mathop{\rm Im}\nolimits} \left( {{O_{COA}}} \right) \otimes {S_y}
\end{equation}

Where $\otimes$ denotes the convolution operator; ${S}_x$ and ${S}_y$  denote the $3 \times 3$ Schaar kernels in $x$  and $y$  directions respectively. Schaar differential kernel is used owing to its mathematical properties in gradient estimation. In most of the cases, the upper portion of the iris is occluded by eyelids. An additional weighing factor ($\beta$) is included to increase the contribution of horizontal gradients. The convolution kernel can be made real-valued as

\begin{equation}
{C_{RCC}} = \beta {\mathop{\rm Re}\nolimits} \left( {{O_{COA}}} \right) \otimes {S_x} + \frac{1}{\beta }{\mathop{\rm Im}\nolimits} \left( {{O_{COA}}} \right) \otimes {S_y}
\end{equation}

Where $\beta $ denotes the weighting factor.
The average intensity of each point in image can be obtained by convolving the weighting kernel with the negated version of the image as,

\begin{equation}
W = \left( {255 - I} \right) \otimes {W_A}
\end{equation}

Where $I$  and  ${W}_A$  denote the image and the kernel for computing the intensity component respectively. The final correlation output ($CO$) can be obtained by combining the convolution results for both gradient and intensity kernels as,

\begin{equation}
CO = \lambda \left( {I \otimes {C_{RCC}}} \right) + \left( {1 - \lambda } \right)W
\end{equation}

Where, $\lambda  \in [0,1]$ is a scalar used to obtain the weighted combination of gradient information and image intensity to reduce spurious detections.  Iris center corresponds to the maximum of the correlation surface $CO$. Further, it is possible to represent all these operations with a single real convolution kernel, which can be applied on the image without any pre-processing, making the iris center localization procedure even faster. For bigger circles, convolution can be carried out in Fourier domain for enhancing the speed of the computation. 

The peak of correlation output alone may lead to false detections in partially occluded images. Here, peak to side lobe ratio (PSR) of the points are used to find the iris location. The PSR values calculated in each of the local maxima and the point with maximum PSR is considered as the iris center. The PSR is estimated as:

\begin{equation}
PSR = \left( {\frac{{C{O_{\max }} - \mu }}{\sigma }} \right)
\end{equation}

Where $C{O_{\max }}$  is the local maxima in the correlation output,  $\mu $ and  $\sigma $  are the mean and standard deviation in the window around the local maxima. We have used a window size of $11 \times 11$ in this work. The point with the maximum  $PSR$ is selected as the iris center.

\subsubsection{Sub-pixel edge refining and ellipse fitting}

In this stage, the approximate center points obtained in the previous stage are used to refine the IC location. The objective is to fit the iris boundary with an ellipse. The constraints on the major and minor axis can be obtained empirically ( $R_{min}$ and $R_{max}$  ). The algorithm presented searches in the radial direction similar to Starburst algorithm \cite{li2005starburst}. However, the search process finds only the strongest edges with similar gradients. Dominant edges with agreeing directions are selected with sub-pixel accuracy. An angle versus distance plot is obtained, and the outlier points are filtered using median filter. An ellipse can be fitted to five points by the least square method using Fitzgibbon's algorithm \cite{fitzgibbon1999direct}. However, we used this algorithm in a random sample consensus (RANSAC) framework for minimizing the effect of outliers.  RANSAC algorithm is employed \cite{fischler1981random} for ellipse fitting, using the gradient agreement \cite{swirski2012robust} of the detected boundary points and the fitted ellipse as the support function. Additionally, a modified goodness of fit (GoF) is evaluated as the integral of dot products of outward gradients over the detected boundary (only agreeing gradients). The parameters obtained are considered as false positives if the goodness of fit is less than a threshold. The detailed algorithm for ellipse fitting is given in Algorithm 1.

\begin{equation}
GoF = \sum\limits_{x,y \in f(\lambda )}^{} {\left( {\min \left( {\frac{{\nabla f(x,y)}}{{\left| {\nabla f(x,y)} \right|}} \bullet \nabla I(x,y),0} \right)} \right)} 
\end{equation}

where, $f(\lambda)$ and $\nabla f(x,y)$ denote the fitted ellipse and its derivative respectively. $\nabla I(x,y)$ denotes the image derivative at position $(x,y)$, belonging to the fitted ellipse.
\subsection{Iris tracking}
A Kalman filter (KF) \cite{yoon2008new} is used to track the IC in a video sequence. The search region for iris detection can be localized with the tracking approach. Once the IC is detected with sufficient confidence, the point can be tracked in subsequent frames using the dynamics of eye motion. Face detection stage can be avoided in this case. The KF \cite{kiruluta1997predictive} estimates can be used as the corrected estimates for iris position. It is to be noted that the objective here is not to model the dynamics of saccade, but to get a smoother estimate of IC location, which can be useful for reducing search region for IC localization in next frame.

\begin{algorithm}[H]
\caption{Algorithm for iris boundary refinement}
\label{alg:irisboundary}
\begin{algorithmic}[1]

\INPUT The grayscale eye region $I$ and the estimated centres
$O = \left( {{c_x},{c_y}} \right)$
\OUTPUT Fitted ellipse parameters $\lambda  = ({c_x},{c_y},a,b,\psi )$
\State \textbf{\textit{Initialize}}: \textit{CandidatePoints}= NULL
\State tempBest=[0,0]

\For {$\theta \leftarrow$ 0 \textbf{to} $2\pi$}

 \For { $r \leftarrow {R_{\min }}$ to ${R_{\max }}$ } 
\State $Pt = ({c_x} + r\cos \theta ,{c_y} + r\sin \theta )$
\State Calculate the gradients and magnitude at the points
\State $\vec g = \frac{{{G_x}\hat x + {G_y}\hat y}}{{\left\| G \right\|}}$

\If{$\left\| g \right\| < threshold$}
\State Break
\Else

\State $\vec r = r\cos \theta \hat x + r\sin \theta \hat y$
\State Calculate the dot product of normalized gradient vector

\If{$\cos \theta  = \vec r \bullet \vec g > threshold$}

\If{ $tempbestmag < Pt.mag,Pt.angle$}

\State $temp\_best = Pt$
\State $tempbestmag = [\cos \theta ,\left\| g \right\|]$

\Else
\State Continue

\EndIf

\EndIf

\EndIf

\EndFor
\State $CandidatePoints.append(temp\_best)$
\EndFor
\State \textit{\textbf{Filter}} the detected points with angular median filter
\State \textit{\textbf{Fit Ellipse}} with RANSAC algorithm
\State \textbf{Return} the parameters of ellipse: $\lambda \leftarrow ({c_x},{c_y},a,b,\psi )$

\end{algorithmic}
\end{algorithm}

In the current tracking application, constant velocity model is chosen as the transition model. Coordinates of the center of iris along with their velocities are used as states. 

\begin{equation}
{X_{k + 1}} = {F_k}{X_k} + {W_k}
\end{equation}

Where, ${X_k}$ is the state containing  $x,y,{v_x},{v_y}$ (coordinates and velocities in  $x$  and $y$   directions respectively) at the ${k^{th}}$ instant.  The measurement noise covariance matrix is computed from the measurements obtained during the gaze calibration stage. The process covariance matrix is computed empirically. Measurements obtained from the IC detector are used to correct the estimated states. 

\subsection{Eye closure detection}
The IC localization algorithm may return false positives when the eyes are closed. Thresholds on the peak magnitude were used to reject false positives. However, the quality of peak may degrade in conditions such as low contrast, image noise, and motion blur. The accuracy of the algorithm may fall in these conditions, and hence a machine learning based approach is used to classify the eye states as open or close. The Histogram of oriented gradients (HOG) \cite{dalal2005histograms} features of eye regions are calculated and a support vector machine (SVM) based classifier is constructed to predict the state of the eye. The HOG features are computed in the detected ROI for left and right eyes separately. The SVM classifier was trained offline from the database. If the eye state is classified as closed, then the predicted value from KF is used as the tentative position of the eye. If eyes are detected as open, then the result from the two-stage method is used to update the KF.

\subsection{Eye corner detection and tracking}
The appearance of inner eye corner exhibits insignificant variations with eye movements and blinks. Therefore, we propose to use inner eye corners as reference points for gaze tracking. The eye corners can be located easily in the eye ROI. The vectors connecting eye corners and iris centers can be used to calculate gaze position. Several methods have been proposed in the literature for the localization of facial landmarks \cite{cristinacce2006facial}. In the proposed method, Gabor jets \cite{vukadinovic2005fully} are used to find eye corners in the eye ROI owing to its high accuracy.  The detected eye corners are tracked in the subsequent frames using optical flow and normalized cross correlation (NCC) \cite{lewis1995fast}, \cite{tomasi1991detection} based method.  The tracker is automatically reinitialized if the correlation score is less than a pre-set value.

\subsection{Gaze estimation}

Gaze point can be computed from the IC location and a reference point. Earlier works \cite{pires2013visible}, \cite{sigut2011iris} have used the eye centre and corneal reflections as reference points. False detection in any of the corners will result in performance degradation of the algorithm. Hence, the inner eye corners are used as reference points in this work. Detection of the eye corner in every frame might increase the error rates and computational load. We avoid this issue by tracking of eye corners in the frames which ensure stable reference points. If $({x_1},{y_1})$ and $({x_2},{y_2})$ denote the coordinates of eye corner and iris centre respectively, the eye corner-iris center (EC-IC) vector can be obtained as: $(x,y) = ({x_2} - {x_1},\,\,{y_2} - {y_1})$ (with reference to the corner). The EC-IC vector is calculated separately for the left and right eye.

\subsubsection{Calibration}
In the calibration stage, subjects were asked to look at uniformly distributed positions on the screen. The EC-IC vectors along with gaze points are recorded. The mapping between EC-IC vector and screen coordinates is nonlinear because of the angular movement of the iris. We used two different models for the mapping between EC-IC vector and point of gaze (PoG), 1) polynomial regression and 2) a radial basis function (RBF) kernel based method. In polynomial regression, a second order regression model is used for determining the point of gaze since it offers the best trade-off between model complexity and accuracy.

\begin{equation}
screen{X_i} = {a_0}{x_i} + {a_1}{y_i} + {a_2}{x_i}{y_i} + {a_3}x_i^2 + {a_4}y_i^2 + {a_5}
\end{equation}
\begin{equation}
screen{Y_i} = {b_0}{x_i} + {b_1}{y_i} + {b_2}{x_i}{y_i} + {b_3}x_i^2 + {b_4}y_i^2 + {b_5}
\end{equation}

Where, $ \left( {{x_i},{y_i}} \right)$  are the components of EC-IC vector and, $\left( {screen{X_i},screen{Y_i}} \right)$  the corresponding screen positions. The data obtained from calibration stage is used in the least square regression framework to calculate unknown parameters.\\

In the RBF kernel based method, we used non-parametric regression \cite{nadaraya1964estimating} for estimating PoG. The components of EC-IC vector are transformed into kernel space using the following expression,

\begin{equation}
k({p_i},{p_l}) = exp( - \frac{{{{\left\| {{p_i} - {p_l}} \right\|}^2}}}{{2{\sigma _k}^2}})
\end{equation}

Where, ${p_i}$ and  ${p_l}$ denote of EC-IC vector and the landmark points respectively. ${\sigma _k}$  denote the standard deviation of the RBF function. We have tested the algorithm in both $3 \times 3$ and $4 \times 4$ calibration grids. Instead of using all the samples as landmark points, we have used only one landmark per calibration point. The number of landmarks used was 9 and 16 for $3 \times 3$ and $4 \times 4$ grids respectively. For each calibration point on the grid, the landmark vector is calculated as the median of the components of EC-IC vector at the particular point. The dimension of the design matrix is reduced by the use of landmark points (since the data points are clustered around the calibration points). Regression is carried out after transforming all the points to kernel space \cite{kohn2001nonparametric} which improved the accuracy of PoG estimation. The training procedure is carried out for left and right eyes.

\subsubsection{Estimation of PoG}

\begin{figure}[!htb]
\centering
\includegraphics[width=0.95\linewidth]{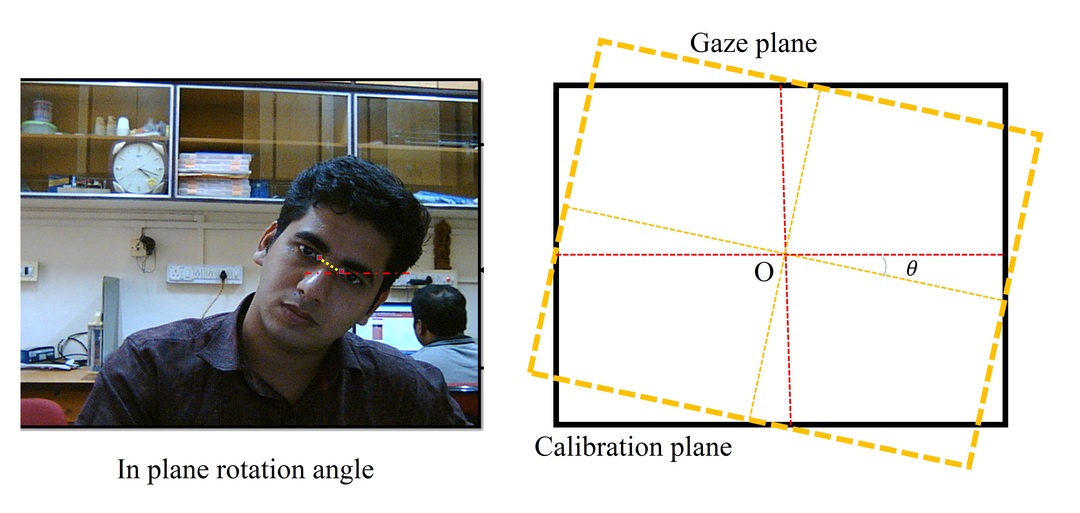}
\caption{Transformation of estimated gaze point to screen coordinates for compensating in-plane rotation.}
\label{fig:2}
\end{figure}

The parameters obtained from calibration procedure are used to determine the gaze position. The regression function obtained is used to map the EC-IC vectors to screen coordinates. The gaze position is computed as the average position returned by the left and right eye models. The head position is assumed to be stable during the calibration stage. After calibration, the estimated gaze point will be on the calibration plane (i.e., w.r.t the position of the face during the calibration stage). Deviation from this face position would cause errors in the estimated gaze locations. The effect of 2D translation is minimal for moderate head movements since the reference points for EC-IC vector are eye corners, which also move along with face (thereby providing a stable reference invariant to 2d translational motion). Even though the method is invariant to moderate amount of translation, the accuracy falls when there is a rotation. This error can be corrected using the face pose information. The in-plane rotation of face can be calculated from the angle of the line connecting the inner corners of the left and right eye as shown in Fig. \ref{fig:2}. The rotation matrix can be computed as,

\begin{equation}
R = \left[ {\begin{array}{*{20}{c}}
{\cos \theta }&{ - \sin \theta }\\
{\sin \theta }&{\cos \theta }
\end{array}} \right]
\end{equation}

Where, $\theta $ is the difference in angle from the calibration stage. The corrected PoG can be found from coordinate transformation with the screen center as the origin. The exact 3D pose variations can be corrected using more computationally intensive models like active appearance models (AAM) \cite{cootes2001active}, constrained local model (CLM) \cite{cristinacce2006feature}, etc.

\section{Experiments}

We have conducted several experiments to evaluate the performance of the proposed algorithm. The algorithm has also been evaluated using standard databases and a custom database.  The IC localization accuracy is evaluated in standard databases and compared with the state of the art methods. The accuracy in PoG estimation and eye closure detection is assessed in the custom dataset.

\subsection{Experiments on IC localization}

\subsubsection{Evaluation method}
Face detection is carried out using Viola-Jones method \cite{viola2004robust}. The eye regions are localized based on anthropometric ratios. 

The normalized error is used as the metric for comparison with other algorithms. The normalized measure for worst eye characteristics (WEC) \cite{jesorsky2001robust} is defined as 
\begin{equation}
{e_{WEC}} = \frac{{\max ({d_l},{d_r})}}{w}
\end{equation}
Where ${d_l}$ and  ${d_r}$ are the Euclidean distances between ground truth and detected iris centres (in pixels) of left and right eye respectively, and    $w$ is the true distance between the eyes in pixels. The average (AEC) and best of eye detection (BEC) errors are also calculated for comparison. They are defined as:
\begin{equation}
{e_{AEC}} = \frac{{({d_l} + {d_r})}}{{2w}},{e_{BEC}} = \frac{{\min ({d_l},{d_r})}}{w}
\end{equation}

Where, ${e_{BEC}}$ is the minimum error in both the eyes and ${e_{AEC}}$ is the average error of both the eyes.

\subsubsection{Experiments in BioID and Gi4E Databases}
A comparison of the proposed method with the state of the art methods is carried out for BioID \cite{BioID} and Gi4E \cite{ponz2012dataset} databases. The BioID database consists of images of 23 individuals taken at different times of the day. The size, position, and pose of the faces change in the image sequences. The contrast is very low in some images. In some images, eyes are closed. There are images where subject wearing glasses and glints are present due to illumination variations. The database contains a total of 1,521 images with a resolution of $384 \times 288$ pixels. The ground truth files for left and right iris centers are also available.

Gi4E dataset consists of 1380 color images of 103 subjects with a resolution of $800 \times 600$. It contains sequences where the subjects are asked to look at 12 different points on the screen. All the images are captured at indoor conditions at varying illumination levels and different backgrounds. The database represents realistic conditions during gaze tracking, head movements, illumination changes and movement of eyes towards corners and occlusions with eyelids. The ground truth of left and right eye positions is also available in the database.

\begin{figure}[!htb]
\centering
\includegraphics[width=1\linewidth]{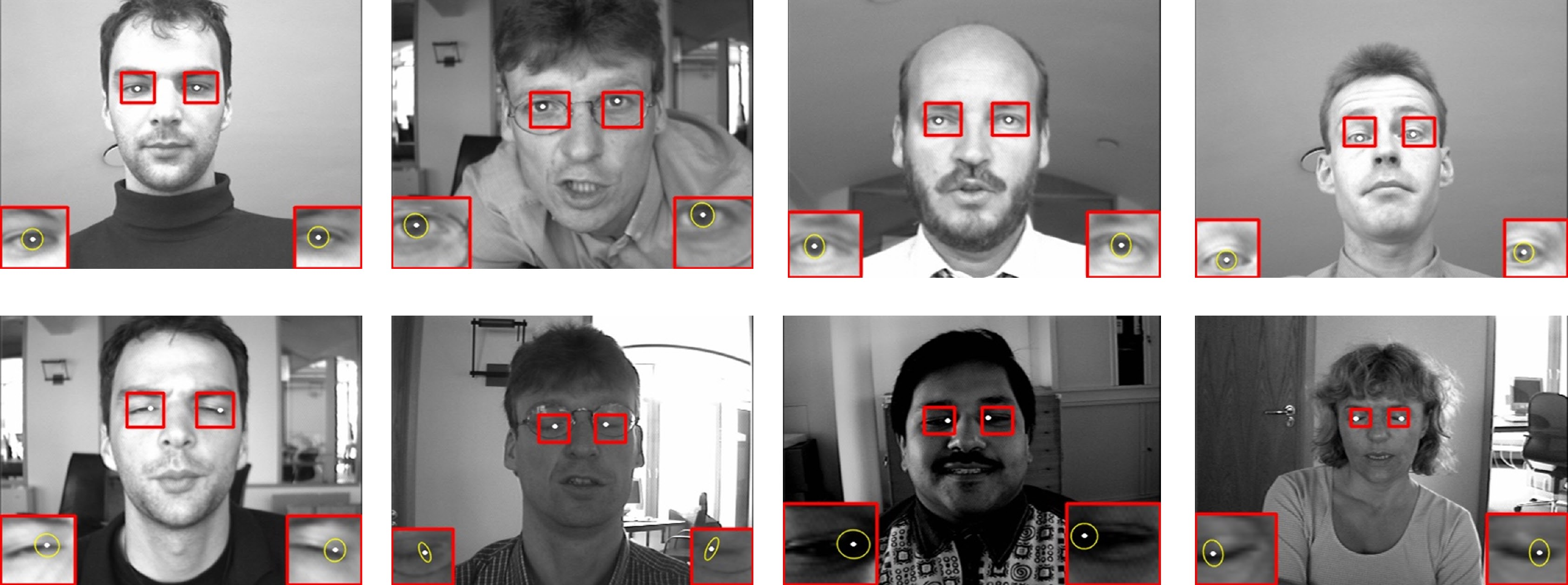}
\vspace{-1.5em}
\caption{Few samples showing successful detections (first row) and failures (second row) in BioID database.}
\label{fig:3}
\end{figure}

\begin{figure}[!htb]
\centering
\includegraphics[width=1\linewidth]{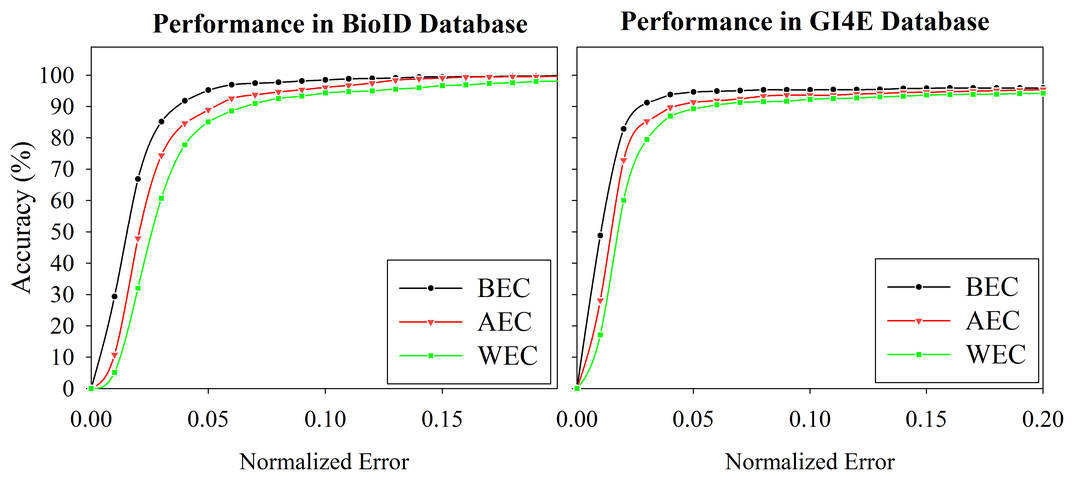}
\caption{Performance of the proposed algorithm in BioID and Gi4E databases. The graph shows three normalized measures corresponding to WEC - Worst eye characteristics, AEC-Average eye characteristics, and BEC- Best Eye characteristics.}
\label{fig:4}
\end{figure}

Fig. \ref{fig:3} show some of the correct detections and failures of the algorithm in BioID database.
An accuracy of 94.74\% is obtained for face detection. In most of the cases, errors are due to partial closure of eyes and eyeglasses. The algorithm performs well when eyes are visible even with low contrast and varying illumination levels. Fig. \ref{fig:4} show the performance of proposed algorithm in BioID and Gi4E database. The value of $\lambda $ and $\beta $ were 0.95 and 2 respectively. The proposed algorithm obtained an accuracy (WEC) of 85.08 for $e \le 0.05$. 


\begin{figure}[!htb]
\centering
\includegraphics[width=1\linewidth]{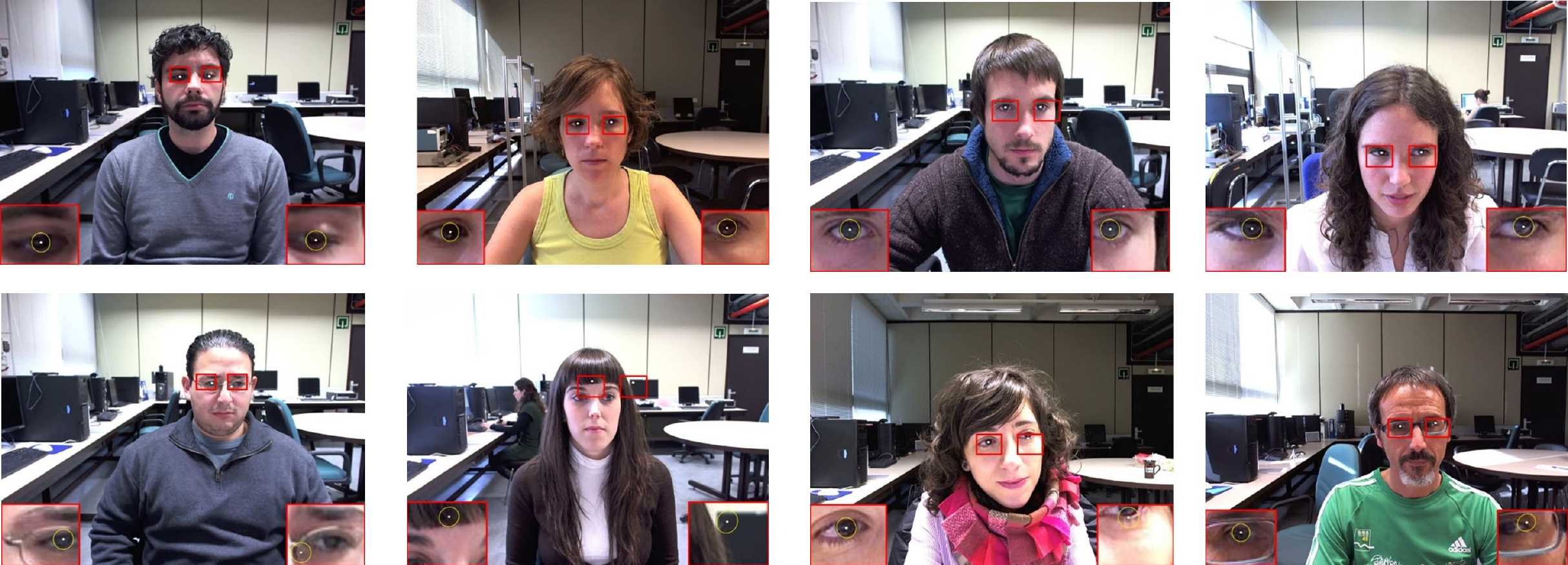}
\caption{Some samples showing successful detections (first row) and failures (second row) in Gi4E Database.}
\label{fig:5}
\end{figure}

In Gi4E database, the worst-case accuracy (WEC) is 89.28 for $e \le 0.05$ . Fig. \ref{fig:5} show results of the algorithm. The accuracy of face detection obtained was 96.95\%.  The main advantage is that the algorithm performs well in different eye gaze positions which is essential in gaze tracking applications.

\begin{figure}[!htb]
\centering
\includegraphics[width=1\linewidth]{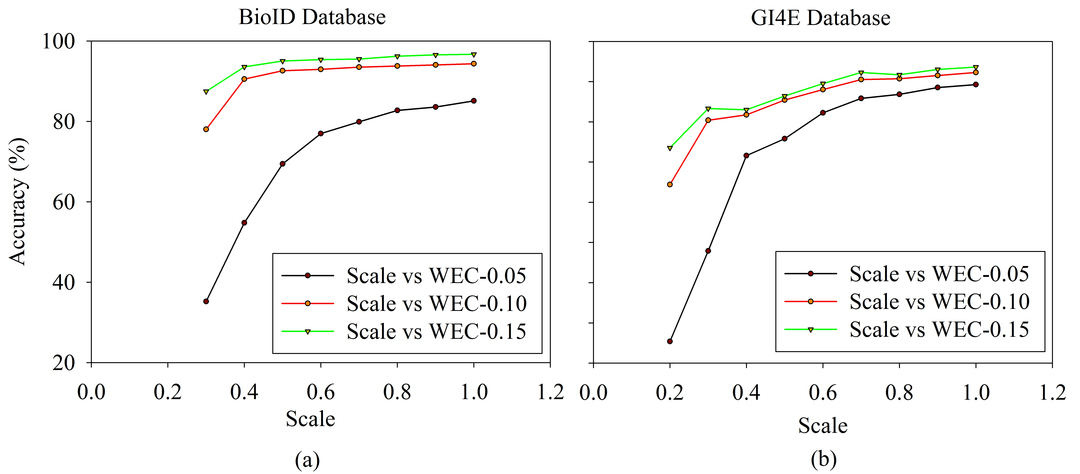}
\vspace{-1.5em}
\caption{WEC Performance of the proposed algorithm in (a) BioID and (b) Gi4E databases with different resolutions. Scaling parameter is w.r.t the original image resolutions in the corresponding databases. }
\label{fig:6}
\end{figure}
The performance of the algorithm may vary depending upon the distance of the user from the monitor. This effect is emulated using images with different spatial resolutions. The performance of the proposed algorithm in different spatial resolutions in BioID and Gi4E database is shown in Fig. \ref{fig:6}. The accuracy of iris localization falls with the image resolution. However, the detection accuracy (WEC-0.5) is more than 80\% for scaling up to 0.8 ($307 \times 230$ resolution) and 0.6 ($480 \times 360$ resolution) in BioID (82.72\%) and Gi4E (82.24\%) databases respectively. 

\subsubsection{Comparison with state of the art methods}
\begin{figure}[!htb]
\centering
\includegraphics[width=1\linewidth]{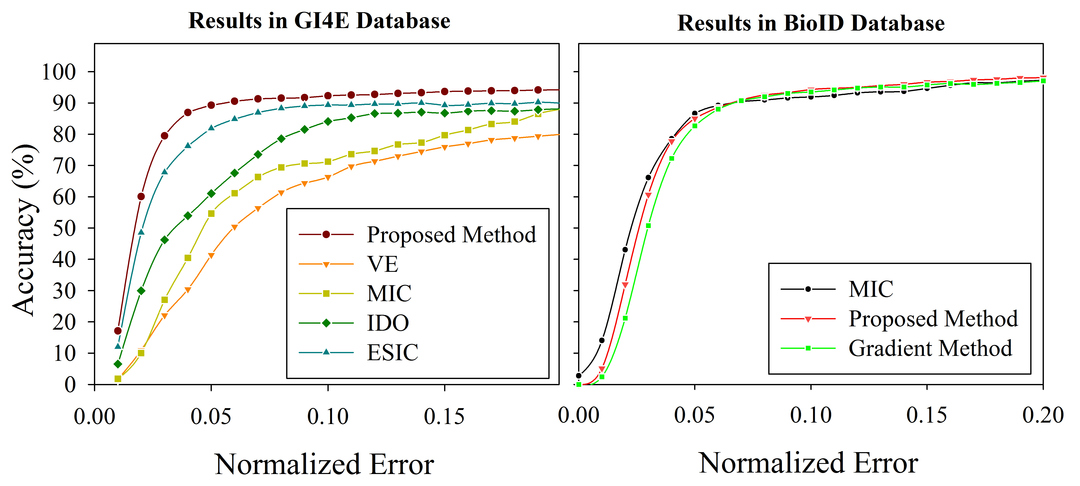}
\vspace{-1.5em}
\caption{ WEC performance comparison of proposed method with state of the art methods in Gi4E and BioID databases. }
\label{fig:7}
\end{figure}

We have compared the algorithm with many state of the art algorithms in BioID and Gi4E databases. The algorithms proposed in Valenti \textit{et al.} \cite{valenti2012accurate} (MIC), Timm \textit{et al.} \cite{timm2011accurate}  and the proposed methods are tested in BioID database. The evaluation is carried out with normalized worst eye characteristics (WEC). The results are shown in Fig. \ref{fig:7}. The WEC data is taken from ROC curves given in author’s papers. The algorithm proposed is the second best in BioID database as shown in Table \ref{tab:2}. Isophote method (MIC) performs well in this database. The proposed algorithm fails to detect accurate positions when eyes are partially or fully closed (eye closure detection stage was not used here). The presence of glints is another major problem. The failure of face detection stage and reflections from the glasses causes false detections in some cases. The addition of a machine learning based classification of local maxima can improve the results of the proposed algorithm.

Gi4E is a more realistic database for eye tracking purposes. It contains images with head and eye movements. The algorithms for comparison are chosen as VE \cite{wang2003eye}, IDO \cite{daugman2004iris}, MIC \cite{valenti2012accurate}, ESIC \cite{baek2013eyeball}. The results are compared with WEC values obtained from ROC curves reported in Baek \textit{et al.} \cite{baek2013eyeball}. It is seen (Table. \ref{tab:1}) that the proposed method outperforms all the other existing methods. The accuracy of MIC method is very low when the eyes move to the corners. The circle approximation of most of the algorithms fails when eyes move towards the corners, making them inapt for eye gaze tracking applications. The performance evaluation is carried out on each frame separately. Addition of temporal information using Kalman filter can increase the accuracy of the algorithm greatly. 
\begin{table}[H]
\centering
\caption{Comparison of proposed method with state of the art algorithms in BioID database}
\label{tab:2}
\begin{tabular}{@{}lcccc@{}}
\toprule
\multicolumn{1}{c}{\textbf{Method}} & $e \le 0.05$     & $e \le 0.10$    & $e \le 0.15$     & $e \le 0.20$      \\ \midrule
MIC{\cite{valenti2012accurate}+Sift kNN }               & 86.09 & 91.67 & 94.5\textsuperscript{*} & 96.9\textsuperscript{*} \\
Proposed                            & 85.08 & 94.3  & 96.67 & 98.13 \\
Timm \textit{et al.} {\cite{timm2011accurate}}                & 82.5  & 93.4  & 95.2  & 96.4  \\ \bottomrule
\multicolumn{5}{l}{\textsuperscript{*}\footnotesize{approximated from graph \cite{valenti2012accurate}}}
\end{tabular}
\end{table}
\begin{table}[H]
\centering
\caption{Comparison of proposed method with state of the art algorithms in Gi4E database}
\label{tab:1}
\begin{tabularx}{1\linewidth}{@{\extracolsep{\fill}}lcccc@{}}
\toprule
\multicolumn{1}{c}{Method} & $e \le 0.05$     & $e \le 0.10$    & $e \le 0.15$     & $e \le 0.20$      \\ \midrule
\textbf{Proposed}          & 89.28 & 92.3 & 93.64 & 94.22  \\
VE                         & 41.4  & 66.3 & 75.9  & 80.0\textsuperscript{*}    \\
MIC                        & 54.5  & 71.2 & 79.7  & 88.1\textsuperscript{*}  \\
IDO                        & 61.1  & 84.1 & 86.7  & 88.15\textsuperscript{*} \\
ESIC                       & 81.4  & 89.3 & 89.2  & 89.9\textsuperscript{*}  \\ \bottomrule
\multicolumn{5}{l}{\textsuperscript{*}\footnotesize{approximated from graph \cite{baek2013eyeball}}}
\end{tabularx}
\end{table}

\begin{figure}[H]
\centering
\includegraphics[width=0.7\linewidth]{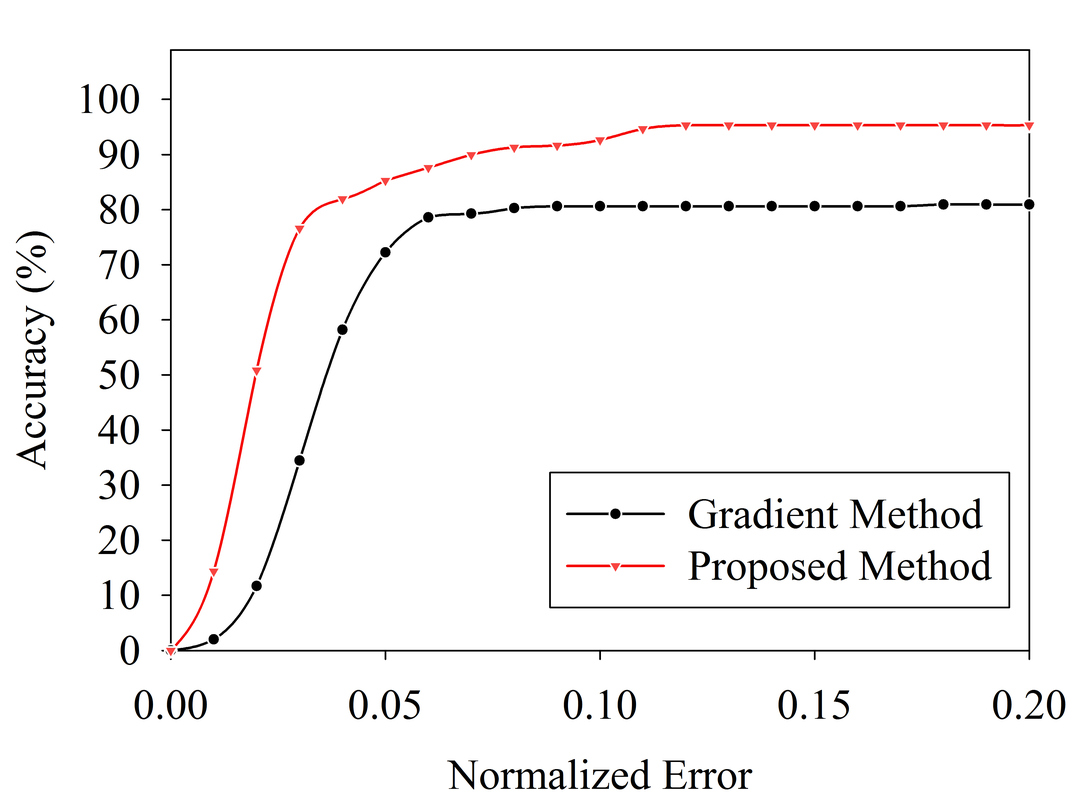}
\caption{WEC performance comparison of the proposed method with gradient based method in extreme corner cases. }
\label{fig:8}
\end{figure}

We have performed additional experiments on the Gi4E dataset to evaluate the performance when the iris moves to the corner. A subset of 299 images was selected according to the position of iris center about the eye corner. We have compared the results with the gradient-based method for evaluating the accuracy with circle as well as the proposed ellipse model. The WEC characteristics comparison is shown in Fig. \ref{fig:8}. It is observed that the ellipse approximation improves the accuracy significantly compared to the circle approximation.


\subsection{Experiment with our database}
\subsubsection{Evaluation of gaze estimation accuracy }

\begin{figure}[!htb]
\centering
\includegraphics[width=1\linewidth]{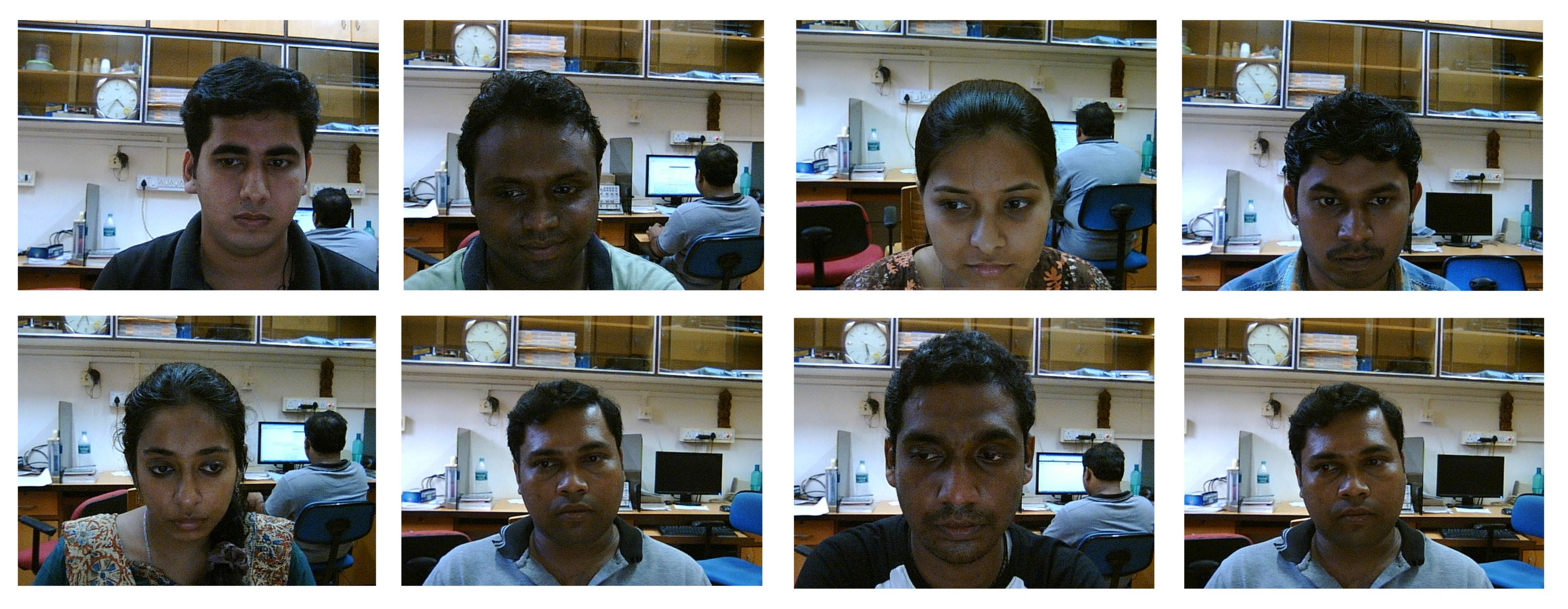}
\caption{ Sample images of subjects in the experiment. }
\label{fig:subjects}
\end{figure}

An experiment was performed on ten subjects using a standard webcam and a 15.6- inch monitor with a resolution of $1366 \times 768$. The subjects were seated 60 cm from the screen and asked to follow the red dot on the monitor. The videos of the eye movements were recorded at 30 fps at a resolution of $640 \times 480$. The subjects were asked to look at the calibration patterns two times. We used both 9 point and 16 point calibration and compared the results. Fig. \ref{fig:subjects} shows some of the images from the dataset.

\begin{figure}[!htb]
\centering
\includegraphics[width=1\linewidth]{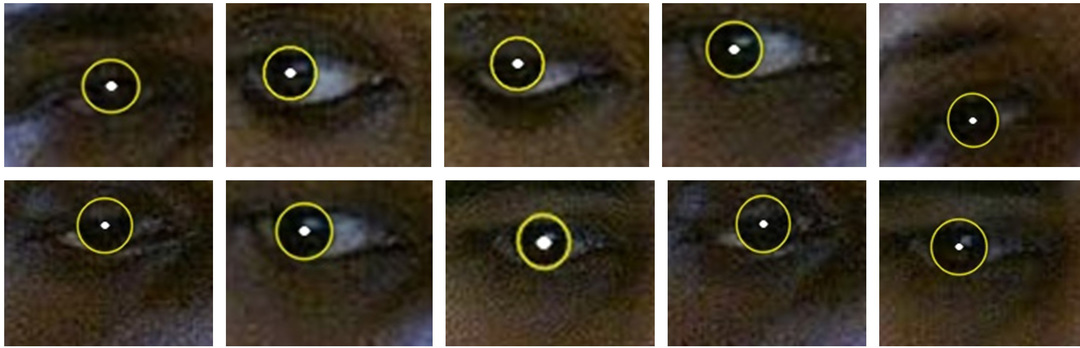}
\caption{ Sample images of detections in the custom dataset }
\label{fig:sampleyee}
\end{figure}

We have evaluated the IC localization accuracy on a subset of images in the in-house dataset. A subset of 1000 images was selected, and the IC localization accuracy was evaluated. Some of the sample detections are shown in Fig. \ref{fig:sampleyee}. The proposed approach obtained WEC accuracy of 90.2\% and 92.9\% for $e \le 0.05$  and  $e \le 0.10$ respectively.

 For $3 \times 3$ and $4 \times 4$ calibration grids, we have tested with polynomial regression and kernel space-based methods. The samples from the first session were used in the training stage. The parameters for regression were found from the training data. In testing stage, the samples in the second session were used to estimate the PoG. The mean position computed from the left and right eye PoG is used as the final gaze point. The error in the estimation is computed using the ground truth. The mean absolute error in visual angles in horizontal, vertical and overall accuracy is computed using the head distance from the screen.
 
 \begin{equation}
Accuracy = {\tan ^{ - 1}}\left( {{\raise0.7ex\hbox{${error}$} \!\mathord{\left/
 {\vphantom {{error} {head\,distance}}}\right.\kern-\nulldelimiterspace}
\!\lower0.7ex\hbox{${head\,distance}$}}} \right)
 \end{equation}

\begin{figure}[!htb]
\centering
\includegraphics[width=1\linewidth]{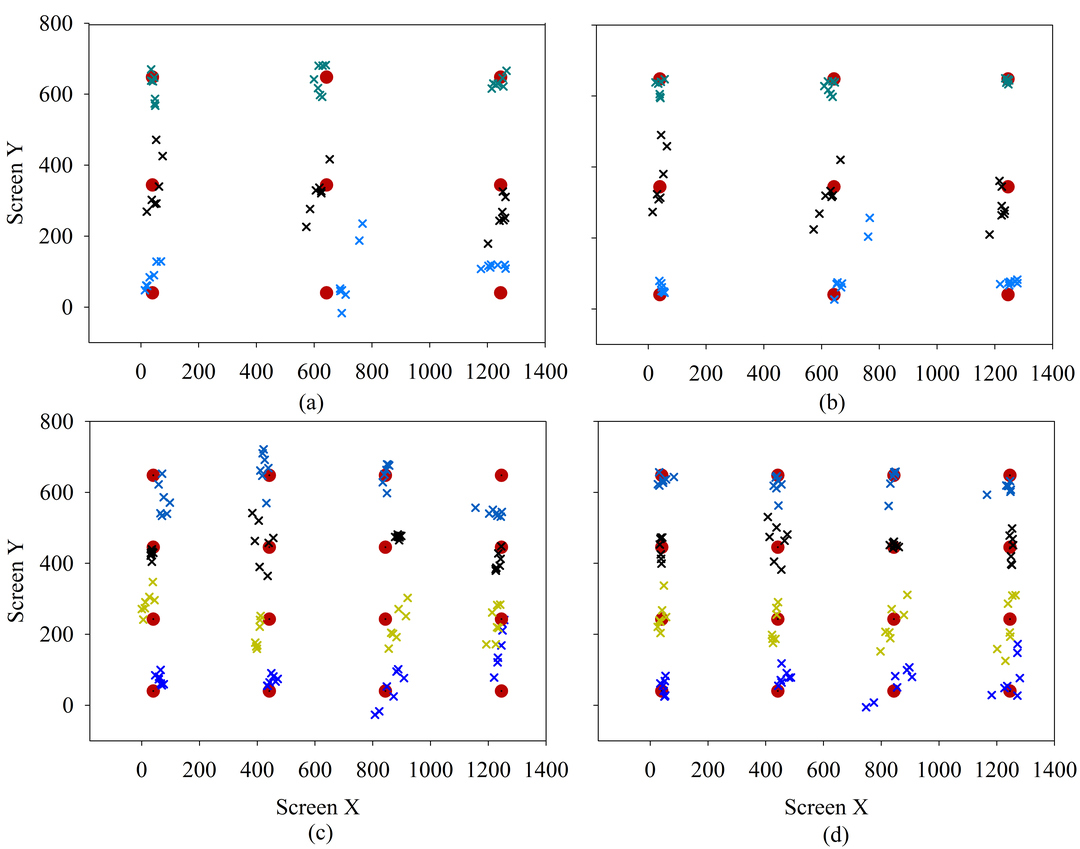}
\caption{ PoG estimates with 16 and 9 point calibration grids (a),(c) polynomial regression (b),(d) RBF kernel. Dots and crosses denote the target points and estimated gaze positions respectively. }
\label{fig:10}
\end{figure}

The average errors are high when the EC-IC vectors are computed on a frame by frame basis. We further computed the PoG using KF estimates which reduced the jitter significantly. The results with and without  KF on $3 \times 3$ and $4 \times 4$ calibration grids are tabulated in Table \ref{tab:3}. The qualitative results of gaze estimation stage are shown in Fig. \ref{fig:10}.

\begin{sidewaystable}
\centering
\caption{Gaze estimation error}
\label{tab:3}
\begin{tabular}{@{}cccccccc@{}}
\toprule
\multicolumn{1}{l}{}                                                        & \multicolumn{1}{l}{}                                          & \multicolumn{3}{l}{Raw gaze position}                                                                                                                                       & \multicolumn{3}{l}{With Kalman Filter}                                                                                                                                      \\ \midrule
Method                                                                      & \begin{tabular}[c]{@{}c@{}}Calibration\\  Points\end{tabular} & \begin{tabular}[c]{@{}c@{}}MAE\\ (degrees)\end{tabular} & \begin{tabular}[c]{@{}c@{}}MHE\\ (degrees)\end{tabular} & \begin{tabular}[c]{@{}c@{}}MVE\\ (degrees)\end{tabular} & \begin{tabular}[c]{@{}c@{}}MAE\\ (degrees)\end{tabular} & \begin{tabular}[c]{@{}c@{}}MHE\\ (degrees)\end{tabular} & \begin{tabular}[c]{@{}c@{}}MVE\\ (degrees)\end{tabular} \\ \midrule
\multirow{2}{*}{Polynomial}                                                 & 9                                                             & 3.46                                                    & 1.05                                                    & 2.36                                                    & 2.03                                                    & 0.67                                                    & 1.36                                                    \\
                                                                            & 16                                                            & 2.97                                                    & 0.98                                                    & 2.01                                                    & 1.95                                                    & 0.62                                                    & 1.32                                                    \\
\multirow{2}{*}{\begin{tabular}[c]{@{}c@{}}RBF Kernel\\ $\sigma_{k}=5$\end{tabular}} & 9                                                             & 2.81                                                    & 0.93                                                    & 1.91                                                    & 1.53                                                    & 0.47                                                    & 1.05                                                    \\
                                                                            & 16                                                            & 2.71                                                    & 0.87                                                    & 1.83                                                    & 1.33                                                    & 0.40                                                    & 0.91                                                    \\ \midrule
\multicolumn{8}{l}{MEA-Mean Absolute Error; MHE-Mean Horizontal Error; MVE-Mean Vertical Error}                                                                                                                                                                                                                                                                                                                                                                                                         \\ \bottomrule
\end{tabular}
\end{sidewaystable}

\subsubsection{Experiment for eye closure detection}
The eye regions obtained from the face detection stage are histogram equalized and resized to the size of $30 \times 30$. A data set of 4000 images containing 2000 samples for open and 2000 samples for closed eyes were formed from our dataset. HOG features were extracted from various pixel per cell windows and eight orientations. The extracted HOG features were used to train the SVM classifier. Ten times ten-fold cross validation was used to examine the accuracy of the trained classifiers. The proposed method achieves an average accuracy of 98.6\% with linear SVM. The results obtained are shown in Table \ref{tab:4}. 
\begin{table}[H]
\centering
\caption{Accuracy of eye closure detection}
\label{tab:4}
\begin{tabular}{@{}lll@{}}
\toprule
Pixel per cell in HOG & RBF Kernel SVM    & Linear SVM        \\ \midrule
2                     & 97.5\%  & 98.3\%) \\
4                     & 97.2\%  & 98.6\%) \\ \bottomrule
\end{tabular}
\end{table}
\subsection{Discussions}
The proposed method contains cascaded stages of many algorithms. The gaze estimation accuracy is a good proxy for the combined accuracy of all the cascaded stages. Face and IC are tracked using Kalman filters independently due to their distinct dynamics. The tracking based framework increases the robustness by reducing the effect of per-frame localization errors.

For successful eye tracking using webcams, the normalized error should be less than 0.05. The proposed algorithm performs better in realistic conditions for webcam-based gaze tracking. The accuracy of gaze estimation was evaluated with the proposed approach in both $3 \times 3$ and $4 \times 4$ calibration grids. RBF kernel-based non-parametric regression method was found to perform better than second order polynomial models. The average error rate obtained with the per-frame based detection was 2.71 degrees. The accuracy of the gaze tracking improved significantly by the use of KF, which uses the temporal information effectively to reduce the error rate to 1.33 degrees. 

The main strength of the proposed algorithm is in the two stage abstraction. We approximate the iris with an ellipse. However, time consuming search in a five parameter space is obviated using the two stage approach. One of the main contribution is the simplification of the ellipse fitting problem with a rather simple two stage scheme using appropriate constraints obtained from the face detection stage. Another advantage here is that small errors in the first stage can be refined in the second stage. Further, the addition of the tracking framework makes the algorithm more robust.

One of the advantages of the proposed algorithm is the low computational load. The eye detection, being a convolution-based method can be implemented in Fourier domain \cite{burrus1991dft} for faster computation. Multi-resolution convolution can be used to reduce the search space even further. The algorithm was implemented in a 2.5 GHz core 2 Duo desktop computer with 2 GB RAM. The C++ implementation using OpenCV library \cite{bradski2000opencv} (without multi-threading) was used for the evaluation experiments in Ubuntu 14.04 OS (32 bit) environment. It detects the face and eye corners in the first frame and tracks the eye corners over time. The temporal information is used to reduce the search space for face detection using a Kalman filter. The images were acquired using a 60 fps, $640 \times 480$ webcam. The online processing speed was limited only by the lower frame rate of the camera. The offline processing speed of the entire algorithm is well over 100 fps on the recorded video. This is suitable for normal desktop based implementation with 30 fps webcams. 

One of the main limitation of the approach is the deterioration of performance due to off-plane head rotations. A larger amount of off-plane rotation might result in the failure of the first stage of the iris center localization. However, for a smaller amount of off plane rotations, while the user is in front of the desktop, the second stage of the algorithm can refine the estimates without much reduction in overall accuracy. 
  
The proposed method can also be implemented in smart devices like mobile phones and tablets due to its low computational overhead. The low computational requirement makes it possible to extend the pose tracking with more complex 3D models. This could make the PoG estimation invariant to out of plane rotations as well.

\section{Summary}

This chapter describes an algorithm for a fast and accurate localization of iris center position in low-resolution grayscale images. A two-stage iris localization is carried out, and the filtered candidate iris boundary points are used to fit an ellipse using a gradient aware RANSAC algorithm. The proposed algorithm is compared with the state of the art methods and found to outperform edge-based methods in low-resolution images. The computational requirement of the algorithm is very less since it uses a convolution operator for iris center localization. We also propose and implement a gaze-tracking framework. Inner eye corners are used as the reference for calculating gaze vector. Kalman filter based tracking is used to estimate the gaze accurately in video. Further, ellipse parameters obtained from the algorithm can be combined with geometrical models for higher accuracy in gaze tracking. We have considered only in-plane rotations in this work. However, pose invariant models can be developed by using more computationally complex 3D models.

\end{onehalfspacing}

%% file: Chapter4/chapter_gazeir_v9.tex
\chapter{Pupil Center Localization Algorithm for Near Infrared Images}{Pupil Center Localization Algorithm for Near Infrared Images}
\graphicspath{{Chapter4/pics/}}
\begin{mdframed}[linecolor=grey!3,backgroundcolor=grey!3] 
This chapter describes a robust algorithm for pupil localization in Near Infrared (NIR) images obtained from a head mounted eye tracker. Most of the existing algorithms fail in localizing pupil center in challenging outdoor environments. The proposed algorithm uses both edges as well as intensity information along with a candidate filtering approach to identify the pupil center. The proposed method outperformed the state of the art algorithms while achieving real-time performance \cite{george2018escaf}.
\end{mdframed}
\vspace{5mm}

\begin{onehalfspacing}

\section{Introduction}

This chapter considers the development an algorithm for head mounted eye trackers. Head mounted eye trackers are useful in investigating human behavior in many practical dynamic tasks. In head-mounted cameras, accurate localization of pupil center is possible with the use of NIR illumination, which can give improved accuracy. Iris-sclera boundary is prominent in visible image based gaze tracking, where as in NIR lighting the pupil iris boundary is much more highlighted. Most of the head mounted eye trackers utilize dark pupil method for localizing the pupil. The challenges are different as compared to the algorithm developed for desktop environments. As head mounted eye trackers are supposed to function in challenging outdoor conditions, the pupil center detection algorithm should be robust against real-world conditions.

 Most of the existing algorithms for pupil localization perform well only under controlled conditions. With the advent of wearable head-mounted devices \cite{starner2013project},\cite{benko2015fovear},  eye tracking holds the potential to become a human-computer interaction channel. The point of gaze gives a lot of information regarding the attention of the user, which can be used to manipulate objects in virtual and augmented reality environments. Gaze tracking can also be used for foveated rendering \cite{guenter2012foveated}, which can reduce the computational load in image rendering. Head mounted trackers have more potential as they are not limited to desktop environments. However, the accuracy of gaze tracking degrades with real world conditions such as illumination variations, blur, partial-occlusions, reflections from external light sources, makeup, contact lenses,  and other sources of noise.  Therefore a robust pupil localization algorithm is necessary to deal with such real-world conditions. 

\begin{figure}[h]
\begin{center}
\includegraphics[width=1\linewidth]{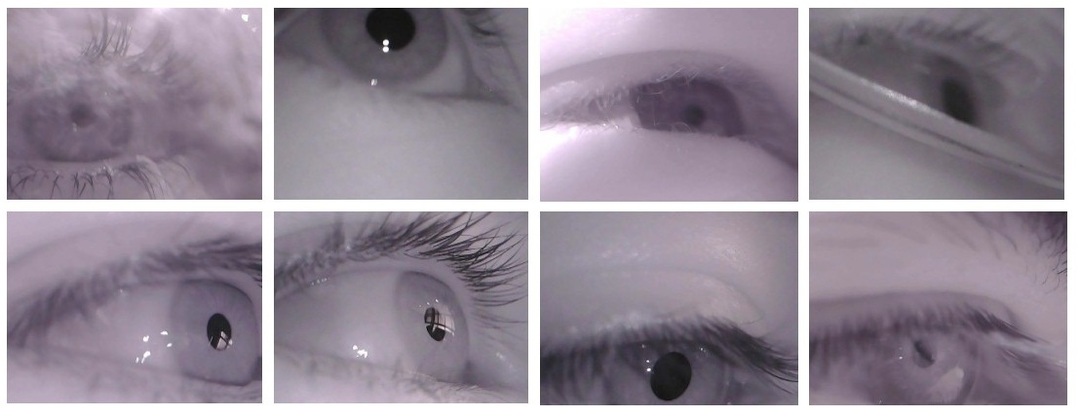}
\end{center}
\caption{Sample images from the LPW dataset}
\label{fig:samples_lpw}
\end{figure}

Pupil localization can be performed easily if the following conditions are true. 1) The gradients of pupil boundary are strong and can be detected by an edge detector 2) Pupil is the darkest region in the image. However, these assumptions may not hold well in uncontrolled environments. In this work, we develop a hybrid approach which can robustly detect PC in the images obtained from a head-mounted camera. The algorithm can be further extended to work with remote eye trackers by adding an eye detection stage. Dark pupil images obtained from a head-mounted camera are shown in Fig. \ref{fig:samples_lpw}.

A hybrid approach is proposed which uses intensity distribution as well as the edges for PC localization. Additionally, a simple tracking scheme is added  which increases the detection rate in real world conditions.
 
The main contributions from this chapter are listed below

\begin{itemize}

\item Proposes a novel framework for pupil center localization in NIR (Near Infrared) images.
\item The proposed approach combines multiple sources of information like intensity and edges for finding the pupil center
\item A multistage filtering of candidates is proposed which reduces the error in the final estimate using a scale space approach
\item A simple yet effective pupil tracking scheme is also included for enhanced detection rates and speed.

\end{itemize}

%
%
%
%

\section{Related works}

There is a significant amount of work related to pupil localization in NIR images. However, most of the works address the issues in controlled conditions. In this section, we review some of the recent works which discuss robust pupil center localization algorithms.

Li \textit{et al.} \cite{li2005starburst} proposed a hybrid algorithm for pupil center localization combining feature-based and model-based approaches. The algorithm detects and removes the corneal reflections in the preprocessing stage. They detected the pupil edges by tracing edges along a ray that extends from the best guess pupil center.This method was iteratively used to detect pupil boundary points. RANSAC-based ellipse fitting was carried out using the candidate boundary points. The ellipse parameters thus obtained were used as initial parameters for model based refinement. 

San Agustin \textit{et al.} \cite{san2010evaluation} proposed a method for eye tracking in low-cost webcam known as `ITU gaze tracker'.  In the first stage, the pupil image is thresholded. The pupil boundary points were detected and fitted to an ellipse using RANSAC-based approach. However, the performance of this approach degraded in real world conditions such as motion blur, glints and reflections. 

Swirski \textit{et al.} \cite{swirski2012robust} presented a pupil localization algorithm which can work in highly off-axis images. The coarse location of the pupil was obtained using Haar-like features. From the regions obtained, the pupil was thresholded using K-means clustering method.  A gradient aware RANSAC ellipse fitting was used for fitting the pupil boundary. The image aware nature of the algorithm made sure that the ellipse boundary lies along strong image edges. However, this method also fails during challenging conditions where external reflections affect the gradients.

Valenti and Gevers \cite{valenti2012accurate} proposed a new method for the localization of iris in visible images. The illumination invariant isophote curvature properties of the edge pixels were used in their approach. The curvatures of edge pixels vote to find the mean iso-center (MIC). This method can be used for IR images as well. However, this method fails to achieve satisfactory accuracy for off axis images. 

Kassner \textit{et al.} \cite{kassner2014pupil} proposed an open source framework for gaze tracking using head-mounted cameras along with an open source hardware design. In their approach, pupil candidates were detected using the center surround Haar-like features. A Canny edge detection stage was carried out followed by an edge filtering stage based on neighboring pixels. From the histogram, edges corresponding to spectral reflections were removed. After this edge pruning, remaining edges were labeled using connected components and split into sub-contours based on curvature continuity.  Ellipses were fitted to these candidate contours and evaluated for the inclusion of other contours.  Finally, a confidence score was calculated based on the ratio of supporting edge length and the circumference of the ellipse. If the confidence is less than a threshold, it reports that no ellipse has been found. One of the major disadvantages of this method is that it depends explicitly on edge detection. If the edge detection stage fails to detect pupil boundary due to motion blur or illumination, subsequent stages could fail. 

Javadi \textit{et al.} \cite{javadi2015set} proposed SET approach, in which a manual threshold was used for thresholding the image. The connected components were treated as pupil candidates, and their convex hull was found. An ellipse fitting stage was followed, and the ellipse which was closest to the circle shape was selected as the final pupil location. 

Fuhl \textit{et al.} \cite{fuhl2015excuse} presented a robust algorithm for PC localization in off-axis images. In the initial stage, the images were normalized, and the peak of the histogram was found. If the peak was found, the pipeline using edge filtering approach was used. The edges detected from Canny algorithm were filtered using morphological operations for removing lines and orthogonal edges. Straight lines were detected and removed using the distance of points to their centroids. The curve with lowest enclosed intensity was selected and fitted with an ellipse. In case peak was not found, the algorithm finds the coarse location of the pupil and refines it based on angular projection functions. The thresholded image was further refined and fitted with an ellipse. Fuhl \textit{et al.} \cite{fuhl2015else} further extended the work in \cite{fuhl2015excuse} improving it by the use of ellipse selection from Canny edges. After detection of edges using the Canny filter, edge segments were evaluated similarly as used in  ExCuSe \cite{fuhl2015excuse}. The segments were evaluated for various constraints including straightness, the inner intensity value and the best one was fitted. In case ellipse detection fails, likely locations of the pupil were found. The image was downscaled and convolved with a surface difference filter and a mean filter. The location of the maximum value in the multiplied result was used as the initial point for position refinement. The position is refined based on the analysis of surrounding pixels. The center of mass of the pixels with the new threshold is used as the updated pupil position. This location is evaluated with a validity check using the surface difference.

Most of the methods use multiple stages for PC localization. However, the rather simple assumptions of the pupil as the darkest region produces a lot of false detections. Head mounted trackers are supposed to work in real world applications, and they should perform robustly in real world conditions. To this end, ElSe approach proposed by Fuhl \textit{et al.} \cite{fuhl2015else} is robust. However, their method relies heavily on the Canny edge detector. Once the detector fails to detect the edges correctly due to glints or reflections, the second stage cannot recover if the edge detection fails.  Further, they do not leverage the temporal information.

Therefore, we propose a novel method which works even with challenging conditions such as glints, extreme angles, partial occlusion, image blur and illumination variations. Further, we introduce a simple yet effective pupil tracking scheme which makes the detection faster. Usage of the temporal information reduces the search space for pupil localization while decreasing the false positives.

\section{Proposed method}
Two basic approaches are commonly used for localizing pupil center in dark pupil images.  The first method uses the intensity distribution of the images. The pupil region is assumed as the darkest region in the image which can be well separated from the background. Some of the approaches use a manual threshold which is adjusted according to the imaging conditions. However, this method fails when other regions appear dark due to the shadows. Further, it may not be possible to find an exact threshold to segment out the pupil due to the varying external lighting and the glints.

Another approach uses edges of the image which can be found using Canny edge detector. The edges detected are filtered morphologically and using several other constraints. Candidate edge segments identified, and ellipse fitting is carried out. However, edge detection stage might fail due to external illumination, glints, and motion blur. In such conditions, the Canny edge detector fails in detecting the pupil boundaries resulting in the failure of subsequent stages.

\begin{figure}[t]
\begin{center}
\includegraphics[width=1\linewidth]{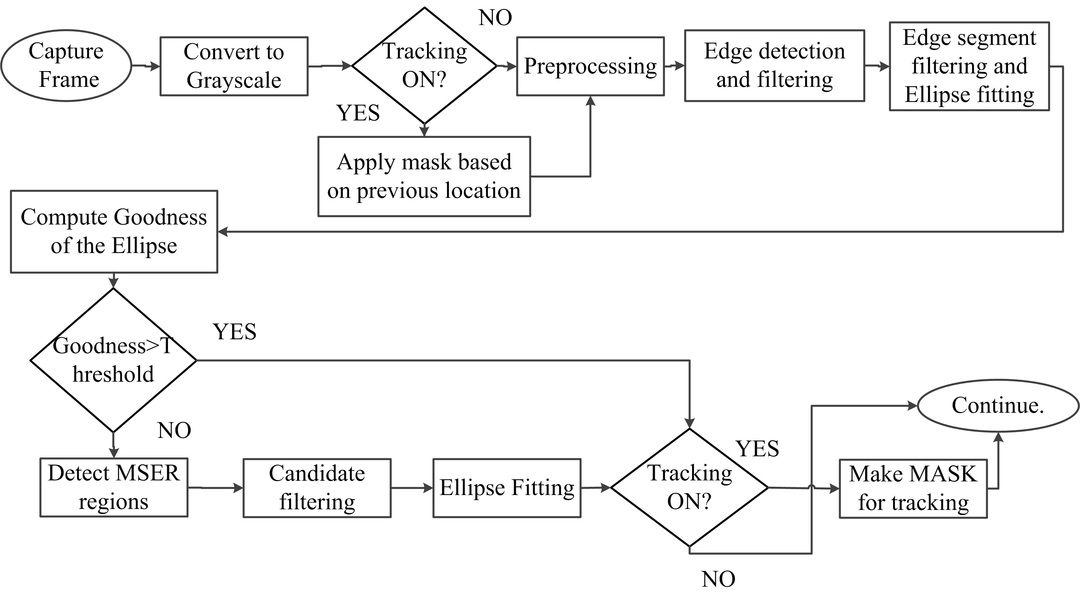}
\end{center}
\caption{Flowchart of the proposed approach}
\label{fig:flowchart}
\end{figure}

In our approach, we used intensity distribution, edges, image gradients and several other parameters to estimate the pupil center. The stages of the proposed method and the overall framework is shown in Fig. \ref{fig:flowchart}.

\subsection{Preprocessing and edge detection}
The native resolution of the camera used is 640 $\times$ 480. The images captured are downsampled by a factor of two to reduce the computational requirement. They are converted to grayscale and are scaled to the range of 0-255. 

After obtaining the normalized image, Canny edge detection algorithm is employed for detecting the edges. However, directly applying Canny algorithm over the eye image results in a lot of spurious edges. In our case, the task is to identify the pupil boundary. Since the region inside pupil is somewhat homogeneous, detection of false edges can be reduced by convolving the image with a Gaussian kernel (a $5 \times 5$ kernel was used). This is followed by a median filter stage, which again reduces the number of edges obtained. The Canny algorithm is applied on this preprocessed image, which results in edge segments which are more continuous. This further reduces the computation in the subsequent stages.

\subsection{Edge selection and candidate filtering}

Once the edges are obtained,  border following algorithm \cite{suzuki1985topological} is used to separate the edge segments. Edge segments with length more than ten are selected for further analysis. Polygonal approximations of the edge segments are found using Douglas-Peucker algorithm \cite{douglas1973algorithms}. Now for each segment, the curvature is computed \cite{kassner2014pupil} and the segments are split into subsegments if curvature inflections are found. The candidate edge segments are evaluated for the suitability of being a pupil boundary. For this, we introduce new criteria based on ellipse fitting, each edge segment is fitted with an ellipse using a least square approach \cite{Fitzgibbon95abuyer's}.  Candidate edge segments are pruned based on the area and the aspect ratio of the fitted ellipses. Edge segments which are too small or too large are rejected at this stage. The median intensity of the inner region of candidate edge segments are found, and candidates with inner intensity less than an empirically determined threshold are selected for further analysis. We use a new method for candidate edge filtering and merging. The edge segments are sorted based on the median of the inner intensities. In the next stage, the edge candidates belonging to the pupil boundary are merged. A combinatorial search is carried out to determine the whether two candidates belong to the same ellipse. Two parameters are considered in this search, i.e. 1) the similarity of median grayscale value enclosed by the segment, and 2) the Euclidean distance between the centers of the fitted ellipses. Edge segments are merged if these two criteria are satisfied. The combined boundary is fitted with an ellipse, and a goodness parameter is computed. The median difference of grayscale values from the inner and outer, along with the edge support is also computed. We use the goodness of fit parameter proposed in chapter 2. The center of the ellipse is returned as the pupil center if the goodness parameter is greater than an empirically selected threshold.

\begin{figure}[h]
\begin{center}
\includegraphics[width=1\linewidth]{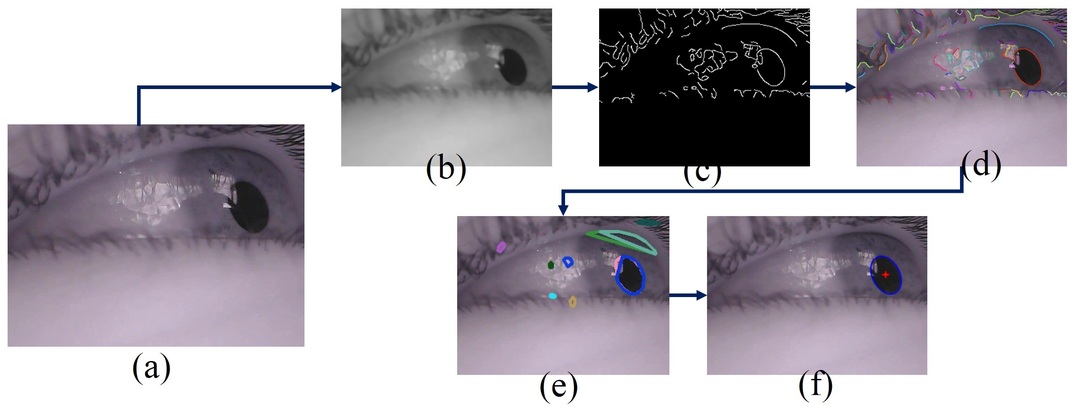}
\end{center}
\caption{Edge based ellipse fitting, a) The original color image captured, b) Downsampled and filtered grayscale image, c) Canny edges, d) Edge segments, e) Candidate edges, f) Fitted ellipse after contour merging }
\label{fig:process_edge}
\end{figure}

 Edge based ellipse fitting can fail when the edges in the image are weak. Motion blur, low contrast, external lighting and noise can also result in the failure of the edge detection stage. If the edge based fitting fails, we use grayscale intensity distribution to identify the pupil center candidates.

\subsection{Candidate detection with MSER}

\begin{figure}[h]
\begin{center}
\includegraphics[width=1\linewidth]{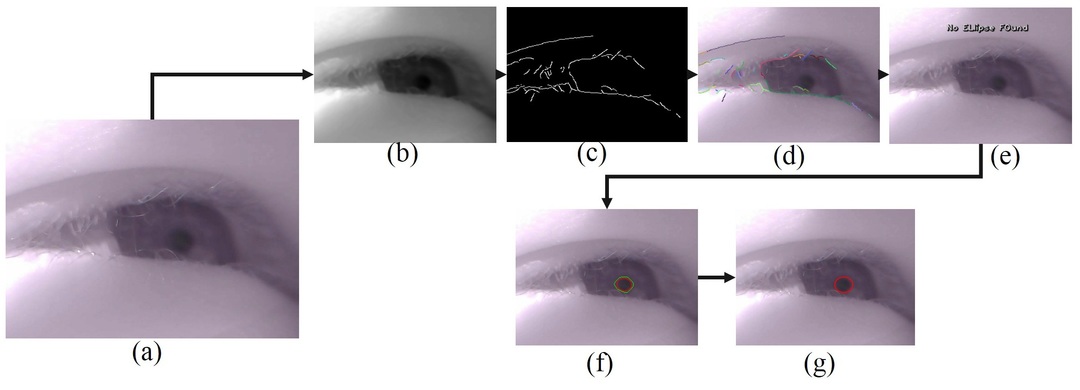}
\end{center}
\vspace{-1.5em}
\caption{Intensity based ellipse fitting, a) The original color image captured, b) Downsampled and filtered grayscale image, c) Detected edges, d) Candidate edge segments, e) Failure of edge based ellipse fitting stage, f) Pruned MSER regions found from the scale space implementation, g) Ellipse fitting corresponding to best pupil candidate }
\label{fig:process_mser}
\end{figure}

If the edge-based fitting fails, the algorithm tries to detect the pupil location using the grayscale intensity as shown in Fig. \ref{fig:process_mser}. We apply a candidate filtering approach for obtaining the pupil center. In the first stage, different candidate regions are identified using a variant of maximally stable extremal regions (MSER) proposed by Matas \textit{et al.} \cite{matas2004robust}. In the second stage, the candidate regions are evaluated for ellipse fitting criterion and the best candidate is selected as the pupil center.

Component tree of an image is the set containing all the connected components of different thresholds, ordered by inclusion. The maximally stable extremal regions can be found from the component tree of a grayscale image. We start with the lowest value in the image grayscale values. Connected components with different thresholds are found out. The region corresponding to a particular threshold is said to be stable if the area of the thresholded regions remains almost stable over a large range of thresholds.  The local maxima of these regions are identified as the maximally stable extremal regions. More details about the fast implementation of MSER algorithm can be found in \cite{nister2008linear}. In our approach, we use three constraints in detecting the MSERs. The minimum and maximum areas of the pupil are assumed to be known, which are used as constraints. The maximum inner intensity of the pupil region is also known. The component tree needs to be computed only up to this level for finding the candidate regions. MSER is known to be sensitive to image blur.  Scale-space pyramid based implementation \cite{forssen2007shape} is used to alleviate the issue of image blur.

Once the MSER regions corresponding to the constraints are obtained, a candidate filtering approach is performed to identify the best pupil candidate. The region boundary contours are fitted with ellipses, and the ratio of the major axis and the minor axis is found. Candidate regions with the ratio less than a 
predefined threshold are identified for further analysis. The goodness parameter is computed, and the candidate with the highest goodness parameter is used as the pupil ellipse.


\subsection{Tracking framework}

The time taken for processing can be reduced significantly by using temporal information. Tracking algorithms like Kalman filter or Particle filter usually assume a motion model for the dynamics of the object to be tracked. However, the dynamics of eye movements involve several subclasses like fixations, saccades, vergence, smooth pursuits, etc. Having different dynamics makes it difficult to implement tracking in practical scenarios. However, based on eye physiology and sampling frequency of the image acquisition system, a rough estimate of the maximum possible change in the position of the pupil center between successive frames can be computed. This information can be used to constrain the search space without the loss of accuracy. In our approach, we have used the previous location of PC to obtain the search region for the current frame. A rectangular region is selected around the last position of the pupil. Search space for PC localization is limited to this area only. However, selection of this mask depends on the confidence of PC localization, the mask from the previous frame is used only when the goodness parameter for ellipse fitting is more than an empirically found threshold. If the ellipse fitting stage in the previous frame does not result in a high Goodness factor, the entire image is searched for localizing PC. This simple approach achieves better frame rates while removing many false detections. The main advantage of this method is that it gets rid of redundant computations and limits the processing to more promising areas using the temporal information.

\subsection{Comparison with state of the art method}

The closest method related to the proposed approach is ElSe \cite{fuhl2015else} proposed by Fuhl \textit{et al.}, as they also use two different pipelines based on the imaging conditions. In the first stage, they have used Canny edge detector followed by complex algorithmic and morphological filtering, whereas in the proposed approach most spurious edges are rejected by downsampling followed by Gaussian filtering. We have introduced new criteria for edge filtering based on the geometric distance and the inner intensity difference of the fitted ellipses. In the event of failure of the edge based stage, the second stage is performed which uses scale space variant of MSER algorithm followed by the proposed candidate filtering to identify the best pupil candidate. Further, the proposed method introduces a tracking approach which reduces the search space based on the confidence levels obtained from the ellipse fitting stage. 

\section{Experiments}

\subsection{Labeled Pupils in the Wild database}
We have used labeled pupils in the wild (LPW) database \cite{tonsen2016labelled} for the algorithm evaluation since the number of images is large and it contains images recorded in real world conditions.  LPW dataset contains 66 high-quality videos of eye regions from 22 subjects, including samples from people of different ethnicities, indoor and outdoor illumination variations in different gaze directions. It also contains images of participants wearing glasses, contact lenses and makeup. Each video in the database contains around 2000 images of resolution $640 \times 480$ recorded at a frame rate of 95 fps. The dataset contains a total of 130,856  images which is much larger than any of the existing datasets. Ground truth pupil locations are also available with the dataset.


\subsection{Evaluation of the algorithm}

We have evaluated the proposed algorithm in 66 videos provided in the dataset. The pixel error in each frame was computed from the ground truth available with the dataset. The comparison with state of the art is made based on the data as in \cite{fuhl2016pupil}.

The result obtained from the proposed algorithm has been compared with six other state of the art methods available in the literature. We have compared the results with Starburst \cite{li2005starburst}, Swirski \cite{swirski2012robust}, SET \cite{javadi2015set}, Pupil Labs \cite{kassner2014pupil}, ExCuSE \cite{fuhl2015excuse}, and ElSE \cite{fuhl2015else}.

\begin{figure}[h]
\begin{center}
\includegraphics[width=1\linewidth]{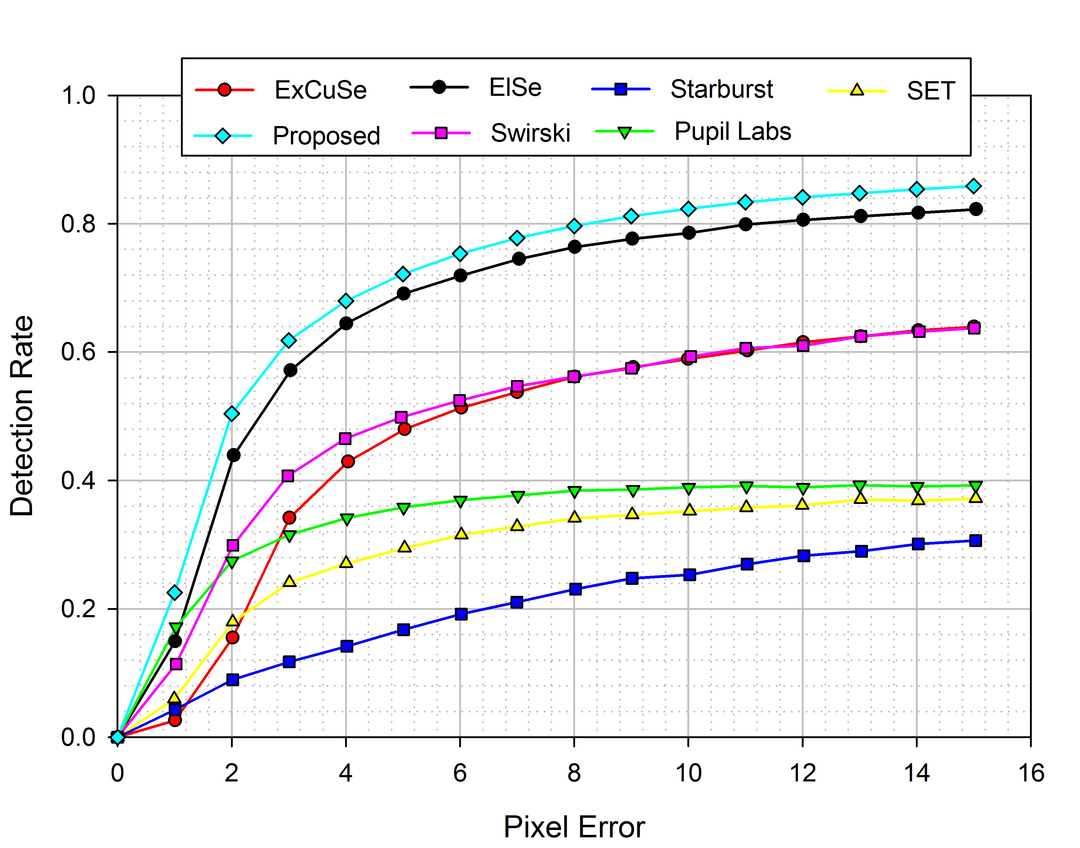}
\end{center}
\vspace{-1.5em}
\caption{Detection rates of the algorithms in LPW dataset ( ElSe, ExCuSe, Pupil Labs, SET, Starburst, Swirski and Proposed).}
\label{fig:lpwfull}
\end{figure}

The results obtained, and the comparison with state of the art are shown in Fig. \ref{fig:lpwfull}. The proposed approach outperforms all the state of the art methods. Comparative results for an error of five pixels is provided in Table. \ref{tab:fivepixel}. The proposed method obtains best overall accuracy.

\begin{sidewaystable}
\centering
\caption{Comparison of detection rates for an error of 5 pixels}
\label{tab:fivepixel}
\begin{tabular}{@{}llllllll@{}}
\toprule
Data set\textsuperscript{*} & \multicolumn{1}{c}{SET (\%)} & \multicolumn{1}{c}{Starburst (\%)} & \multicolumn{1}{c}{Swirski (\%)} & \multicolumn{1}{c}{ExCuSe (\%)} & \multicolumn{1}{c}{ElSe (\%)} & \multicolumn{1}{c}{Pupil Labs (\%)} & \multicolumn{1}{c}{Proposed (\%)} \\ \midrule
1        & 56.86                        & 39.79                              & 84.48                            & 63.53                           & 87.95                         & 65.58                               & \textbf{92.05}                    \\
2        & 48.68                        & 19.70                              & 41.58                            & 29.90                           & 69.87                         & 24.72                               & \textbf{82.83}                    \\
3        & 27.55                        & 6.75                               & 31.43                            & 34.83                           & \textbf{57.50}                & 21.82                               & 51.58                             \\
4        & 7.70                         & 9.27                               & 16.87                            & 25.38                           & 37.42                         & 14.53                               & \textbf{50.92}                    \\
5        & 6.75                         & 0.00                               & 8.38                             & 19.08                           & \textbf{22.95}                & 13.15                               & 19.75                             \\
6        & 11.10                        & 13.30                              & 63.48                            & 53.44                           & 84.10                         & 38.72                               & \textbf{87.91}                    \\
7        & 43.55                        & 7.65                               & 66.17                            & 66.48                           & 73.60                         & 68.43                               & \textbf{85.93}                    \\
8        & 42.17                        & 34.32                              & 77.68                            & 75.32                           & 81.00                         & 64.22                               & \textbf{87.25}                    \\
9        & 35.65                        & 30.90                              & 56.40                            & 60.42                           & \textbf{61.97}                & 45.20                               & 61.53                             \\
10       & 10.42                        & 3.65                               & 71.23                            & 59.00                           & 72.65                         & 41.62                               & \textbf{82.93}                    \\
11       & 31.07                        & 18.10                              & 31.58                            & 49.52                           & 71.48                         & 9.45                                & \textbf{73.68}                   \\
12       & 54.92                        & 24.10                              & 71.82                            & 72.58                           & \textbf{89.73}                & 49.58                               & 89.32                             \\
13       & 14.75                        & 16.52                              & 27.03                            & 45.04                           & \textbf{51.51}                & 16.68                               & 45.43                             \\
14       & 30.25                        & 23.50                              & 76.07                            & 58.60                           & 70.50                         & 57.52                               & \textbf{79.40}                    \\
15       & 27.17                        & 8.15                               & 37.80                            & 44.83                           & 53.95                         & 43.78                               & \textbf{58.48}                    \\
16       & 23.24                        & 17.24                              & 74.11                            & 72.73                           & 82.13                         & 82.01                               & \textbf{85.67}                    \\
17       & 20.90                        & 2.42                               & 68.88                            & 42.10                           & 72.97                         & 47.52                               & \textbf{73.15}                    \\
18       & 50.67                        & 33.48                              & 61.18                            & 66.25                           & 78.57                         & 48.73                               & \textbf{82.48}                    \\
19       & 11.97                        & 3.45                               & 24.87                            & 21.88                           & 54.05                         & 2.60                                & \textbf{76.72}                    \\
20       & 19.83                        & 16.58                              & 41.40                            & 11.72                           & \textbf{83.35}                & 0.98                                & 76.38                             \\
21       & 30.43                        & 25.63                              & 55.90                            & 47.45                           & 88.92                         & 28.18                               & \textbf{92.05}                    \\
22       & 41.60                        & 11.85                              & 6.48                             & 31.23                           & 70.00                         & 0.83                                & \textbf{75.65}                    \\ \hline
Overall  & 29.42                        & 16.65                              & 49.76                            & 47.79                           & 68.92                         & 35.72                               & \textbf{73.23}                   \\
\bottomrule
\multicolumn{8}{l}{\textsuperscript{*}\footnotesize{Data taken from \cite{fuhl2016pupil} for comparison.}}\\

\multicolumn{8}{l}{\footnotesize{Best results obtained in each dataset is shown in bold scripts.}}

\end{tabular}
\end{sidewaystable}

\begin{figure}[h]
\begin{center}
\includegraphics[width=0.8\linewidth]{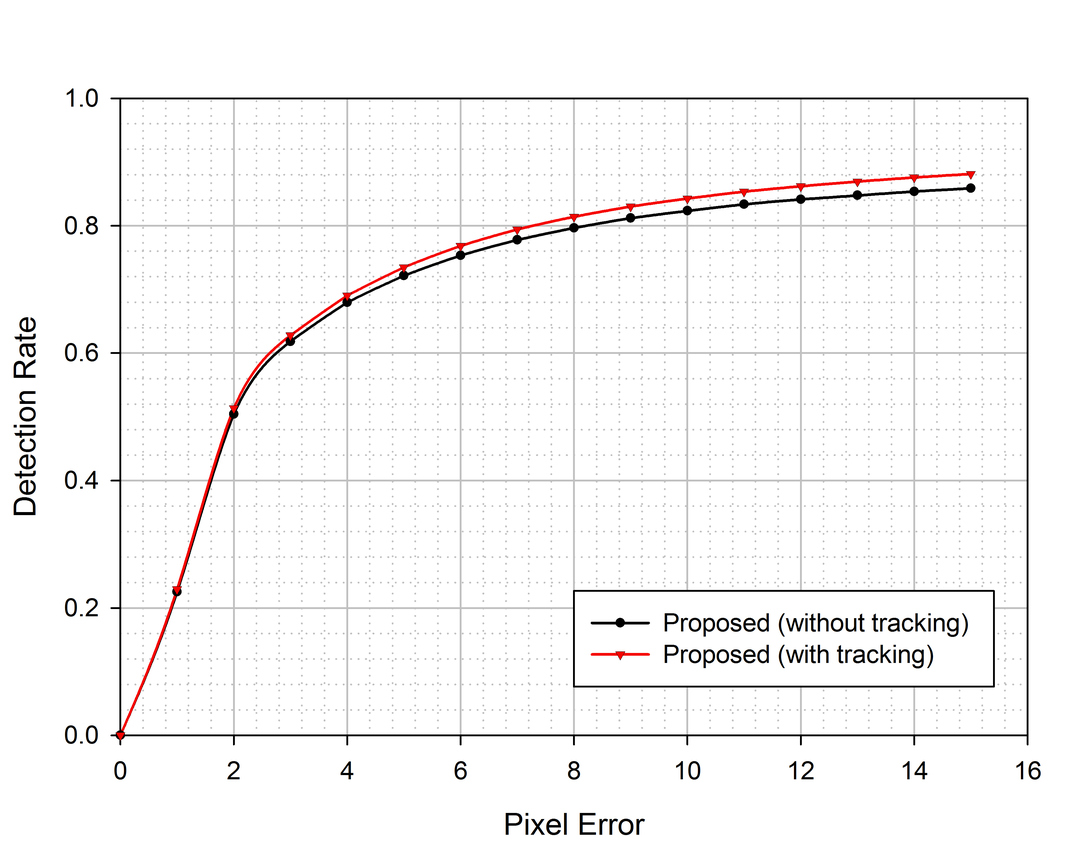}
\end{center}
\caption{Detection rates with and without tracking.}
\label{fig:tracking}
\end{figure}

The results obtained with and without tracking are shown in Fig. \ref{fig:tracking}. The addition of tracking decreased the processing load without any decrease in accuracy. The results obtained with tracking are slightly better than individual frame based detections. This can be attributed to the reduction in false detections due to the masking used in the tracking approach.

\begin{figure}[H]
\begin{center}
\includegraphics[width=1\linewidth]{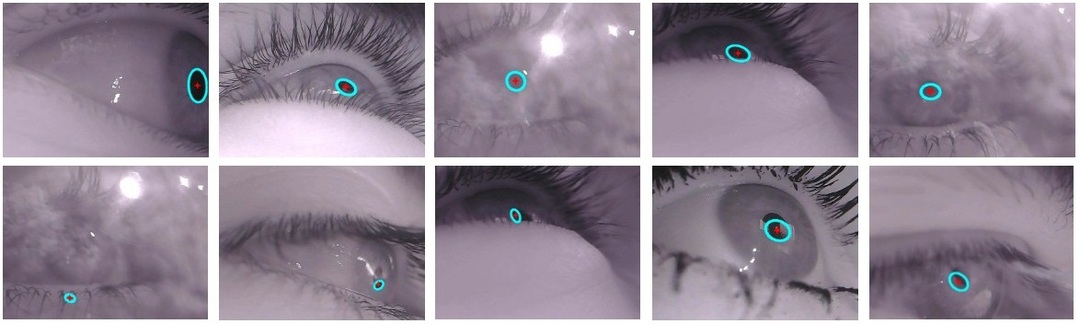}
\end{center}
\vspace{-1.5em}
\caption{Sample results from the detections, the first row shows the successful detection and second row shows detection failures.}
\vspace{-.5em}
\label{fig:success_failure}
\end{figure}

\begin{figure}[H]
\begin{center}
\includegraphics[width=1\linewidth]{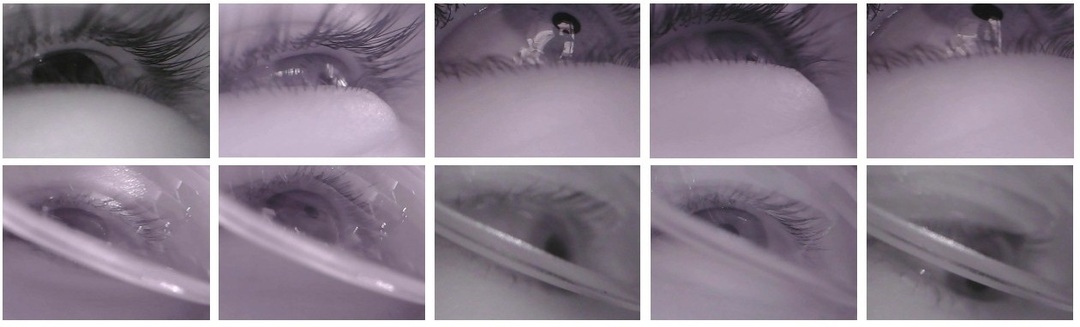}
\end{center}
\vspace{-1.5em}
\caption{Some examples of the challenging images from datasets 4 and 5}
\vspace{-1.5em}
\label{fig:challenging}
\end{figure}

Some of the successful detections and failures are shown in Fig. \ref{fig:success_failure}. Fig. \ref{fig:challenging} shows some of the challenging images from the LPW dataset.

%
%

\section{Discussions}
The overall performance of the algorithm in the entire dataset is shown in Fig. \ref{fig:lpwfull}. The proposed algorithm outperformed all the state of the algorithms.  

The algorithm was designed to be robust against real world conditions. Detection using one particular feature may not work in all practical usage scenarios. The proposed algorithm switches to either edge based method or intensity based method depending upon the image conditions. One significant advantage here is that, even if the first edge based stage of the algorithm fails due to some reflections, the second stage can identify the pupil (though more computation is required). Further, the tracking approach reduces false detection rate at the same time reduces computational load as the search space is considerably less. Most of the algorithms are designed to maximize the per frame based detection rates. Here, we have added the tracking framework which directly extends the algorithm for video. Results with and without tracking are shown in Fig. \ref{fig:tracking}. The tracking scheme achieved slightly better results with a better runtime performance.

\subsection{Execution time }

The algorithms were implemented on a desktop computer with 64 bit Ubuntu 13.10 Operating system having  3.33 GHz core i5 processor, and  8GB RAM. Implementation with unoptimized python code obtained an average processing time of 14.28 ms/frame without tracking and 9.90 ms/frame with tracking. There is a scope for improving the processing time by code optimization.
\subsection{Limitations}

The algorithm detects the pupil centers accurately when the edges are correctly detected. Intensity based candidate filtering approach detects the pupil when the edge-based approach fails. However, the algorithm fails when the pupil is occluded as shown Fig. \ref{fig:challenging} (dataset 5). The edges are not properly detected because of blur. Intensity based approach could also fail since the candidates are occluded. This reduces the goodness of regions obtained. Another failure case can occur when the image contrast is poor, and the surrounding regions have low contrast and reflections.  A machine learning based approach can be used to compensate for the detection-failures in such challenging cases.


\section{Summary}

In this chapter, we have presented a framework for pupil center localization in dark pupil images. The algorithm works in images captured with a head-mounted eye tracker using dark pupil method. The primary objective of the work was to develop an accurate algorithm which would work in real world conditions. The algorithm takes care of both intensity and edge information to estimate the pupil center accurately. A candidate filtering approach is chosen which maximizes a goodness function returning the best possible pupil candidate. A simple tracking method has also been used. It reduces the computations required without compromising on the accuracy. The proposed approach has been evaluated on LPW dataset and found to outperform all the state of the art methods. The python-based implementation achieves frame rates close to 100, which can be improved with an optimized implementation in C/C++. The high frame rates obtained can be useful in adding additional post processing and more computationally sophisticated tracking algorithms for improving the accuracy even further. Online identification of the eye movement types can be helpful in using appropriate model for tracking eye movements during different movement types. 

\end{onehalfspacing}

%% file: Chapter5/chapter_eac_v6.tex
\chapter{Gaze Direction Classification Using Convolutional Neural Network}{Eye Gaze Direction Classification Using Convolutional Neural Network}
\graphicspath{{Chapter5/pics/}}
\begin{mdframed}[linecolor=grey!3,backgroundcolor=grey!3] 

This chapter presents a real-time framework for the classification of eye gaze direction. A convolutional neural network is employed in this work for the classification of eye gaze direction. The proposed gaze direction obtained can be readily used for various HCI applications since it does not require a person dependant calibration stage.
\end{mdframed}
\vspace{5mm}

\begin{onehalfspacing}

\section{Introduction}

Human eyes provide rich information about human cognitive processes and emotions. The gaze  patterns of eyes also contains information about fatigue \cite{di2012towards}, diseases \cite{terao2011initiation}, etc. Estimation of gaze direction can be useful in various domains, including psychology, disease diagnosis, and human computer interaction. Gaze direction changes can be used as an interaction channel virtual environments.  They can also be used in applications like gaze based gestures, eye contact identification, eye based typing, etc.

 Most of the existing eye trackers require a cumbersome calibration procedure.  A person independent gaze classification system can be useful in scenarios where obtaining calibration data is difficult, such, gaze tracking in public displays and experiments with children. 

In this chapter, we present a real-time framework which can detect eye gaze direction using off-the-shelf, low-cost cameras in desktops and other smart devices. We treat the gaze direction classification as a multi-class classification problem, avoiding the need for calibration.

\begin{figure}[h]
\begin{center}
\includegraphics[width=1\linewidth]{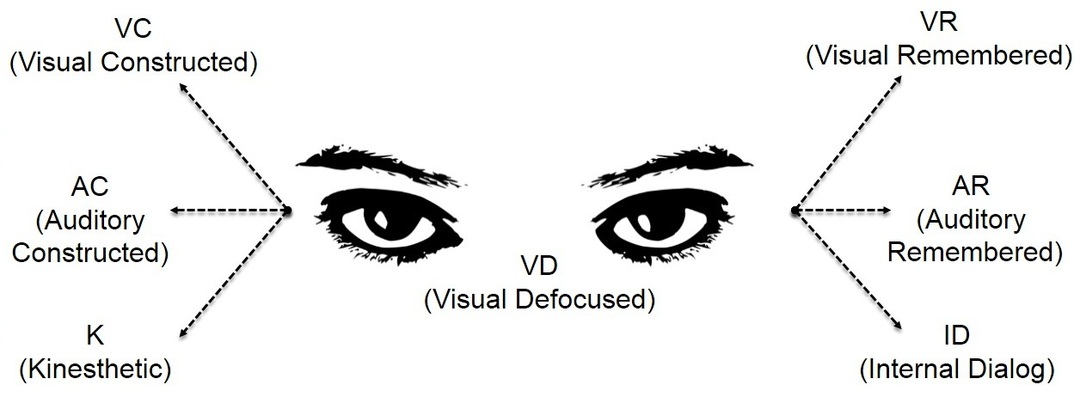}
\end{center}
\caption{Different EACs in NLP theory}
\label{fig:eac}
\end{figure}

\subsection{Application in finding  Eye Accessing Cues}

The patterns in which the eyes move when humans access their memories is known as eye accessing cues (EAC).  The neuro-linguistic programming (NLP)  EAC theory suggests \cite{bandlerfrogs} that there is a correlation between eye-movements and cognitive processing while accessing experiences.  EAC theory suggests that the meaning of non-visual gaze directions may be directly related to the internal mental processes. These movements are reported to be related to the neural pathways which deal with memory and sensory information. The direction of the iris in the socket can give information regarding various cognitive processes. Each direction of non-visual gaze is associated with different cognitive processes. The meanings of the various EACs are shown in Fig. \ref{fig:eac}. More details about EAC model can be found from  \cite{bandlerfrogs}. Even though the EAC theory is not 100 \% accurate, recent studies \cite{sturt2012neurolinguistic},\cite{vranceanu2013computer} have found correlation which incites further research in the field. A critical review of EAC method can be found in \cite{diamantopoulos2009critical}.

The eye directions obtained from the proposed framework can be used to find the Eye Accessing Cues (EAC) and thereby infer the user's cognitive process. The information obtained can be useful in the analysis of interrogation videos, human-computer interaction, information retrieval, etc. Identifying the affective and cognitive states of humans can make the interaction between computers and humans more natural. The knowledge of mental processes can help computer systems to interact intelligently with humans. 

\section{Related works}

There are many works related to gaze tracking in desktop environments, an excellent review of the methods can be found in \cite{hansen2010eye}. In this section, we limit the discussion to the recent state of the art works related to eye gaze direction estimation.

Vr{\^a}nceanu \textit{et al}. proposed a method \cite{vranceanu2011fast} for automatic classification of eye gaze direction using the information from color space. The relative position of iris and sclera in the eye bounding box is used to classify the eye gaze direction. Vr{\^a}nceanu \textit{et al}. proposed another method \cite{vranceanu2013automatic} for finding gaze direction using iris center detection and facial landmark detection. They used isophote curvature based method for iris center localization. The relative position of iris center is used with the fiducial points for a better estimate of eye gaze direction. In \cite{radlak2014gaze} Radlak \textit{et al.} presented a method for gaze direction estimation in static images. They used an ellipse detector with a support vector based verifier. The eye bounding box is obtained using the hybrid projection functions \cite{song2013literature}. Finally, the gaze direction is classified using support vector machine (SVM) and Random Forests. Recently Vr{\^a}nceanu \textit{et al}. \cite{vranceanu2015gaze} proposed another approach for eye direction detection using component separation. Iris, sclera, and skin are segmented and the features obtained are used in a machine learning framework for classifying the eye gaze direction. Zhang \textit{et al}. \cite{zhang2015appearance} applied convolutional neural network (CNN) for gaze estimation.  They combined the data from face pose estimator and eye region using a CNN model. They have trained a regression model in the output layer.

In most of the related works, the general framework uses three cascaded stages, i.e. face detection, eye localization, and classification. The localization or classification errors in any of the cascaded stages will result in the reduction of overall accuracy. The computational complexity of the methods is another bottleneck. In this work, we aim at increasing the accuracy of eye gaze direction classification. The developed algorithm is robust against noise, blur, and localization errors. The computational load is less in the testing phase, and the proposed algorithm achieves an average 24 fps in a Python-based implementation without using graphical processing units (GPU).

\section{Proposed algorithm}
\begin{figure}[h]
\begin{center}
\includegraphics[width=1\linewidth]{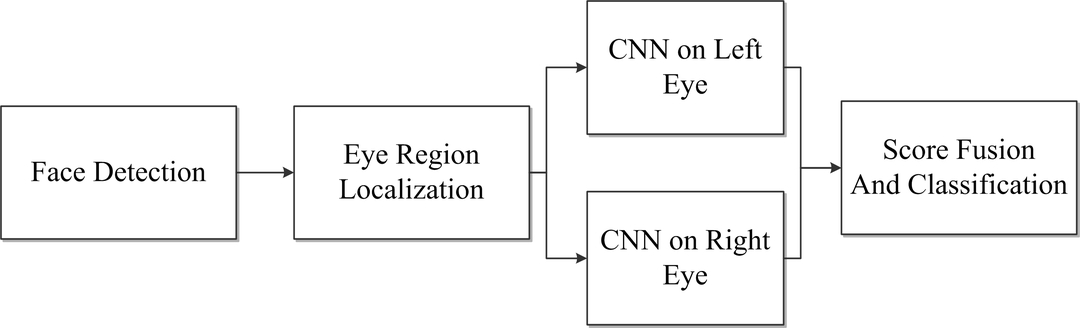}
\end{center}
\caption{Schematic of the overall framework}
\label{fig:overall}
\end{figure}

The overall framework is shown in Fig. \ref{fig:overall}. Different stages of the algorithm are described below.

\subsection{Face detection and eye region localization}
\begin{figure}[h]
\begin{center}
\includegraphics[width=1\linewidth]{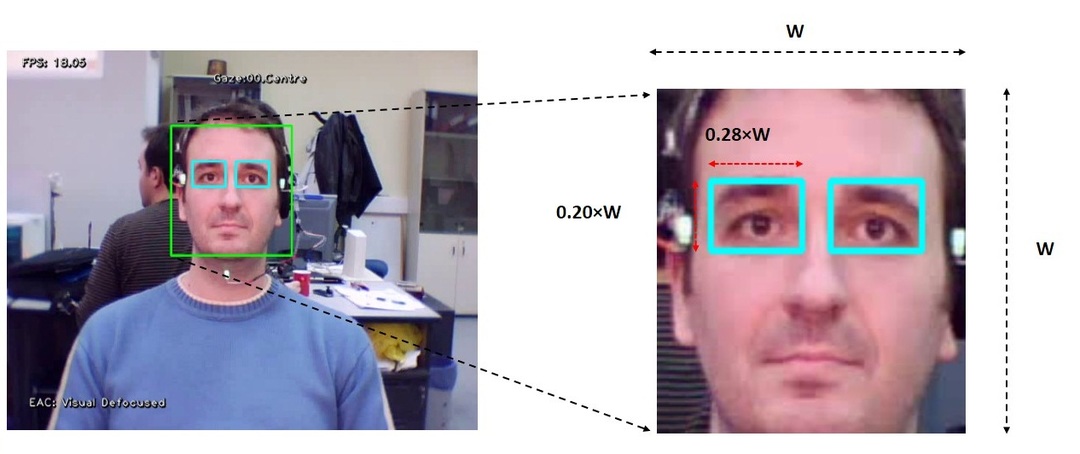}
\end{center}
\caption{Eye region localization using geometric approach (ROI)}
\label{fig:eyeloc}
\end{figure}

The first stage in the algorithm is face detection. The framework described in Chapter 2 is used for face detection. Once the face region is localized, next stage is to obtain the eye region. We have used two different methods for obtaining the eye region. In the first method, the eye region for classification is obtained geometrically from the face bounding box returned by the face detector (ROI). The dimension of the eye region is shown on an image from HPEG database \cite{asteriadis2009natural} in Fig. \ref{fig:eyeloc}. The eye regions obtained are re-scaled to a resolution of $42\times50$ for the subsequent stages (ROI). In the second method, we used a facial landmark detector to find the eye corners and other fiducial points.

\begin{figure}[h]
\begin{center}
\includegraphics[width=1\linewidth]{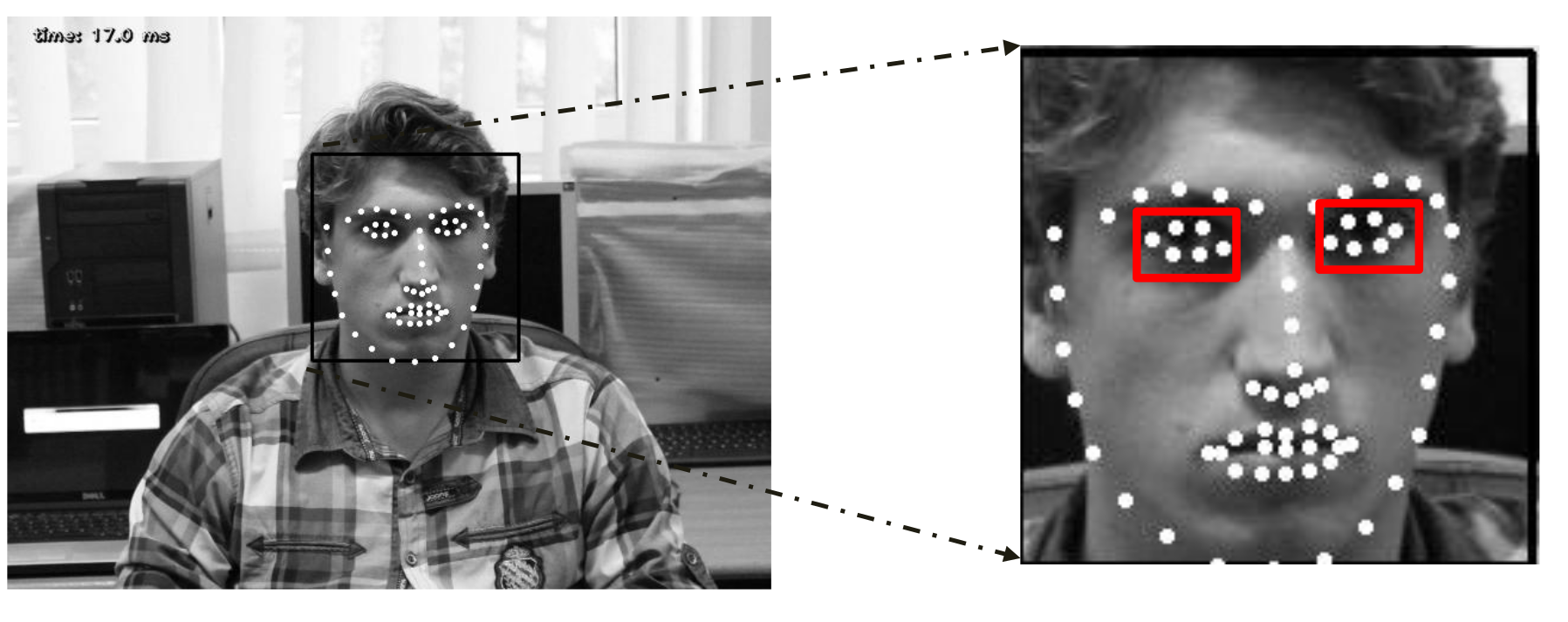}
\end{center}
\caption{Eye region localization using ERT approach}
\label{fig:eyelocclm}
\end{figure}
\subsubsection{Facial landmark localization}
Localization of facial landmarks helps in constraining the eye region for classification. Ensemble of randomized tree approach (ERT) \cite{kazemi2014one} approach is used for localizing facial landmarks. The face bounding box obtained from the preceding stage is used as the input to the algorithm. The locations of the facial landmarks are regressed using a sparse subset of pixels from the face region. The algorithm is very fast and works even with partial labels. The details of the algorithm can be found in \cite{kazemi2014one}. Figure. \ref{fig:eyelocclm} represents the selection of eye patch from the facial landmarks around the eye region. The inner and outer eye corners along with the upper and lower eyelid points are used to define a rectangular region. The rectangle area including all the eyelid boundary points returned from the landmark detector is applied to select the eye patch. The selected patch is aligned using the inner and outer eye corners which is used in the subsequent classification stage.

\subsection{Eye gaze direction classification}

The eye region obtained from the previous stage is used in a multiclass classification framework for
predicting the eye gaze direction. Convolutional neural network (CNN) is used for the classification. 

\subsubsection{Convolutional Neural Network (CNN)}

The convolutional neural network represents a type of feed-forward neural network which can be used for a variety of machine learning tasks. Krizhevsky \textit{at al}. \cite{krizhevsky2012imagenet} used a large CNN model for the classification of images in Imagenet database. Even though the training time is huge, the accuracy and robustness of CNNs are better than most of the standard machine learning algorithms. In our approach, we have used a CNN model with three convolution stages.

We follow the popular LeNet \cite{lecun1998gradient} architecture in this paper. The LeNet architecture consists of convolutional layers followed by nonlinearity and pooling layers. Usually, there are multiple stages depending on the complexity of the problem. The first convolution layer acts mostly like edge detectors. In gaze direction classification problem the position of the iris with respect to the eyelids and eye corners would be different for different classes. The feature maps obtained from the convolution layers would contain information regarding the spatial position and direction of the edges. Through sub-sampling, we improve the translation invariance. Further, the activation introduces nonlinearity which makes the separation of close classes possible. In this particular application, computational requirement during evaluation was another constraint, as we wanted the algorithm to run at camera frame rate. Deeper CNN's take more time to evaluate as the stages are sequential. Hence we arrived at a three layer network empirically as a good tradeoff between speed and accuracy.

\begin{figure}[!t]
\begin{center}
\includegraphics[width=1\linewidth]{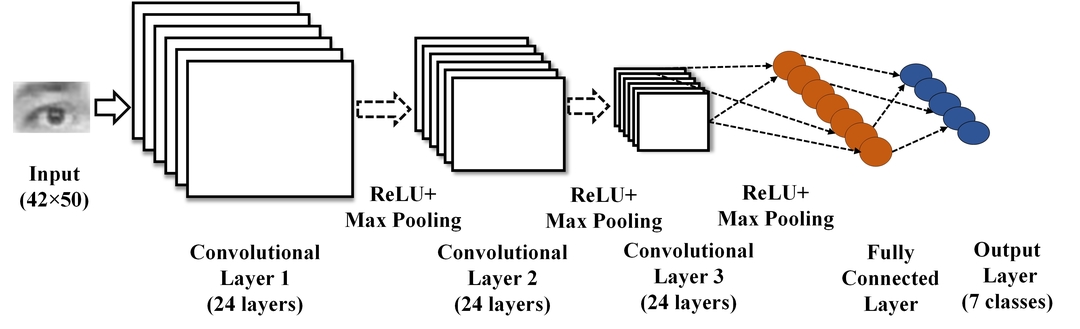}
\end{center}
\caption{Architecture of the CNN  used}
\label{fig:cnn}
\end{figure}

The input stage of the CNN consists of images of dimension $42\times50$ (or $25\times 15$ in the case of ERT). In the first convolutional layer, 24 filters of dimension $7\times7$ are used. This stage was followed by a rectifier linear unit (ReLU). ReLU layer introduces a non-linearity to the activations. The non-linearity function can be represented as:
\begin{equation}
f(x) = \max (0,x)
\end{equation}
where, $x$ is the input and $f(x)$ the output after the ReLU unit.
A max pooling layer is added after the ReLU stage. Max pooling layer performs a spatial sub-sampling of each output images.
We have used  $2\times 2$ max-pooling layers which reduce the spatial resolution to half. Two similar stages with filter dimensions $5\times5$ and $3\times3$ are also added. After the convolutional, ReLU, and max-pooling layers in the third convolutional layer, the outputs from all the activations are joined in a fully connected layer.
The number of output nodes corresponds to the number of classes in the particular application.  The structure of the network is shown in Fig. \ref{fig:cnn}. The softmax loss is used over classes as the error measure. Cross entropy loss is minimized in the training.

The cross entropy loss ($L$) is defined as:
\begin{equation}
L\left( {f(x),y} \right) =  - y\ln (f(x)) - (1 - y)\ln (1 - fx(x))
\end{equation}

where $x$ is the vector to be classified, $y \in \{ 0,1\} $
, where $y$ is the label

The cross entropy loss is convex and can be minimized using stochastic gradient descent (SGD) \cite{bottou2010large} algorithm. The size of convolution kernels remains same for both ERT and ROI ($7\times7$, $5\times5$ and $3\times3$ ).

\subsubsection{Classification of eye gaze direction}
Two CNNs are trained independently for left and right eyes. The scores from both the networks are used to obtain the average score.
\begin{equation}
score = \left( {{{scor{e_L} + scor{e_R}} \over 2}} \right)
\end{equation}
where, $scor{e_L}$ and $scor{e_R}$ denote the scores obtained from left and right CNNs respectively.\\
The class can be found out as the label with maximum probability:
\begin{equation}
class = \mathop {\arg \max }\limits_{label} (score)
\end{equation}

\section{Experiments}
We have conducted experiments in Eye Chimera database \cite{florea2013can}, \cite{vranceanu2013nlp} which contains all the seven gaze directions classes. 
This dataset contains images of 40 subjects. For each subject, images with different gaze directions are available. The total number of images in the dataset is 1170. The ground truth for class labels and fiducial points are also available. 

\subsection{Evaluation procedure}
The database was randomly split into two equal proportions. Training and testing are performed on two completely disjoint 50\% subsets to avoid over-fitting. CNN require a large amount of data in the training phase for better results. The size of the database is relatively small. We have used data augmentation in the training set images to solve this issue. Rotations, blurring, and scaling are performed in the images in the training subset to increase the number of training samples. Two CNNs were
trained separately for left and right eye. In the testing phase, the scores from both left and right eye CNN models are combined to obtain the label of the test image.

In this work, we have considered both 7-class and 3-class classification. All the seven  classes are used in the first case, only a subset of the labels are used in the latter. In 3-class case, we use only classes left, center and right. The classes are denoted as follows :\\
C- Center, CL- Center Left, CR- Center Right, UL-Upper Left, UR- Upper Right, DL- DOwn Left, and DR- Down Right.

 The methodology followed is same in both the cases.

\begin{figure}[!htb]
\begin{center}
\includegraphics[width=.9\linewidth]{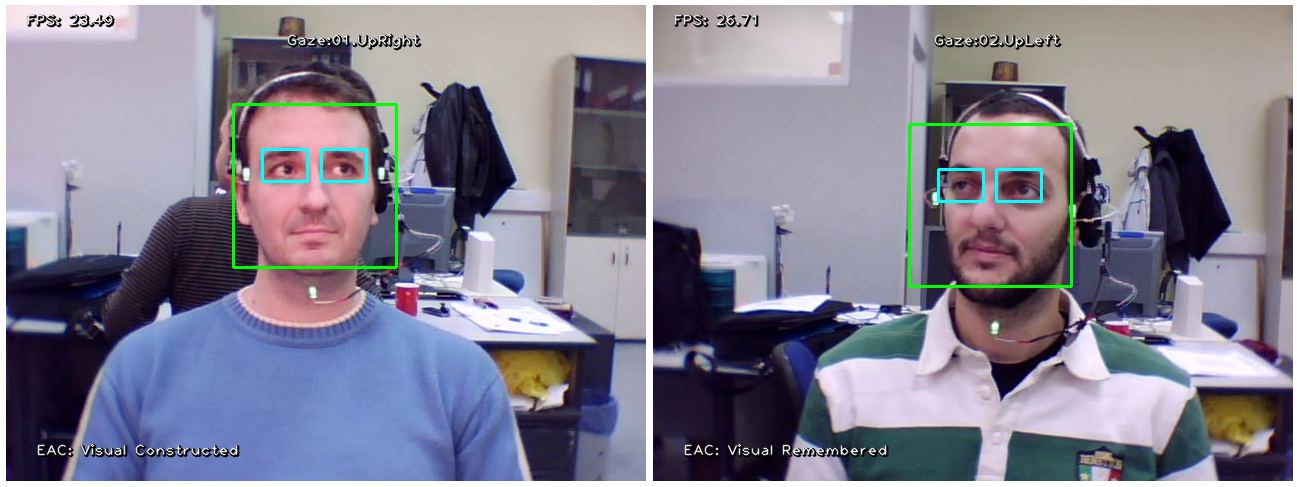}
\end{center}
\caption{Sample results from the framework}
\label{fig:sample}

\end{figure}

\subsection{Results}
The results obtained from the experiments in Eye Chimera dataset is described in this section. The classification accuracy was high in 3-class scenario compared to  7-class case.

We have conducted experiments with the two different methods proposed. In the first case, the eye region localization is carried out using geometrical relations. Explicit landmark detection is avoided in this case. This approach is denoted as ROI. This method reduces one stage in the overall framework. In the second algorithm, we use the ERT based landmark detection scheme. The eye corners obtained are used to constrain the region for subsequent classification stage. The eye region obtained in each image is resized to a resolution of $20 \times 15$ for further processing.

In both the cases (ROI and ERT), the data was divided into two 50\% subsets. CNNs were trained separately for left and right eyes using data augmentation. Testing was carried out with 50\% disjoint testing set to avoid over-fitting effects. All the experiments were repeated in both 3 class and 7 class scenario. 

The results obtained by using only one eye are shown in Table \ref{tab:componeeye}. 

Combining the information from both eyes improves the accuracy. The results obtained using both the eyes and the comparison with state of the art is shown in Table \ref{tab:comptwoeye}. 

In both the cases, the proposed method outperforms all the state of the art algorithms in eye gaze direction classification. Highest accuracy is obtained with ERT+CNN algorithm. The individual accuracies achieved in the 7 class case is shown in Table \ref{tab:classacc}.

\begin{table}[]
\centering
\caption{Comparison of accuracy of classification using only one eye}
\label{tab:componeeye}
\begin{tabularx}{1\linewidth}{@{\extracolsep{\fill}}llcc@{}}
\toprule
\begin{tabular}[c]{@{}l@{}}Eye Boundingbox\\ Localization method\end{tabular} & \begin{tabular}[c]{@{}l@{}}Eye direction \\ classification\\ \\ Method\end{tabular} & \begin{tabular}[c]{@{}c@{}}Recognition \\ Rate\\ 7 class (\%)\end{tabular} & \begin{tabular}[c]{@{}c@{}}Recognition \\ Rate\\ 3 class (\%)\end{tabular} \\ \midrule
BoRMaN \cite{valstar2010facial}\textsuperscript{*}                                                           & Valenti \cite{valenti2008accurate}                                                                    & 32.00                                                                      & 33.12                                                                      \\
Zhu \cite{zhu2012face}\textsuperscript{*}                                                                          & Zhu                                                                                \cite{zhu2012face} & 39.21                                                                      & 45.57                                                                      \\
Vr{\^a}nceanu \cite{vranceanu2015gaze}\textsuperscript{*}                                                                     & Vr{\^a}nceanu \cite{vranceanu2015gaze}                                                                                                                                            & 77.54                                                                      & 89.92                                                                      \\
\textbf{\begin{tabular}[c]{@{}l@{}}\textbf{Proposed}\\ (\textbf{Geometric})\end{tabular}}       & \textbf{\begin{tabular}[c]{@{}l@{}}Proposed\\ (CNN)\end{tabular}}                   & \textbf{81.37}                                                             & \textbf{95.98}                                                             \\
\textbf{\begin{tabular}[c]{@{}l@{}}\textbf{Proposed}\\ \textbf{(ERT)}\end{tabular}}             & \textbf{\begin{tabular}[c]{@{}l@{}}Proposed\\ (CNN)\end{tabular}}                   & \textbf{86.81}                                                             & \textbf{96.98}                                                             \\ \bottomrule
\multicolumn{4}{l}{\textsuperscript{*}\footnotesize{Data taken from \cite{vranceanu2015gaze} for comparison}}

\end{tabularx}
\end{table}

\begin{table}[]
\centering
\caption{Accuracy in classification  of each class (\%)}
\label{tab:classacc}
\begin{tabularx}{1\linewidth}{@{\extracolsep{\fill}}llllllll@{}}
\toprule
Method                                                   & \multicolumn{1}{c}{C} & \multicolumn{1}{c}{UR} & \multicolumn{1}{c}{UL} & \multicolumn{1}{c}{CR} & \multicolumn{1}{c}{CL} & \multicolumn{1}{c}{DR} & \multicolumn{1}{c}{DL}\\ \midrule
\begin{tabular}[c]{@{}l@{}}Proposed\\ (ROI)\end{tabular} & 96                     & 79                     & 87                     & 77                     & 79                     & 75                     & 93                    \\
\begin{tabular}[c]{@{}l@{}}Proposed\\ (ERT)\end{tabular} & 97                     & 93                     & 94                     & 84                     & 87                     & 71                     & 91                    \\ \bottomrule

\end{tabularx}

\end{table}

\begin{table}[]
\centering
\caption{Classification accuracy  (\%)  when both the eyes are used}
\label{tab:comptwoeye}
\begin{tabularx}{1\linewidth}{@{\extracolsep{\fill}}lcccccc@{}}
\toprule
Dataset                                                                      & \multicolumn{1}{l}{Classes} & \multicolumn{1}{l}{\begin{tabular}[c]{@{}l@{}} Valenti\cite{valenti2008accurate}+\\Valstar\cite{valstar2010facial}\textsuperscript{*}\end{tabular}} & \multicolumn{1}{l}{\begin{tabular}[c]{@{}l@{}} Zhu\\ \cite{zhu2012face}\textsuperscript{*}\end{tabular}}  & \multicolumn{1}{l}{\begin{tabular}[c]{@{}l@{}} Vr{\^a}nceanu\\ \cite{vranceanu2015gaze}\textsuperscript{*}                                                                    \end{tabular}}  & \multicolumn{1}{l}{\begin{tabular}[c]{@{}l@{}}\textbf{Proposed}\\ \textbf{(ROI)}\end{tabular}} & \multicolumn{1}{l}{\begin{tabular}[c]{@{}l@{}}\textbf{Proposed}\\ \textbf{(ERT)}\end{tabular}} \\ \midrule
\multirow{2}{*}{\begin{tabular}[c]{@{}l@{}}Still Eye\\ Chimera\end{tabular}} & 7                           & 39.83                                                                            & 43.29                        & 83.08                  & \textbf{85.58 }                                                                       & \textbf{89.81}                                                               \\
                                                                             & 3                           & 55.73                                                                            & 63.01                        & 95.21                  & \textbf{97.65 }                                                                       & \textbf{98.32}                                                               \\ \bottomrule
\multicolumn{7}{l}{\textsuperscript{*}\footnotesize{Data taken from \cite{vranceanu2015gaze} for comparison}}

\end{tabularx}
\end{table}

\begin{figure}[H]
\begin{center}
\includegraphics[width=.5\linewidth]{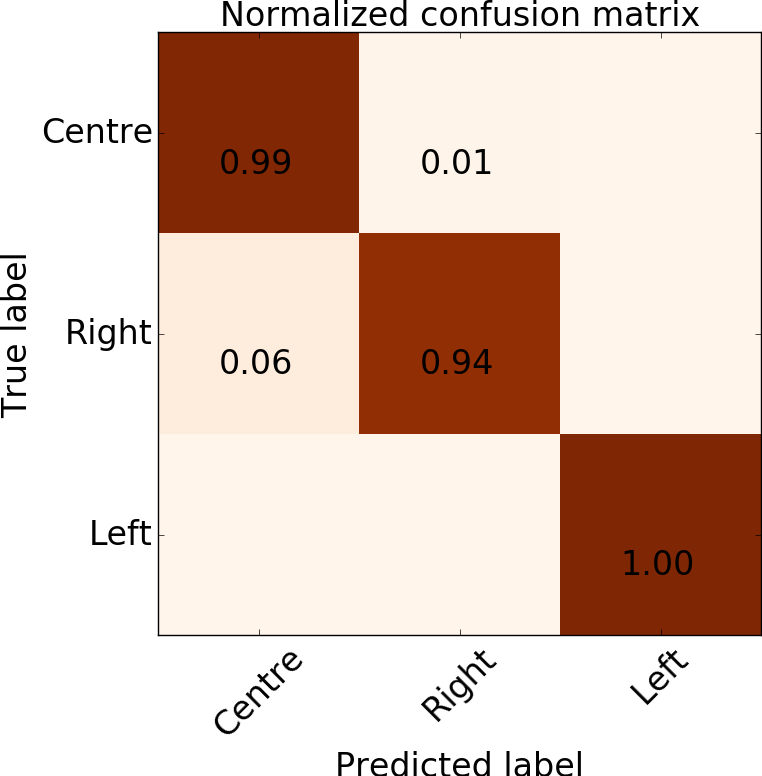}
\end{center}
\vspace{-1.5em}
\caption{Confusion matrix for 3 classes (ERT+CNN)}
\vspace{-1.5em}
\label{fig:conf3}

\end{figure}

\begin{figure}[H]
\begin{center}
\includegraphics[width=.7\linewidth]{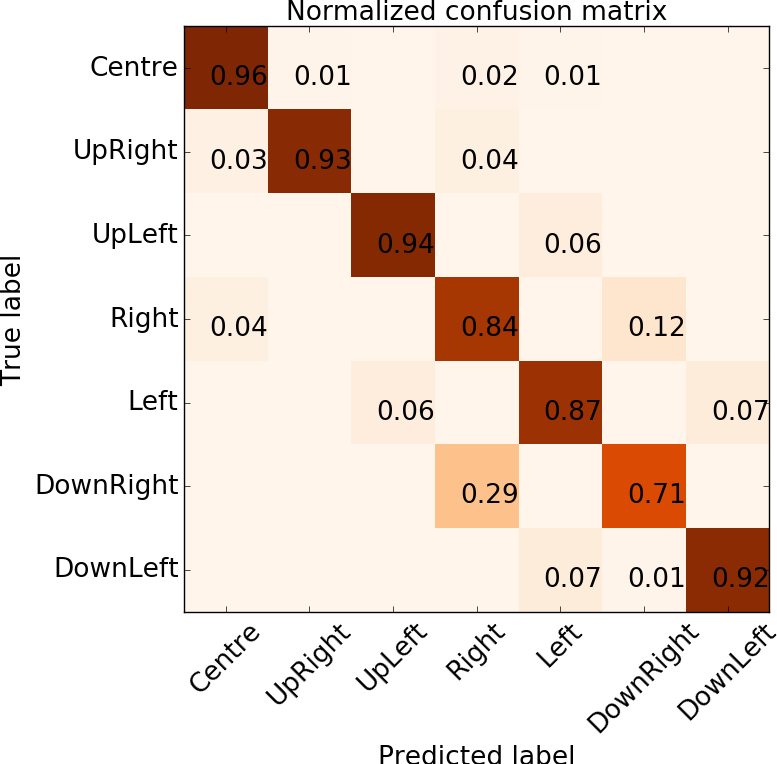}
\end{center}
\vspace{-1.5em}
\caption{Confusion matrix for 7 classes (ERT+CNN)}
\vspace{-1.5em}
\label{fig:conf7}

\end{figure}

The confusion matrix for 3 class and 7 class case (ERT+CNN) are shown in Fig. \ref{fig:conf3} and Fig. \ref{fig:conf7}.

\subsection{Discussion}

The proposed algorithm outperforms all the state of the art results reported in the literature.  The proposed algorithm leverages the information obtained from the facial landmark detection stage to limit the computation to eye regions.  The alignment using eye corners reduces the intra-class variability. Score level fusion from the two separately trained CNN's further improve the accuracy. 

From the confusion matrix, it can be seen that most of the miss-classifications occur in differentiating between right and down right. The classification accuracy is poor in the vertical direction (similar to the observations in \cite{vranceanu2015gaze}).
 This can be attributed to the lack of spatial resolution in the vertical
direction. Most of the cases iris is partly occluded by eyelids in extreme corners. This makes it difficult to classify them accurately. With the larger amount of labeled data, the algorithm could perform even better.

\section{Summary}

In this chapter,  a framework for real-time classification of eye gaze direction is presented. 
The estimated eye gaze direction can also be used to infer eye accessing cues, giving information about the cognitive states. The computational load is very less; we achieved frame rates upto 24 Hz in Python implementation in a 2.0 GHz Core i5 desktop computer running Ubuntu 64 bit OS with 4GB RAM. The per-frame computational time is 42 ms, which is much less than that of the other state of the art methods (250 ms in \cite{vranceanu2015gaze}). Off the shelf webcams can be used for computing the eye gaze direction. The eye gaze direction obtained can be used for HCI applications. The computational requirements of the algorithm in testing phase is less, which makes it suitable for smart devices with low-resolution cameras using pre-trained models. A temporal filtering of the predicted scores can be used in the case of video data. Using the color information in the CNN is another path to be explored.
\end{onehalfspacing}

%% file: Chapter6/chapter_prl_v7.tex
\chapter{Eye Movement based Biometric Authentication}{Eye Movement based Biometric Authentication}

\graphicspath{{Chapter6/pics/}}
\begin{mdframed}[linecolor=grey!3,backgroundcolor=grey!3] 

This section presents a novel framework for biometric identification based on eye movements.  A large set of features is extracted from fixations and saccades to characterize each individual. A backward selection approach is used to identify important features. Two different Gaussian RBF networks are trained using features from fixations and saccades separately.  Score level fusion approach is adopted for biometric authentication. 
The developed framework was evaluated in BioEye 2015 dataset and was found to outperform state of the art methods.
\end{mdframed}
\vspace{5mm}

\begin{onehalfspacing}

\section{Introduction}

Biometrics is an active area of research in pattern recognition and machine learning community. Potential applications of biometrics include forensics, law enforcement, surveillance, personalized interaction, access control \cite{jain2007handbook}, etc. Physiological features like fingerprint, DNA, earlobe geometry, iris pattern, facial recognition, \cite{jain2004introduction} are widely used in biometrics. Recently, several behavioral biometric modalities have been proposed including gait, eye movement patterns, keystroke dynamics \cite{wang2009behavioral} signature, etc. Even though many such parameters like brain signals \cite{marcel2007person} (using electroencephalogram) and heart beats \cite{plataniotis2006ecg} have been proposed as biometric modalities, their invasive nature limits their practical applications.

An effective biometric should have the following characteristics \cite{jain2007handbook}: (1) the features should be unique for each individual, (2) they should not change with time (template aging effects), (3) acquisition of parameters should be easy (low computational complexity and noninvasive), (4) accurate and automated algorithms should be available for classification, (5) counterfeit resistance, (6) low cost, and (7) ease of implementation. Other characteristics that might make the system more robust are portability and the ability to extract features from non-co-operative subjects.

Out of many biometric modalities, iris recognition has shown the most promising results \cite{ib2005independent} obtaining equal error rates (EER) close to 0.0011\%. However, it can only be used when the user is co-operative. Such systems can be spoofed by contact lenses with printed patterns. Even though most of the biometric modalities perform well on evaluation databases, one may be able to spoof such systems with mechanical replicas or artificially fabricated models \cite{roberts2007biometric}. In this regard, several approaches have been presented \cite{schuckers2002issues} to detect the liveliness of tissues or body parts presented to the biometric system. However, such methods are also vulnerable to spoofing.

Biometrics using patterns obtained from eye movements is a relatively new field of research. Most of the conventional biometrics use physiological characteristics of the human body. Eye movement-based biometrics tries to identify the behavioral patterns as well as information regarding physiological properties of tissues and muscles generating eye movements \cite{leigh1999neurology}. They provide abundant information about cognitive brain functions and neural signals controlling eye movements.

Eye movements as observed from outside is generated from an oculomotor plant which consists of six extra-ocular muscles.  Four of them are responsible for horizontal and vertical movements namely the lateral and medial recti (horizontal) and the superior and inferior recti (vertical). The torsional and coordinate rotations of the eye are controlled by the other two muscles, the superior and the inferior oblique.  These muscles are controlled by axons of oculomotor, trochlear and abducent nerves. There are complex mechanisms which control eye movements, both physiological (structure of oculomotor system) and behavioral (the neural circuitry guiding visual attention).  
The dynamics of the eye movement can be modeled as an input output system where the neuron input to the muscles as inputs and the eye movement as the output. The dynamics of the movement is affected by the parameters of the system. Here the physiological properties of the tissues and muscles including their elasticity, rotational inertia are manifested as the parameters of the model. Manifestations of the unique combination of these parameters could be used as a biometric trait.

 Saccadic eye movement is the fastest movement (peak angular velocities up to 900 degrees per second) in the human body. Mechanically replicating such a complex oculomotor plant model is extremely difficult. These properties make eye movement patterns a suitable candidate for biometric applications. The dynamics of eye movement along with these properties can give inbuilt liveliness detection capability.

Initially, eye movement biometrics has been proposed as a soft biometric. However, with the high level of accuracy achieved, it seems there are more opportunities regarding its application as an independent biometric modality. Eye movement detection can be integrated easily into already existing iris recognition systems. A combination of iris recognition and eye movement pattern recognition may lead to a robust counterfeit-resistant biometric modality with embedded liveliness detection and continuous authentication properties. Eye movement biometrics can also be made task-independent \cite{kinnunen2010towards} so that the movements can be captured even for non-co-operative subjects.

\section{Related works}

Initial attempts to use eye movements as a biometric modality were carried out by Kasprowski and Ober \cite{kasprowski2004eye}. They recorded the eye movements of subjects following a jumping dot on a screen. Several frequency domain and Cepstral features were extracted from this data. They applied different classification methods like naive Bayes, C45 decision trees, SVM and KNN methods. The results obtained further motivated research in eye movement-based biometrics. Bednarik \textit{et al.} \cite{bednarik2005eye} conducted experiments on several tasks including text reading, moving cross stimulus tracking and free viewing of images. They used FFT and PCA on the eye movement data. Several combinations of such features were tried. However, the best results were obtained using the distance between eyes, which is not related to eye dynamics. Komogortsev \textit{et al.} \cite{komogortsev2010biometric} used an Oculomotor Plant Mathematical Model (OPMM) to model the complex dynamics of the oculomotor plant. The plant parameters were identified from the eye movement data. This approach was further extended in \cite{komogortsev2012biometric}. Holland and Komogortsev \cite{holland2013complexb} evaluated the applicability of eye movement biometrics with different spatial and temporal accuracies and various types of stimuli. Several parameters of eye movements were extracted from fixations and saccades. Weighted components were used to compare different samples for biometric identification. A temporal resolution of 250 Hz and spatial accuracy of 0.5 degrees were identified as the minimum requirements for accurate gaze-based biometric systems. Kinnunen \textit{et al.} \cite{kinnunen2010towards} presented a task-independent user authentication system based on eye movements. Gaussian mixture modeling of short-term gaze data was used in their approach. Even though the accuracy rates were fairly low, the study opened up possibilities for the development of task-independent eye movement-based verification systems. Rigas \textit{et al.} \cite{rigas2012biometric} explored variations in individual gaze patterns while observing human face images. Eye movements resulted were analyzed using a graph-based approach. The Multivariate Wald\textendash Wolfowitz runs test was used to classify the eye movement data. This method achieved 70\% rank-1 IR and 30\% EER on a database of 15 subjects. Rigas \textit{et al.} \cite{rigas2012human} extended this method using features of velocity and acceleration calculated from fixations.  The feature distributions were compared using Wald\textendash Wolfowitz test.

Zhang \textit{et al.} \cite{zhang2012biometric} used saccadic eye movements with machine learning algorithms for biometric verification. They used multilayer perceptron networks, support vector machines, radial basis function networks and logistic discriminant for the classification of eye movement data. Recently Cantoni \textit{et al.} \cite{cantoni2015gant} proposed a gaze analysis technique called GANT in which fixation patterns were denoted by a graph-based representation. For each user, a fixation model was constructed using the duration and number of visits at various points. Frobenius norm of the density maps was used to find the similarity between two recordings. Holland and Komogortsev presented an approach (CEM) \cite{holland2011biometric} using several scan path features including saccade amplitudes, average saccade velocities, average saccade peak velocities, velocity waveform, fixation counts, average duration of fixation, length of scan path, area of scan path, regions of interest, number of inflections, main sequence relationship, pairwise distances between fixations, amplitude duration relationship, etc. A comparison metric of the features was computed using Gaussian cumulative density function. Another similarity metric was obtained by comparing the scan paths. A weighted fusion of these parameters obtained the best case EER of 27\%. Holland and Komogortsev proposed a method (CEM-B)\cite{holland2013complexa}, in which the fixation and saccade features were compared using statistical methods like Ansari\textendash Bradley test, two-sample t-test, two-sample Kolmogorov\textendash Smirnov test, and the two-sample Cramer\textendash von Mises test. Their approach achieved 83\% rank-1 IR and 16.5\% EER on a dataset of 32 subjects.

To the best knowledge of the authors, the best case EER obtained is 16.5\% \cite{holland2013complexa}. Most of the works presented in the literature were evaluated on smaller databases. The effect of template aging was not considered in these works. For the application of eye movement as a reliable biometric, the patterns should remain consistent with time. In this work, we try to improve upon the existing methods. The proposed algorithm can reach an EER up to 2.59\% and rank-1 accuracy of 89.54\% in RAN\_30min dataset of BioEye 2015 database \cite{bioeye} containing 153 subjects. Template aging effect has also been studied using data taken after an interval of 1 year. The average EER obtained is 10.96\% with a rank-1 accuracy of 81.08\% with 37 subjects.
\section{Proposed method}
In the proposed approach, eye movement data from the experiment are classified into fixations and saccades, and their statistical features are used to characterize each individual. For each individual, the properties of saccades of same durations have been reported to be similar \cite{collewijn1988binocular}. We use this knowledge and extract the statistical properties of the eye movements for biometric identification. Different stages of the algorithm are described below.

\subsection{Details about the data recording}

Gaze sequences were obtained using two distinct types of visual stimuli. In one set (RAN), a white dot moving in a dark background was used as the stimulus, and the subjects were asked to follow the dot. Text excerpt shown on the screen was used as the stimulus in the other set (TEX).  Eye tracking data was recorded with 1000 Hz followed by downsampling to 250 Hz with anti-aliasing filtering.

Data recorded in three different sessions are available. The data from the first session was used for the enrollment.  Other two sessions were used for testing the accuracy. The second session was conducted after 30 minutes containing recordings of 153 subjects.  A third session, conducted after one year, (37 subjects) is also available to evaluate the robustness against template aging. The dataset used was part of BioEye 2015 competition.

\subsection{Data pre-processing and noise removal}
\begin{figure}[H]
\centering
\includegraphics[width=0.8\linewidth]{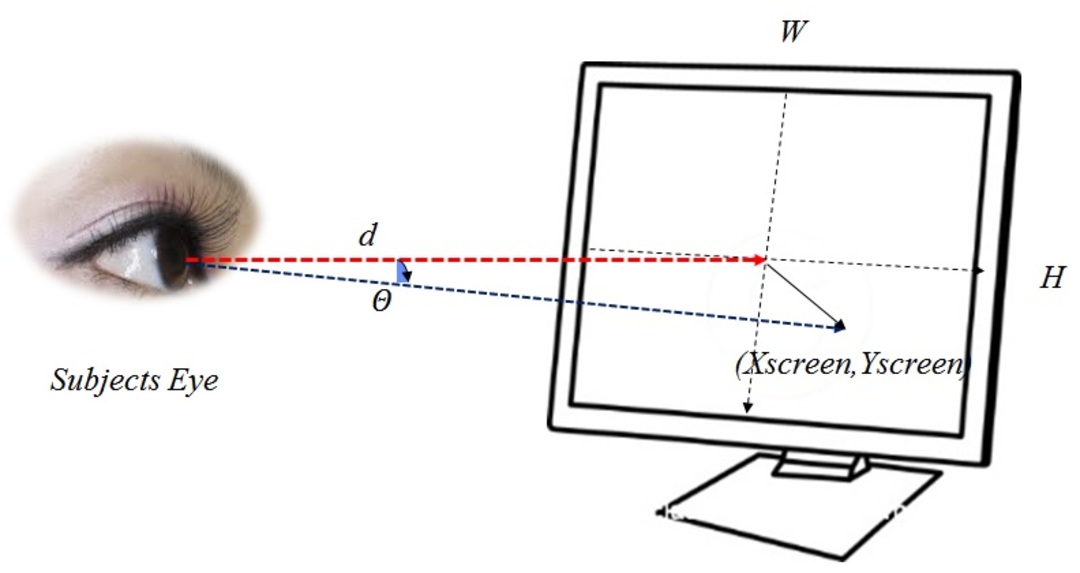}
\caption{The arrangement for gaze recording}
\label{fig:geometry}
\end{figure}

The data contains visual angles in both $x$ and $y$ directions along with stimulus angles. Information about the validity of samples is also available. Eye movement data has been captured at a sampling frequency of 1000 Hz. The data obtained is decimated to 250 Hz using an anti-aliasing filter. In the proposed feature extraction method, most of the parameters are computed with reference to the screen coordinate system. Hence, in the pre-processing stage, the data obtained is converted to screen
coordinates based on head distance and geometry of the acquisition system as
\begin{equation}
{x_{screen}} = \left( {{{d * {w_{pix}}} \over w}} \right)\tan ({\theta _x}) + {{{w_{pix}}} \over 2}
\end{equation}
\begin{equation}
{y_{screen}} = \left( {{{d * {h_{pix}}} \over h}} \right)\tan ({\theta _y}) + {{{h_{pix}}} \over 2}
\end{equation}
where, $d,{\theta _x}$ and ${\theta _y}$ denote distance from the screen and visual angles in $x$ and $y$ direction (in radian) respectively. ${x_{screen}}$ and ${y_{screen}}$ denote the position of gaze on the screen. ${w_{pix}},{h_{pix}}$,$w,h$ denote resolution and physical size of the screen in horizontal and vertical directions respectively (Fig. \ref{fig:geometry}). The distance of the face from the screen and the dimensions of the recording systems were provided with the dataset. However, most of the commercial eye tracking systems report 2D (or 3D) gaze position directly, obviating the need for this step.

Raw eye gaze positions may contain noise. Most of the features used in this work are extracted from velocity and acceleration profiles. The presence of noise makes it difficult to estimate the velocity and acceleration parameters using differentiation operation. Eye movement signals contain high-frequency components, especially during saccades. High-frequency components would be more prominent in velocity and acceleration profiles \cite{harris1984instrument}. Savitzky\textendash Golay filters are useful for
filtering out the noise when the frequency span of the signal is large \cite{krishnan2013selection}. They are reported to be optimal \cite{savitzky1964smoothing} for minimizing the least-square error in fitting a polynomial to frames of the noisy data. We use this filter with polynomial order of 6 and frame size of 15 in our approach.

\subsection{Eye movement classification and feature extraction }
\subsubsection{Eye movement classification}
The I-VT (velocity threshold) algorithm \cite{holland2012biometric}, \cite{salvucci2000identifying} is used to classify the filtered eye movement data into a sequence of fixations and saccades (Algorithm \ref{alg:eyeclassification}). Most of the earlier works specify the velocity threshold for angular velocity. The angular velocity computed from the filtered data is used to classify the eye movements. A velocity of 50 degrees per second is used as the threshold in I-VT algorithm.

\begin{algorithm}[H]
\caption{Fixation and Saccade classification algorithm}
\label{alg:eyeclassification}
\begin{algorithmic}[1]
\INPUT [Time,Gazex,Gazey]
\OUTPUT Res
\State \textbf{Constants}: VT=Velocity threshold,
 MDF=Minimum duration for fixation
\State \textbf{States} $=$[FIXATION,SACCADE]
\State fixationStart=1
\State Velocity=\textit{smoothDiff}(data)
\State $ N\leftarrow$ Number of samples of data
\For {$index \leftarrow$ 1 \textbf{to} N}
\If{Velocity[index] $<$ VT}
\State currentState=FIXATION
\If{lastState $\neq$ currentState}
\State fixationStart = index
\EndIf
\Else
\If{lastState$ =$ FIXATION}
\State  duration = data(index,1) - data(fixationStart,1)
\If {duration $<$ MDF}
\For {$i \leftarrow$ fixationStart \textbf{to} index}
\State res[i]= SACCADE
\EndFor
\EndIf
\EndIf
\State currentState=SACCADE
\EndIf
\State lastState=currentState
\State res[index]=currentState
\EndFor
\State Res $\leftarrow$ res
\end{algorithmic}
\end{algorithm}
\begin{figure}[H]
\begin{center}
\includegraphics[width=0.7\linewidth]{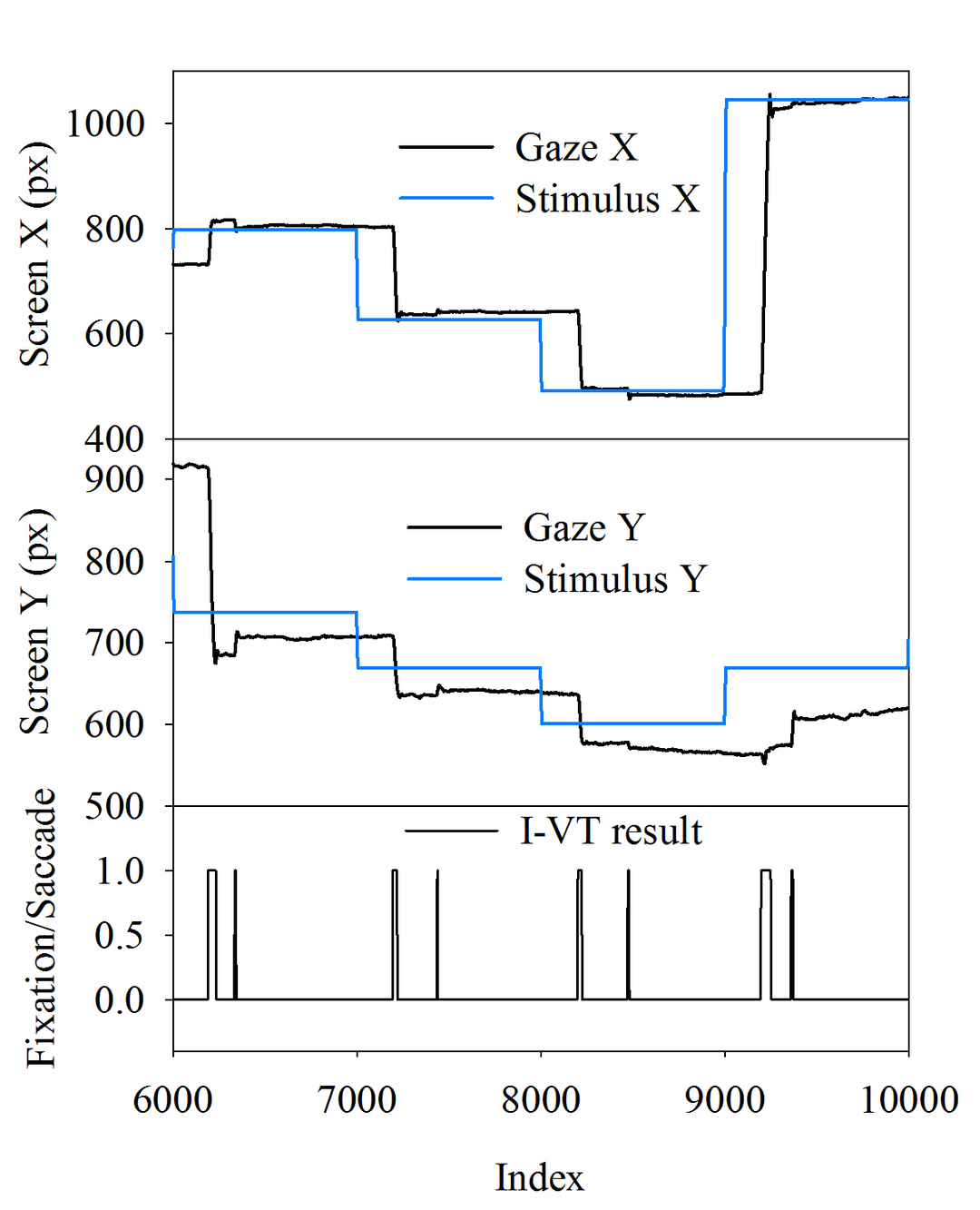}
\end{center}
\caption{Gaze data and stimulus for RAN\_30min sequence}
\label{fig:gazedata}
\label{fig:onecol}
\end{figure}
A minimum duration threshold of 100 ms has been chosen to reduce the false positives in
fixation identification. Algorithm \ref{alg:eyeclassification} returns the classification results for each data point as either fixation or saccade. Points that are not a part of fixations are considered as saccades in this stage. In the proposed approach, we consider saccades with their
durations more than a specified threshold to minimize the effect of spurious saccade segments. From the results of Algorithm \ref{alg:eyeclassification}, a list containing starting index and duration of all fixations and saccades is created. A post-processing stage is carried out to remove small-duration saccades. Saccades with duration less than 12 ms are removed in this stage.

\subsubsection{Feature extraction}
After the removal of small-duration saccades, each eye movement data
is arranged into a sequence of fixations and saccades. The sequence of gaze locations and corresponding visual angles are also available for each fixation and saccade. Several statistical features are extracted from the position, velocity and acceleration profiles of
the gaze sequence. Other features like duration, dispersion, path length and co-occurrence features
are also extracted for both fixations and saccades. Earlier works \cite{komogortsev2010biometric} suggested that saccades provide a rich amount of information about the dynamics of the oculomotor plant. Hence, we extract several other parameters including the saccadic ratio, main sequence, angle, etc. Saccades in horizontal and vertical directions are generated by different areas of the brain \cite{harwood2008optimally}. We use the statistical properties of the gaze data in $x$ and $y$ directions to incorporate this information. The distance and angle with the previous fixation/saccade are also used as features to leverage the temporal properties. The method used for computation of features is described below.

\begin{figure}[!htb]
\centering
\includegraphics[width=0.9\linewidth]{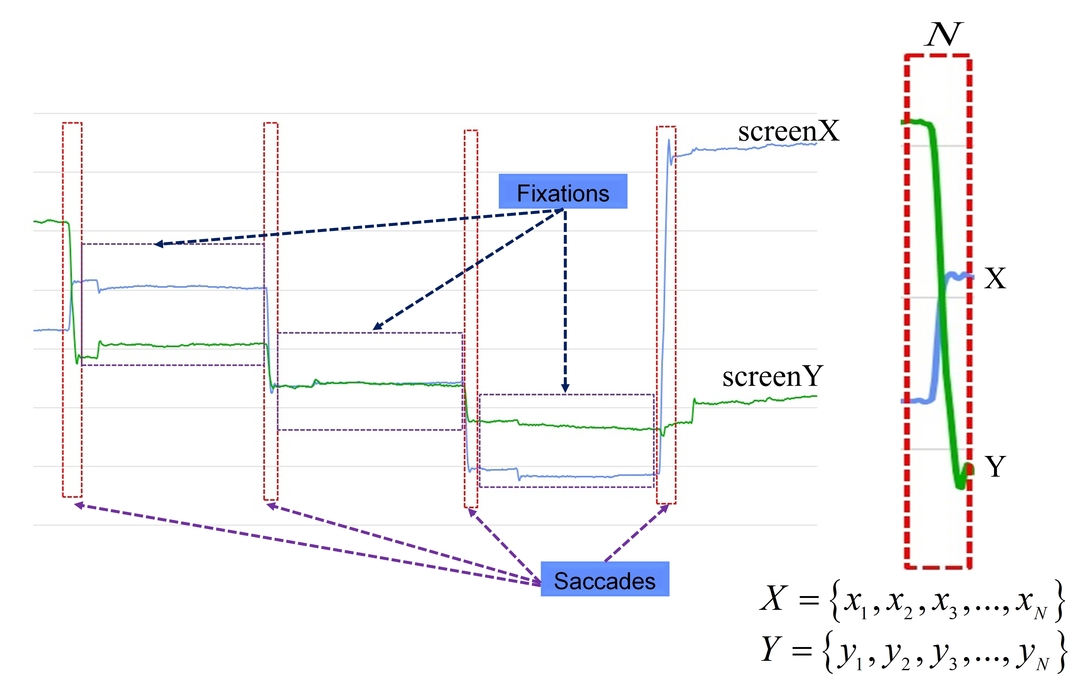}
\caption{Classification of the raw sequence}
\label{fig:raw}
\end{figure}

Figure. \ref{fig:raw} depicts the time series of $x$ and $y$ positions of gaze. The dotted red rectangles denote the saccade sections segmented out using the I-VT algorithm. The region between two saccade regions constitutes a fixation. For each fixation or saccade segment we represent, the fixation or saccade using sets of $x$ coordinates and $y$ coordinates as $X$ and $Y$.

Let $X{\rm{ }} = {\rm{ }}\left\{ {{x_1},{x_2},{x_3},...,{x_N}} \right\}$
and $Y{\rm{ }} = {\rm{ }}\left\{ {{y_1},{y_2},{y_3},...,{y_N}} \right\}$ denote the set of coordinate positions of gaze in
each fixation/saccade and let $N$ denotes the number of data points in any fixation or saccade.
$\left( {{x_i},{y_i}} \right)$ denotes gaze location on the screen coordinate system and $\left( {\theta _i^x,\theta _i^y} \right)$ denotes the corresponding horizontal and vertical visual angles.

A large number of features are extracted from the gaze sequence in each fixation and saccade. Some features are derived from the angular velocity. The differentiation operation for finding velocity and acceleration is carried out using forward difference method on the smoothed data. List of features extracted from fixations and saccades along with the methods of computation are shown in Table \ref{Tab:fixation_featuresused} and Table \ref{Tab:saccade_featuresused}. The features are extracted independently for each fixation and saccade.

\begin{sidewaystable}

\caption{\label{Tab:fixation_featuresused}List of features extracted from fixations}
\centering
\begin{tabularx}{1\linewidth}{@{\extracolsep{\fill}}llll}
\hline
\multicolumn{2}{l}{Used in} & \multicolumn{1}{c}{Fixation features} & Description \\ \cline{1-2}
TEX & RAN & & \\ \hline
N & Y & Fixation duration & Obtained from I-VT result \\
N & N & Standard deviation (X) & \begin{tabular}[c]{@{}l@{}}From screen coordinates\\ during fixation\end{tabular} \\
Y & N & Standard deviation (Y) & '' \\
Y & Y & Path length & \begin{tabular}[c]{@{}l@{}}Length of path traveled in screen\\ $Path\,Length = \sum\limits_{i = 1}^{N - 1} {\sqrt {{{\left( {{x_{i + 1}} - {x_i}} \right)}^2} + {{\left( {{y_{i + 1}} - {y_i}} \right)}^2}} } $\end{tabular} \\
Y & Y & Angle with previous fixation & \begin{tabular}[c]{@{}l@{}}Angle with centroid of \\ previous fixation\end{tabular} \\
Y & Y & Distance from the last fixation & Euclidean distance from the previous fixation \\
Y & Y & Skewness(X) & From Screen coordinates \\
Y & Y & Skewness(Y) & '' \\
N & N & Kurtosis(X) & '' \\
Y & Y & Kurtosis(Y) & '' \\
Y & Y & Dispersion & \begin{tabular}[c]{@{}l@{}}Spatial spread during a fixation, Computed as\\$D= \left( {\max (X) - min(X)} \right) + \left( {\max (Y) - min(Y)} \right)$\end{tabular} \\
Y & Y & Average velocity & $AV = ppath\,length/duration$ \\ \hline
\multicolumn{4}{l}{$Y$ and $N$ denote inclusion or exclusion of the feature in the particular stimulus after feature selection}
\\ \hline
\end{tabularx}

\end{sidewaystable}



\begin{sidewaystable}
\caption{\label{Tab:saccade_featuresused}List of features extracted from saccades}
\centering
\begin{tabularx}{1\linewidth}{@{\extracolsep{\fill}}llll}
\hline
\multicolumn{2}{l}{Used in} & \multicolumn{1}{c}{Saccade features} & Description \\
TEX & RAN & & \\ \hline
N & N & Saccadic duration & Obtained from I-VT result \\
Y & Y & Dispersion & $D= \left( {\max (X) - min(X)} \right) + \left( {\max (Y) - min(Y)} \right)$\\
NYYYYY & NNNYYY & M3S2K(angular velocity) & Features from angular velocity \\
YYYYYN & YYYYYY & M3S2K(angular acceleration) & Features from angular acceleration \\
Y & Y & Standard deviation(X) & Obtained from screen positions \\
Y & Y & Standard deviation(Y) & '' \\
Y & Y & Path length & \begin{tabular}[c]{@{}l@{}}Distance traveled in screen, \\$\sum\limits_{i = 1}^{N - 1} {\sqrt {{{\left( {{x_{i + 1}} - {x_i}} \right)}^2} + {{\left( {{y_{i + 1}} - {y_i}} \right)}^2}} }$\end{tabular} \\
Y & Y & Angle with previous saccade & \begin{tabular}[c]{@{}l@{}}Difference in saccadic angle with \\ previous saccade\end{tabular} \\
Y & Y & Distance from the previous saccade & \begin{tabular}[c]{@{}l@{}}Euclidean distance between\\ the centroid of the previous\\ saccade\end{tabular} \\

Y & Y & Saccadic ratio & $SR = \max (angular\,velocity)/saccade\,duration$ \\
Y & Y & Saccade angle & \begin{tabular}[c]{@{}l@{}}Obtained from first and last points\\ as, $saccade\,angle = {\tan ^{ - 1}}\left( {\frac{{{y_N} - {y_1}}}{{{x_N} - {x_1}}}} \right)$\end{tabular} \\
Y & Y & Saccade amplitude &Obtained as: $\sqrt {{{\left( {{x_N} - {x_1}} \right)}^2} + {{\left( {{y_N} - {y_1}} \right)}^2}} $\\
YYYYYY & YYYYYY & M3S2K(Velocity\_X\_direction) & Features from screen positions \\
YYYYYY & YYYYNY & M3S2K(Velocity\_Y\_direction) & '' \\
YYYYYY & YYYYYY & M3S2K(Acceleration\_X\_direction) & '' \\
YYYYYY & YYNYYY & M3S2K(Acceleration\_Y\_direction) & '' \\
& & & \\ \hline
\multicolumn{4}{l}{*M3S2K - Statistical features:} \\
\multicolumn{4}{l}{Mean,Median,Max,Std, Skewness, Kurtosis}
\\ \hline
\multicolumn{4}{l}{$Y$ and $N$ denote inclusion or exclusion of the feature in the particular stimulus after feature selection} \\ \hline
\end{tabularx}
\end{sidewaystable}


The control mechanisms generating fixations and saccades are different. The number of fixations and saccades is also different in each recording. There is a total of 12 and 46 features extracted from fixations and saccades respectively. A feature normalization scheme is used to scale each feature into a common range to ensure equal contribution in the final classification stage.

\subsubsection{Feature selection}

The large number of features extracted may contain redundancy and correlation. A backward feature selection algorithm, as shown in Algorithm \ref{alg:backward} is used to retain a minimal set of discriminant features. We use the wrapper-based approach \cite{kohavi1997wrappers} for selecting the features. An RBFN classifier is used for finding the Equal Error Rate (EER) in each iteration. Cross-validation has been carried out in the training set to avoid overfitting. We used a random 50\% subset of the development dataset for the feature selection algorithm.
Feature selection algorithm starts with a set of all the features. Now in each iteration, the EER with inclusion and exclusion of a particular feature is found. The feature is retained if the EER with the use of the feature is better than EER with exclusion. The procedure is repeated for all the features in a sequential manner. The feature selection algorithm is
iterated ten times each time on a random 50\% subset for cross-validation. After these iterations, a set of important features is retained. To evaluate the generalization ability of the selected features, we have tested the algorithm (with the selected features) on an entirely disjoint set
that was not used in the feature selection process. The results with the evaluation set \cite{bioeye}(as shown in
the public results of BioEye 2015 competition) show the stability and generalization capability of the selected features. The subset of
features selected were different for different stimuli (TEX and RAN sets). The list of features selected for TEX and RAN stimuli is shown in Table \ref{Tab:fixation_featuresused} (Fixation features) and Table \ref{Tab:saccade_featuresused} (Saccade features). The features thus selected are used as inputs to the classification algorithm.


\begin{algorithm}[H]
\caption{Backward feature selection}
\label{alg:backward}
\begin{algorithmic}[1]
\INPUT Feature matrix
\OUTPUT featureList[1: Included,0:Excluded]
\State $N\leftarrow$ Number of features
\State $ featureList\leftarrow ones(N)$
\For {$i \leftarrow$ 1 \textbf{to} N}
\State $W \leftarrow featureList$
\State $E \leftarrow +Inf$
\For {$j \leftarrow$ 0 \textbf{to} 1}
\State $ W[i] \leftarrow j$
\State T $\leftarrow$ EER with included features using RBFN
\If{$T < E$}
\State $featureList[i]\leftarrow j$
\State $ E \leftarrow T$
\EndIf
\EndFor
\EndFor
\end{algorithmic}
\end{algorithm}
After obtaining the set of features from fixations and saccades, we develop a model to represent the data. It has been empirically observed that the performance of classification approaches with Kernel-based methods is better than linear classifiers. It has also been reported that the parameters like amplitude-duration and amplitude-peak velocity may vary with the angle of saccade \cite{goossens1997human}. The nature of saccade dynamics may be different in different directions as the stimulus is changing randomly at various points on the screen. For each person, saccades of different amplitudes and directions form clusters in the feature space. In order to use the multi-mode nature of the data, we represent them by clustering them in the feature space. Representative vectors from each cluster are used to characterize each person. We use Gaussian radial basis function network (GRBFN) to model these data. The multiple cluster centers in the feature space are used as representative vectors in this approach. These vectors are selected using the \textit{K}-means algorithm. Two different RBFNs are trained separately for fixation and saccade. Details about the structure of network and score fusion stage are described in the following section.
\subsection{RBF network} 
\label{sec:rbf}
Radial basis function network (RBFN) is a class of neural networks initially proposed by Broomhead and Lowe \cite{broomhead1988radial}. Classification in RBFN is done by calculating the similarity between training and test vectors. Multiple prototype vectors corresponding to each class are stored in each neuron. The Euclidean distance between the input vector and the prototype vector is used to calculate neuron activations.

\begin{sidewaysfigure}
\centering
\includegraphics[width=1\linewidth]{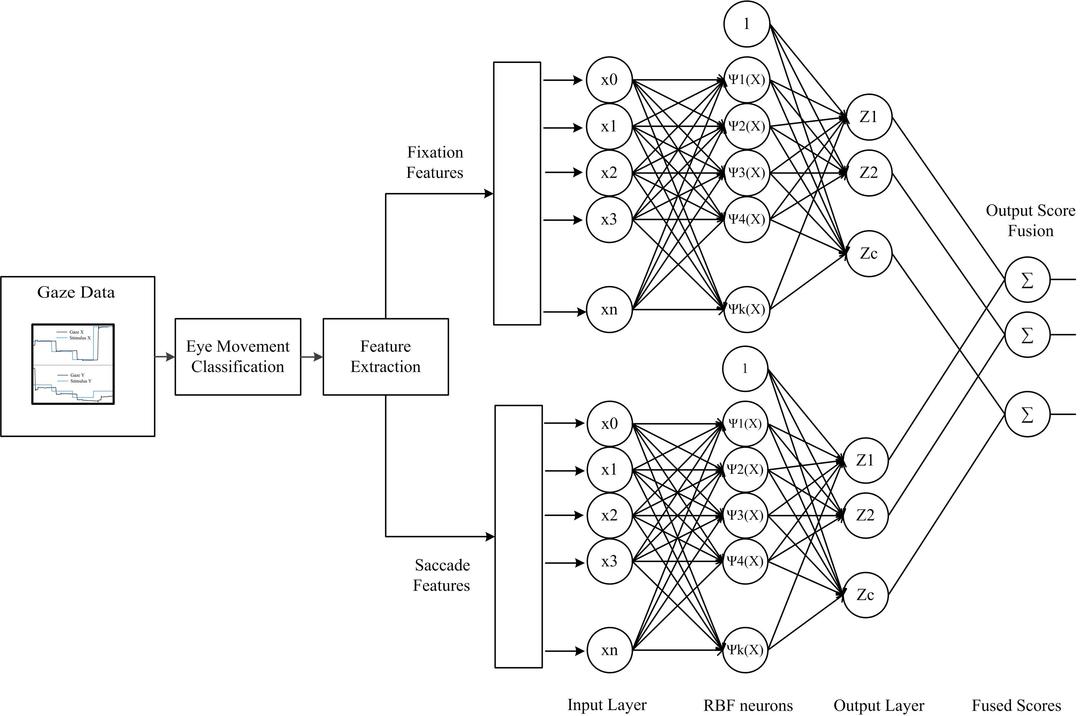}
\caption{Schematic of the proposed framework.}
\label{fig:rbfncombined}
\end{sidewaysfigure}

In the RBF network, input layer is made of feature vectors (Fig. \ref{fig:rbfncombined}). $\varphi (x)$ is a radial basis function that finds the Euclidean distance between the input vector and the prototype vector. A weighted combination of scores from the RBF layer is used to classify the input into different categories.

The number of prototypes per class can be defined by the user, and these vectors can be found from the data using different algorithms like \textit{K}-means, Linde\textendash Buzo\textendash Gray (LBG) algorithm, etc.

The Gaussian activation function of each neuron is chosen as:
\begin{equation}
\varphi (x) = {e^{ - \beta {{\left\| {x - \mu } \right\|}^2}}}
\end{equation}
where,
$\mu $ is the mean of the distribution. The parameter $\beta $ can be found from the data.

In this work, we have used \textit{K}-means algorithm for selecting the representative vectors. For each
person, 32 clusters for fixations and 32 cluster centers for saccades are kept, resulting in $32N$
clusters for each RBFN (where $N$ is the number of persons in the dataset). The number of clusters
to keep is obtained empirically. We have clustered the fixations/saccades of each
individual separately to obtain a fixed number of representative vectors for each person.
A maximum of 100 iterations is used to form the clusters. A standard \textit{K}-means algorithm is used with squared Euclidean distance, and the centers are updated in each
iteration. Each data point is assigned to the closest cluster center obtained from the \textit{K}-means algorithm. For a particular neuron, the value of $\beta$ is computed from
the distance of all points belonging to that particular cluster as
\begin{equation}
\beta = {1 \over {2{\sigma ^2}}}
\end{equation}
where $\sigma$ is the mean Euclidean distance of the points (assigned to the specific neuron) from the centroid of the corresponding cluster.

\subsubsection{Notations}
The biometric identification problem is similar to a multiclass classification problem. Let there be $n$ samples of a $p$ dimensional data. Assume there are $m$ classes (corresponding to $m$ different individuals) with $c$ samples per class ($n = mc$). Let
${y_i}$ be the label corresponding to
${i^{th}}$ sample. Let $K$ be the number of representative vectors from each class. The value of $K$ is chosen empirically ($K=32$).

\subsubsection{Network learning}
The activations can be obtained as:
$A = {\varphi _{i,j}}({x_k}),\,\,i = 1,...,K,\,\,j = 1,...,m,\,\,k = 1,...,n$

The output of the network can be represented as a linear combination of the RBF activations as

\begin{equation}
f(x) = \sum\limits_{j = 1}^m {{w_j}{\varphi _j}(x)}
\end{equation}

where,
$f(x)$ contains the class membership in vector form. Given the activations and output labels, the objective of the training stage is to find the weight parameters of the output layer. The weights are obtained by minimizing the sum of squared errors.

The output layer is represented by a linear system as:
\begin{equation}
A\hat w = \hat y
\end{equation}
The optimal set of weights can be found using the Moore\textendash Penrose pseudoinverse. 
Alternatively, these weights can be learned through gradient descent method.
In the learning phase, features extracted from each fixation and saccade are used to train the model. Each fixation/saccade
is treated as a sample in the training process.

The method described here uses two-phase learning. RBF layer and weight layer trainings are carried out separately. However, a joint training similar to back-propagation is also possible \cite{schwenker2001three}.

\subsubsection{Training stage}
Only the session 1 data from the datasets are used in the training stage. Cluster centers and corresponding $\beta$ values are computed separately for each person (resulting in $32N$ neurons for both fixation and saccade RBFNs). The output weights (${\hat w_{fix}}$ and ${\hat w_{sacc}}$) are found using all fixations and saccades from all the subjects in the dataset.

\subsubsection{Testing stage}
Session 2 data is used in the testing stage. Parameters of RBFN are computed separately for fixations and saccades in the training session. The scores from both RBFNs are combined to obtain the final result. The overall configuration of the scheme is shown in Fig. \ref{fig:rbfncombined}.

 For an unlabeled probe, the activations for each fixation and saccade (${A_{fix}}$ and ${A_{sacc}}$) are found separately using the cluster centers obtained in the training stage.
The final classification is carried out using the combined score obtained from all saccades and
fixations. 
 Let $n{}_{fix}$ and $n{}_{sacc}$ be the number of fixations and saccades in an unlabeled gaze sequence.
The combined score can be obtained as:
\begin{equation}
score = \lambda {1 \over {{n_{fix}}}}\sum\limits_{i = 1}^{{n_{fix}}} {A_{fix}^i{{\hat w}_{fix}} + \left( {1 - \lambda } \right){1 \over {{n_{sacc}}}}} \sum\limits_{i = 1}^{{n_{sacc}}} {A_{sacc}^i{{\hat w}_{sacc}}}
\end{equation}

where, $\lambda \in \left[ {0\,1} \right]$ is the weight used in the score fusion.
The parameter $\lambda$ decides the contribution of fixations and saccades in the final decision stage.
This value can be obtained empirically. In the present work, $\lambda$ value of 0.5 is
used.

The label of the unknown sample can be obtained as
\begin{equation}
label = \mathop {\arg \max }\limits_m (score)
\end{equation}

\section{Experiments and results}

\subsection{Datasets}
The data used in this work are part of the development phase of BioEye 2015 \cite{bioeye} competition. Data recorded in three different sessions are available. First two sessions are separated by a time interval of 30 min containing recordings of 153 subjects (ages 18-43). A third session, conducted after one year, (37 subjects) is also available to evaluate the robustness against template aging. The database contains gaze sequences obtained using two distinct types of visual stimuli. In one set (RAN), a white dot moving in a dark background was used as the stimulus. The subjects were asked to follow the dot. Text excerpt shown on the screen was used as the stimulus in the other set (TEX). The samples were recorded with an EyeLink eye-tracker (with a reported spatial accuracy of 0.5 degrees) at 1000 Hz and down-sampled to 250 Hz with anti-aliasing filtering. The development dataset contains the ground truth about the identity of the persons. An additional evaluation set is also available without ground truth. 

In each recording, visual angles in $x$ and $y$ direction, stimulus angle in $x$ and $y$ direction and information regarding the validity of the samples are available. Details about the stimulus types in BioEye 2015 database are given below.

\subsubsection{Random dot stimulus (RAN\_30min \& RAN\_1year)}
The stimulus used was a white dot appearing at random locations on a black computer screen. The position of the stimulus would change every second. The subjects were asked to follow the dot on the screen and recording was carried out for 100 s.

\begin{table}[H]
\centering
\caption{\label{datasetdetails} Details about the database}
\label{my-label}
\begin{tabularx}{1\linewidth}{@{\extracolsep{\fill}}lllll@{}}
\toprule
\textbf{Dataset name}                                              & \textbf{RAN\_30min} & \textbf{RAN\_1year} & \textbf{TEX\_30min} & \textbf{TEX\_1year} \\ \midrule
Subjects                                                             & 153                   & 37                    & 153                   & 37                    \\
Stimulus                                                             & Moving dot            & Moving dot            & Text                  & Text                  \\
Duration                                                             & 100 s                 & 100 s                 & 60 s                  & 60 s                  \\
\begin{tabular}[c]{@{}l@{}}Interval between \\ sessions\end{tabular} & 30 min                & 1 year                & 30 min                & 1 year                \\ \bottomrule
\end{tabularx}
\end{table}

\subsubsection{Text stimulus (TEX\_30min \& TEX\_1year)}
The task, in this case, was reading text excerpts from the poem of Lewis Carroll ``The Hunting of the Snark''. The duration of this experiment was 60 s.

 A comprehensive list of the datasets and parameters are shown in Table \ref{datasetdetails}.
\subsection{Evaluation metrics}
The proposed algorithm has been evaluated in the labeled development set. Rank-1 accuracy and EER are used for evaluating the algorithm. Rank-1 (R1) accuracy is defined as the ratio of the total number of correct detections to the number of samples used. EER is the percentage at which false acceptance rate (FAR) and false rejection rate (FRR) are equal. Detection error trade-off (DET) curves are shown for all the datasets. Rank(\textit{n}) accuracy is the number of correct detections in the top $n$ candidates. Cumulative match characteristics (CMC) is the cumulative plot of rank(\textit{n}) accuracy. CMC curves are also plotted for all the four datasets. The evaluation set in the BioEye 2015 dataset is unlabeled. However, we report the R1 accuracy as obtained from the public results \cite{bioeye} of the competition.
\subsection{Results}
\subsubsection{Performance in the development datasets}
The model was trained using 50\% of data in the development datasets.
We have trained and tested the algorithm on completely disjoint sessions to test its generalization ability. For example, in RAN\_30min
sequence there are 153 samples available for two different sessions. We have trained the
Algorithm only on the first session (using a random 50\% subset of the data). The evaluation was
carried out on the session 2 data. We have not used the data from the same session for training
and testing since it won't account for intersession variability. 

The average R1 accuracy and EER were calculated from random 50\% subsets of development datasets. This procedure was repeated 100 times and the average R1 accuracy and EER were
obtained. The results obtained along with the standard deviations are given in Table \ref{resultsdev}.

The R1 accuracy in RAN\_30min and TEX\_30min databases are above 90\% indicating the robustness of the proposed framework. The EER on RAN\_30min database is found out to be 2.59\%, comparable to the accuracy levels of fingerprint (2.07\% EER) \cite{maio2004fvc2004}, voice recognition systems, and facial geometry (15\% EER) \cite{phillips2010frvt} biometrics.
\begin{table}[H]
\caption{\label{resultsdev} Results in the development datasets}
\begin{tabularx}{1\linewidth}{@{\extracolsep{\fill}}lllll@{}}
\toprule
& RAN\_30 & RAN\_1yr & TEX\_30 & TEX\_1yr \\ \midrule
R1 & 90.10$\pm$2.76 & 79.31$\pm$6.86 & 92.38$\pm$2.56 & 83.41$\pm$6.98 \\
EER & 2.59$\pm$0.71 & 10.96$\pm$4.59 & 3.78$\pm$0.77 & 9.36$\pm$3.49 \\ \bottomrule
\end{tabularx}
\end{table}

\begin{figure}[H]
\centering
\includegraphics[width=1\linewidth]{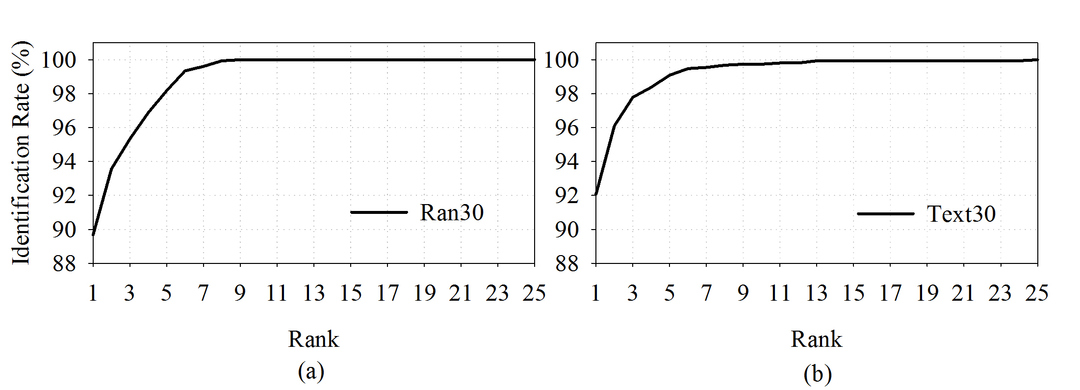}
\vspace{-1.5em}
\caption{CMC curve for (a) RAN\_30min and (b) TEX\_30min}
\label{fig:cmc_30_dev}
\end{figure}

\begin{figure}[H]
\centering
\includegraphics[width=1\linewidth]{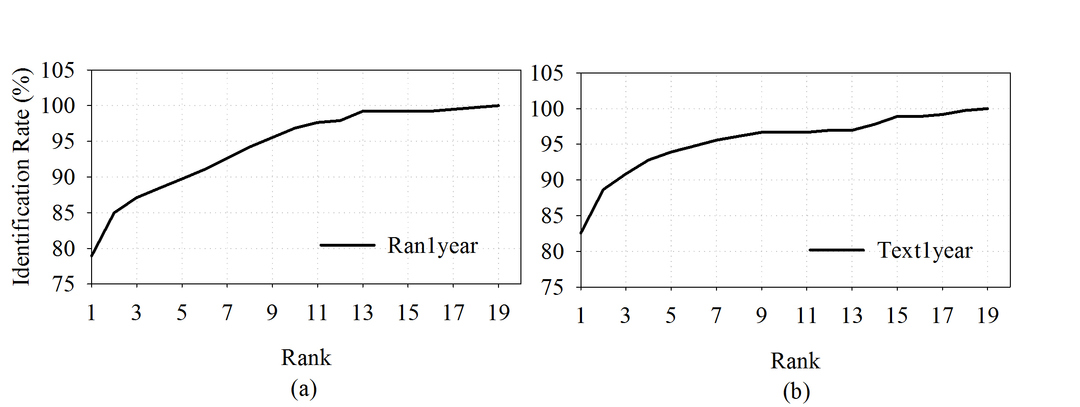}
\vspace{-1.5em}
\caption{CMC curve for (a) RAN\_1year and (b) TEX\_1year}
\label{fig:cmc_1yr_dev}
\end{figure}

R1 accuracy (Table \ref{comparedev}) of the proposed algorithm obtained from the development set was compared with the baseline algorithm (CEM-B) \cite{holland2013complexa}.
The average cumulative matching characteristics curves for the four datasets are shown in Fig. \ref{fig:cmc_30_dev} and Fig. \ref{fig:cmc_1yr_dev}.

\begin{table}[H]
\caption{\label{comparedev} Comparison of R1 accuracy in the entire development dataset}
\begin{tabularx}{1\linewidth}{@{\extracolsep{\fill}}lllll@{}}
\toprule
& RAN\_30 & RAN\_1yr & TEX\_30 & TEX\_1yr \\ \midrule
Our method (\%)& 89.54 & 81.08 & 85.62 & 78.38 \\
\begin{tabular}[c]{@{}l@{}} Baseline \cite{holland2013complexa} (\%) \end{tabular} & 40.52 & 16.22 & 52.94 & 40.54 \\ \bottomrule
\end{tabularx}
\end{table}

\begin{figure}[H]
\centering
\includegraphics[width=1\linewidth]{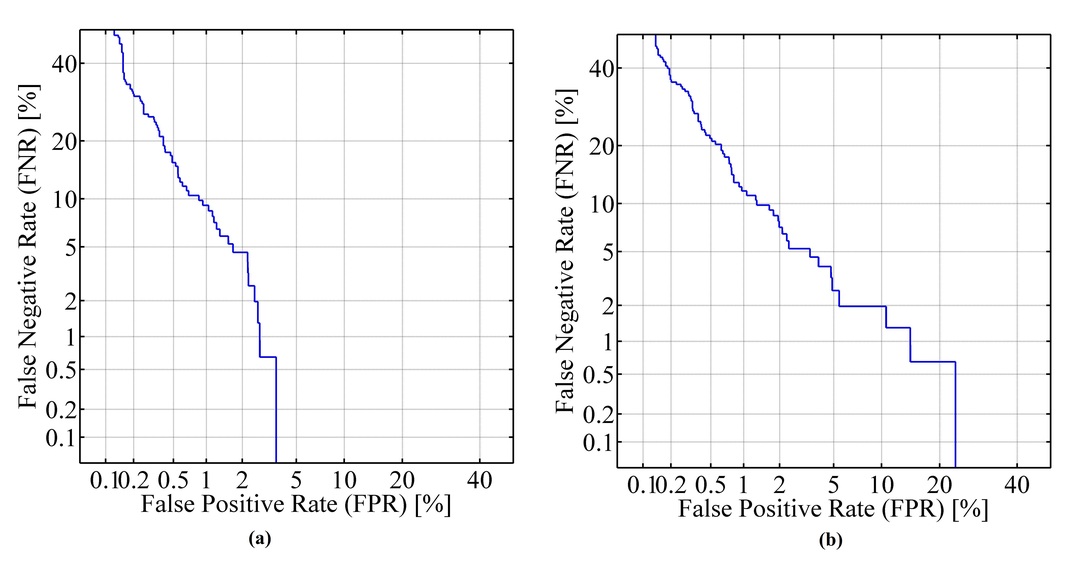}
\vspace{-1.5em}
\caption{DET curve for (a) RAN\_30min and (b) TEX\_30min}
\label{fig:eer_30_dev}
\end{figure}

\begin{figure}[H]
\centering
\includegraphics[width=1\linewidth]{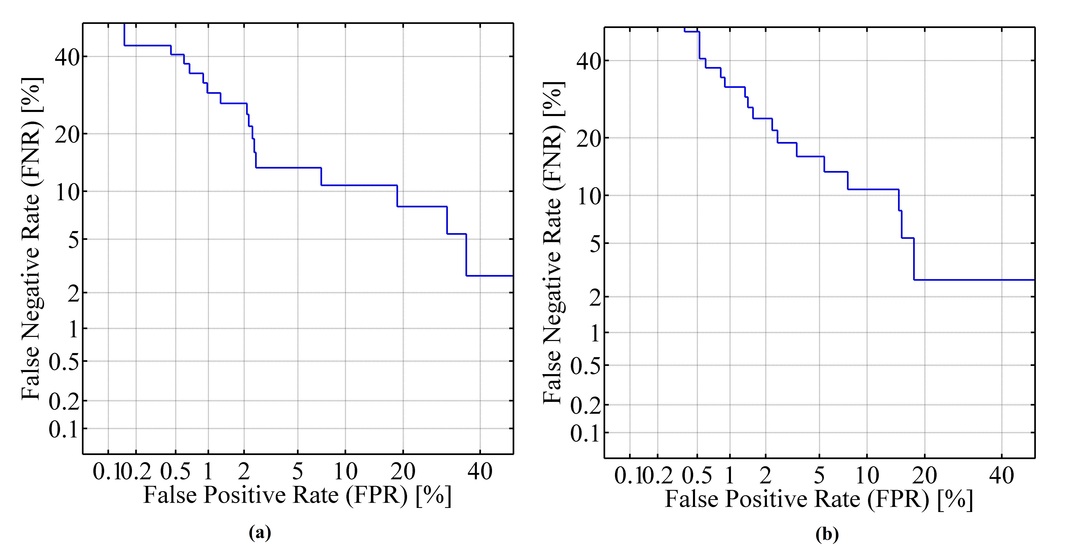}
\vspace{-1.5em}
\caption{DET curve for (a) RAN\_1year and (b) TEX\_1year}
\label{fig:eer_1yr_dev}
\end{figure}

 The detection error trade-off (DET) curves for the development datasets are shown in Fig. \ref{fig:eer_30_dev} and Fig. \ref{fig:eer_1yr_dev}. In Fig. \ref{fig:eer_30_dev} (a) and (b), FNR becomes very small as FPR increases indicating a good separation from
impostors. The reduction in FNR may be because of the addition of scores of all the fixations and saccades
in the score fusion stage. Impostor scores are considerably smaller than genuine scores in the proposed approach. The performance in
1-year sessions are poor compared to 30-min sessions indicating template aging effects.

\subsubsection{Performance in the evaluation sets}
The evaluation part of the database is unlabeled. However, the results of the competition are available on the website \cite{bioeye}.  The evaluation set of the dataset had only one unlabeled data for every labeled sample. We have used this one to one correspondence assumption in the final stage of the algorithm.

Let there be $n$ labeled and $n$ unlabeled recordings. The task is to assign each unlabeled file to a labeled file.  The scores obtained from RBF output stage were stored in a matrix $D$ (with dimension $n$x$n$).  $D(i,j)$ denotes the normalized similarity score  between ${i^{th}}$ labeled and ${j^{th}}$ unlabeled samples. We have selected the best match for each unlabeled recording using Algorithm \ref{alg:onetoone}.  The use of this one to one assumption improved the results. However, this assumption may not be suitable for practical biometric identification/verification scenarios. The proposed method has been found to outperform all the other methods even without the one to one assumption indicating the robustness for biometric applications. The results with and without this assumption are shown in Table \ref{compareeval}. 

\begin{algorithm}[H]
\caption{One to one matching}
\label{alg:onetoone}
\begin{algorithmic}[1]
\INPUT $D$ (Score matrix)
\OUTPUT Matches
\State $[n,n]=size(D)$
\For{ $i \leftarrow$ 1 \textbf{to} n} 
\State $[row,col]=find(D==max(D(:)))$
\State $D(row,:)=-\infty $
\State $D(,:col)=-\infty $
\State pair=$[row,col]$
\State Matches.\textit{append}(pair)
\EndFor
\end{algorithmic}
\end{algorithm}

\begin{table}[H]
\caption{\label{compareeval} Comparison of R1 accuracy with baseline method in evaluation dataset}
\label{my-label}
\begin{tabularx}{1\linewidth}{@{\extracolsep{\fill}}lllll@{}}
\toprule
& RAN\_30 & RAN\_1yr & TEX\_30 & TEX\_1yr \\ \midrule
\begin{tabular}[c]{@{}l@{}}Our method (\%) \end{tabular} & 93.46 & 83.78 & 89.54 & 83.78 \\
\begin{tabular}[c]{@{}l@{}}Our method\textsuperscript{*} (\%) \end{tabular} & 98.69 & 89.19 & 98.04 & 94.59 \\
Baseline \cite{holland2013complexa} (\%) & 33.99 & 40.54 & 58.17 & 48.65 \\ \bottomrule
\multicolumn{4}{l}{\textsuperscript{*}\footnotesize{With one to one assumption}}

\end{tabularx}
\end{table}

\subsection{Execution time}
The algorithm has been implemented in an Intel Core i5 CPU, 3.33 GHz desktop computer with 4 GB RAM. The average training time for the network without code optimization (single-threaded) in MATLAB is about 400 s (with 153 samples). In the testing phase, for predicting one unlabeled recording, it takes on an average 0.21 s (in TEX\_30min). The time taken for training and testing phase can be improved considerably by implementation in C, C++, using parallel processing platforms like graphical processing units (GPU).
\subsection{ Discussions}
\subsubsection{Performance of the algorithm}
The R1 accuracy of the proposed method is high in both TEX and RAN datasets, which indicates the possibility of developing a task-independent biometric system. The EER and R1 accuracy achieved show the robustness of the proposed score fusion approach. The selected features show good discrimination ability in both stimuli. The accuracy with 1-year datasets is comparatively lesser than that with the 30-min datasets.
 This lower accuracy may be attributed to template aging effects.  
 Some of the selected features may show variability over time \cite{komogortsev2014template} \cite{kasprowski2013impact}. 
 
 The amplitudes and directions of the saccades were random in the RAN dataset. This indicates that once we have a proper enrollment done, biometric identification can be performed just by using the eye movements during natural interaction conditions, even without the cooperation of subjects (as there are no restrictions on amplitude or direction). The normal eye movement during normal daily tasks can be used for the authentication purposes.

The feature selection was carried out in 30-min datasets due to the availability of a large number of subjects.  Feature selection with 1-year datasets may lead to overfitting because of fewer subjects. This issue can be solved by using the feature selection in 1-year datasets with a larger number of subjects, which may identify features that are robust against template aging.  However, the results show significant improvement compared to the state of the art methods. The proposed algorithm was ranked first in the BioEye 2015 \cite{bioeye} competition.
\subsubsection{Limitations}
Controlled experimental setup was used to collect the data used in this work. The sampling rate and quality of data used in the present work were very high since it was collected in lab conditions using chinrest. It is to be noted that the data used in this work was captured at 1000 Hz. The performance of the algorithm at lower sampling rates needs to be evaluated further. Accurate estimation of the features in noisy, low sampling rates is necessary for the use in a practical biometric scenario. The nature of eye movements may be affected by the level of alertness, fatigue, emotions, cognitive loading, etc. Consumption of caffeine and alcohol by the subjects may affect the performance of the proposed algorithm. The features selected for biometrics should be invariant to such variations. Only two sessions of data were available for each subject. Intersession variability and template aging effects need to be studied further. Lack of publicly available databases containing a large number of samples (to account for template aging, uncontrolled environment, affective states, intersession variability) is another problem. Creation of a large database with such variability could provide more robust solutions.

\section{Summary}
A novel framework for biometric identification based on dynamic characteristics of eye movements is proposed in this chapter. The raw eye movement data is classified into a sequence of fixations and saccades. We extract a large set of features from fixations and saccades to characterize each individual. The important features extracted from fixations and saccades are identified based on a backward selection framework. Two different Gaussian RBF networks are trained using features from fixations and saccades separately. In the detection phase, scores obtained from both RBF networks are used to get the subject's identity. The high accuracy obtained shows the robustness of the proposed algorithm. The proposed framework can be easily integrated into the existing iris recognition systems. Even though iris recognition technology is very accurate, it is susceptible to spoofing. A high-quality printout of an NIR iris pattern printed on a contact lens worn by an impostor can spoof the system.  Incorporating eye movement features along with iris recognition systems might make spoofing attacks impractical.  A combination of the proposed approach with conventional iris recognition systems may give rise to a new counterfeit-resistant biometric system. The comparable accuracy in distinct types of stimuli indicates the possibility of developing a task-independent system for eye movement biometrics. The proposed method can also be used for continuous authentication in desktop environments. 

\end{onehalfspacing}

%% file: Chapter7/chapter_gazeactivity_v8.tex
\chapter{Activity Recognition from Head Mounted Eye Tracker}{Activity Recognition from Head Mounted Eye Tracker}
\graphicspath{{Chapter7/pics/}}
\begin{mdframed}[linecolor=grey!3,backgroundcolor=grey!3] 

This chapter presents a framework for recognition of human activity from egocentric video and eye tracking data obtained from a head-mounted eye tracker.  Three channels of information such as eye movement, ego-motion, and visual features are combined for the classification of activities. Image features were extracted using a pre-trained convolutional neural network. Eye and ego-motion are quantized, and the windowed histograms are used as the features. The combination of features obtains better accuracy for activity classification as compared to individual features \cite{george2018recognition}.
\end{mdframed}
\vspace{5mm}

\begin{onehalfspacing}

\section{Introduction}
Activity recognition from videos is an important topic in computer vision community. Recognition of actions has several applications in many areas such as human-computer interaction (HCI), robotics, surveillance, image and video retrieval. Most of the literature in this field deals with action recognition from video streams captured by a camera which may be situated far away from the subjects (third person view) \cite{poppe2010survey},\cite{weinland2011survey}, \cite{turaga2008machine}.

 Recently with the proliferation of wearable devices, there has been an upsurge in research in the field of activity recognition from wearable devices.  Recent works in egocentric video-based (first person view) activity recognition \cite{fathi2011understanding},\cite{pirsiavash2012detecting},\cite{yan2015egocentric} has shown great promise in providing insights into various activities. The egocentric video gives direct information regarding user's environment. Head-mounted eye trackers can provide gaze locations and head movements along with the ego-centric video.  

Nowadays a lot of virtual and augmented reality (VR and AR) devices are coming up in the consumer market such as Oculus Rift, Hololens, Google Glass \cite{starner2013project}, etc. They hold the potential to augment human capabilities. Eye tracking and egocentric video could give important cues about the user's point of attention and actions. Usage of visual features along with the eye movement behavior as observed through eye tracking can lead to the understanding of activities and cognitive processes. Identification of human actions and intentions in real-time could result in human-machine systems which are more natural and `pro-active' .

\begin{figure}[t]
\begin{center}
\includegraphics[width=1\linewidth]{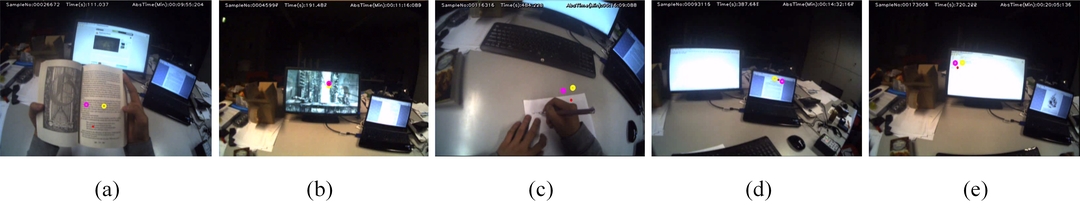}
\end{center}
\vspace{-1.5em}
\caption{The activity classes considered in the work, a) Read, b) Watching Video, c) Write, d) Copying text, and e) Browsing.}
\label{fig:samples}
\end{figure}

In this chapter, a framework for activity classification using egocentric information obtained from a head-mounted eye tracker is presented. Three channels of information, namely, eye movement patterns, ego-motion patterns and visual features as observed through the camera, are used for activity classification.  We consider activities performed in office environments which are difficult to classify by other modalities alone. Combining all these modalities can improve the accuracy of classification. The activity classes used in this work are shown in Fig. \ref{fig:samples}.

\section{Related works}

An excellent review of recent works in egocentric activity recognition can be found in \cite{nguyen2016recognition}.  Some of the recent works related to activity recognition from eye gaze are described here.

Bulling \textit{et al}.\cite{bulling2011eye} presented an activity recognition scheme based on eye movement parameters obtained using Electro Oculogram (EOG). They extracted a large number of features from fixations, saccades, and blinks. A feature selection approach was used to select the best features for activity classification. They considered five activities performed in the office environment, along with a null class. A support vector machine based classification was adopted for recognizing the activities. This work paved the way for further investigations using eye gaze where activity recognition using other modalities are difficult. Hipiny and Mayol-Cuevas \cite{hipiny2012recognising} presented an activity classification scheme using the gaze data. They represented each activity as a record of fixation locations. A Bag of words based weighted voting scheme, along with the Bhattacharya distance between templates and samples were used for classification. Ogaki \textit{et al}.\cite{ogaki2012coupling} presented an approach for egocentric activity recognition by fusing eye movement and ego-motion features.  They estimated ego-motion from the global optical flow computed from the “outward looking” camera. The eye tracking data was obtained from a head-mounted eye tracker. Both eye motion and ego-motion parameters were encoded to a string sequence using the motion pattern. The ‘N-gram’ statistics, computed over a sliding window, was used as a feature for classification.  From the experiments, they demonstrated that the combination of features improves the accuracy compared to eye movement features alone. Li \textit{et al}.\cite{li2015delving} presented a novel scheme for combining different modalities of information for egocentric action recognition. From the egocentric video, they extracted dense trajectories and a set of local descriptors across the trajectories. The features included motion binary histograms along $x$ and $y$ directions, histogram of flow, histogram of gradients and Lab color histogram. They computed these features within a grid, and the features were then concatenated. Egocentric features such as head motion and hand manipulation point were also extracted. They encoded the features using Improved Fisher Vector (IFV). Finally, the IFVs of different features were concatenated as a representation of the video. Support vector machine (SVM) was used for classification. However, they did not use eye movement patterns in their framework. Fathi \textit{et al}.\cite{fathi2012learning} demonstrated the relation between the task being performed and the locations of visual attention.  They showed that the information regarding hand-eye coordination could be beneficial in two different scenarios, predicting the probable gaze sequence given an action and predicting the likely action given the gaze sequence. Shiga \textit{et al}.\cite{shiga2014daily} proposed a method for egocentric activity recognition by combining eye motion and visual features.  The eye movement feature extraction scheme was similar to the method used in \cite{bulling2011eye}. They used ‘N-gram’ statistics computed over sliding windows. The visual features were obtained by selecting a patch around the gaze location and extracting local features using SIFT-PCA and dense sampling. A Bag of words approach was used for the classification. They trained separate multi-class SVMs for visual and eye movement features, and score fusion methodology was adopted for the final activity classification. Yan \textit{et al}.\cite{yan2015egocentric} proposed a multi-task clustering approach for egocentric activity classification.  They proposed two different algorithms for activity classification in unsupervised settings. Kunze \textit{et al}.\cite{kunze2013activity} provided a description of possibilities of eye tracking in various use cases such as detection of fatigue and reading. Data from mobile eye trackers can be utilized for the analysis of reading habits, type of document read, reading speed comprehension level and identifying alertness levels.

While there are many approaches for activity classification in egocentric videos, classification in indoor environments is still a challenge. This can be mainly attributed to the lack of significant motion patterns and limited variations in the environment. In most of the office activities (like reading, copying, browsing, watching a video, writing ), the variability in image background, as observed from the egocentric video is limited. This yields poor accuracy due to the lack of sufficient discriminative information. However, a fusion of these features could improve the performance. The visual features can provide a context for the action, and the combination of ego-motion and eye movement pattern can result in better accuracy in the overall classification.

\section{Proposed method}

\begin{figure}[!ht]
\begin{center}
\includegraphics[width=1\linewidth]{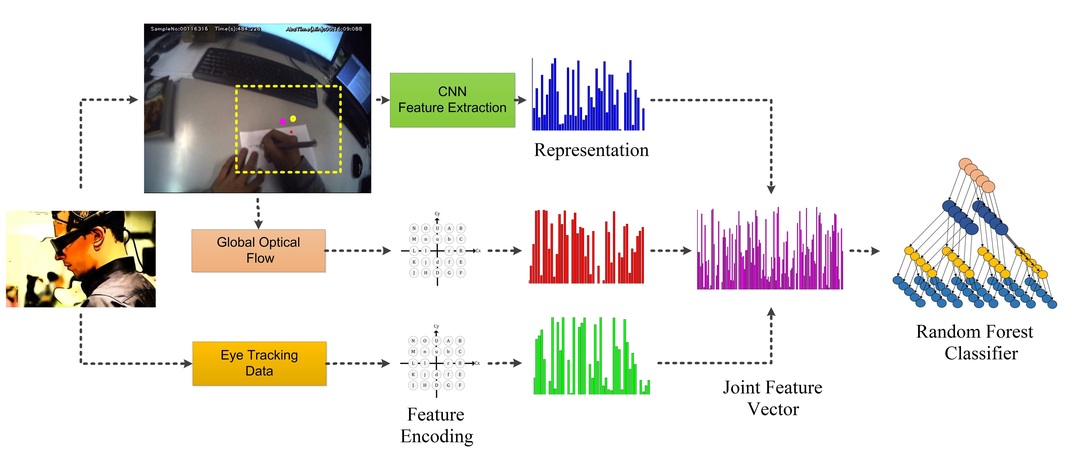}
\end{center}
\caption{The proposed framework, three channels of information are fused to classify the activities.}
\label{fig:framework}
\end{figure}

In this work we propose to use information from the image, gaze locations and ego-motion for the recognition of activities. The features extracted from each domain along with the proposed fusion scheme is described below. A schematic diagram of the proposed framework is shown in Fig. \ref{fig:framework}.

\subsection{Feature extraction from image}

\begin{figure}[h]
\begin{center}
\includegraphics[width=1\linewidth]{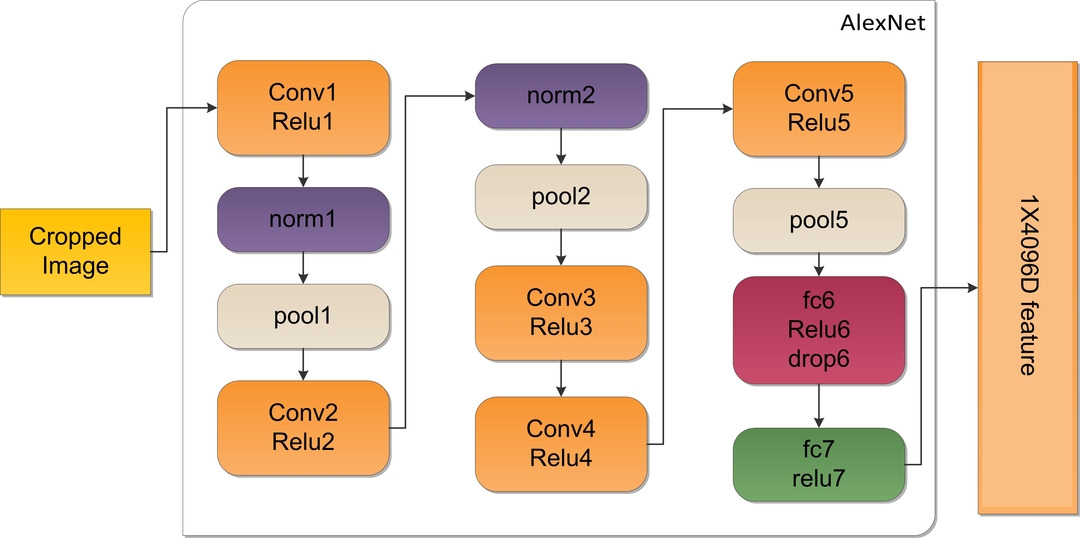}
\end{center}
\caption{CNN feature extraction scheme, cropped and resized image is fed into the pretrained network, outputs from \textit{fc7} are used as the feature.}
\label{fig:alex}
\end{figure}

Location of gaze on the images captured from a first person view (ego-centric) cameras carries valuable information which might be useful for activity classification. Previous works \cite{shiga2014daily} have used dense SIFT descriptor with PCA in a Bag of words (BoW) framework. Features were extracted from the patch around the point of gaze.  They computed the descriptors for each frame separately. The accuracy of this method could fall when the training and testing environments are different. For example, the appearance of a book might differ with variations in size, pose, color, and different types of binding. Ideally, the feature representation should be invariant to such changes as it is intended to give a context to the actions. We have used convolutional neural network \cite{sharif2014cnn} based feature extractor in this work owing to its high representation power. A pre-trained Alexnet model \cite{krizhevsky2012imagenet} (trained on the Imagenet dataset) is employed for this purpose. The fully connected output layer was removed, and a feature descriptor of dimension 4096 was obtained. The architecture of Alexnet excluding the final fully connected layer is shown in Fig. \ref{fig:alex}. We take the output from \textit{fc7} layer after applying the rectified linear unit (ReLU) transformation \cite{gong2014multi}.
For each image in the training set, a patch of size $200 \times 200$ was selected around the gaze location. The image patch obtained was resized and fed to the CNN to obtain a 4096-dimensional feature vector. We have extracted features from all the images in the training set in a similar manner. K-means clustering was performed on this data, and 15 cluster centers were kept. Now, for each image, the feature representation is computed, and the cluster center closest to it is found out. Histogram Voting across the cluster centers are carried out, and the normalized votes are computed in a temporal window of 25 seconds. The histogram obtained is used as the feature input for the activity classification.

\subsection{Feature extraction from eye tracking data}

The eye movement sequence is of the form
\begin{equation}
E = \left\{ {{e_{x,t}},{e_{y,t}}} \right\}_{t = 1}^{{T_E}}
\end{equation}

where,  ${e_{x,t}},{e_{y,t}}$ denote the $x$ and $y$ components of gaze position at the time instant $t$. $T_E$ denotes the duration of the sequence. The raw sequence is median filtered to remove noise.  Let ${e_{x,t}}$ be the input signal corresponding to the $x$ component of eye movement. The wavelet coefficient $Cx_b^a$
of ${e_{x,t}}$ at scale $a$ and position $b$ is defined as

\begin{equation}
Cx_b^a = \int\limits_\Re  {{e_{x,t}}{1 \over {\sqrt a }}} \overline {\psi \left( {{{t - b} \over a}} \right)} dt
\end{equation}

Continuous 1D wavelet coefficients are computed at a scale 10 using Haar-wavelet function.

Now, the wavelet coefficients are computed separately for $x$ and $y$ directions. The coefficients obtained are quantized as
\begin{equation}
 \hat Cx_b^a = \left\{
\begin{array}{ll}
      2 & {\tau _{large}} \le {Cx_b^a} \\
      1 & {\tau _{small}} < Cx_b^a \le {\tau _{large}} \\
      0 & { - {\tau _{small}} \ge Cx_b^a \le {\tau _{small}}} \\
      -1 & { - {\tau _{large}} < Cx_b^a \le  - {\tau _{small}}} \\
      -2 & {Cx_b^a \le  - {\tau _{large}}}  \\
\end{array}
  \right.
\end{equation}

where, $\tau _{large}$ and $\tau _{small}$ are empirically decided thresholds.

\begin{figure}[h]
\begin{center}
\includegraphics[width=0.6\linewidth]{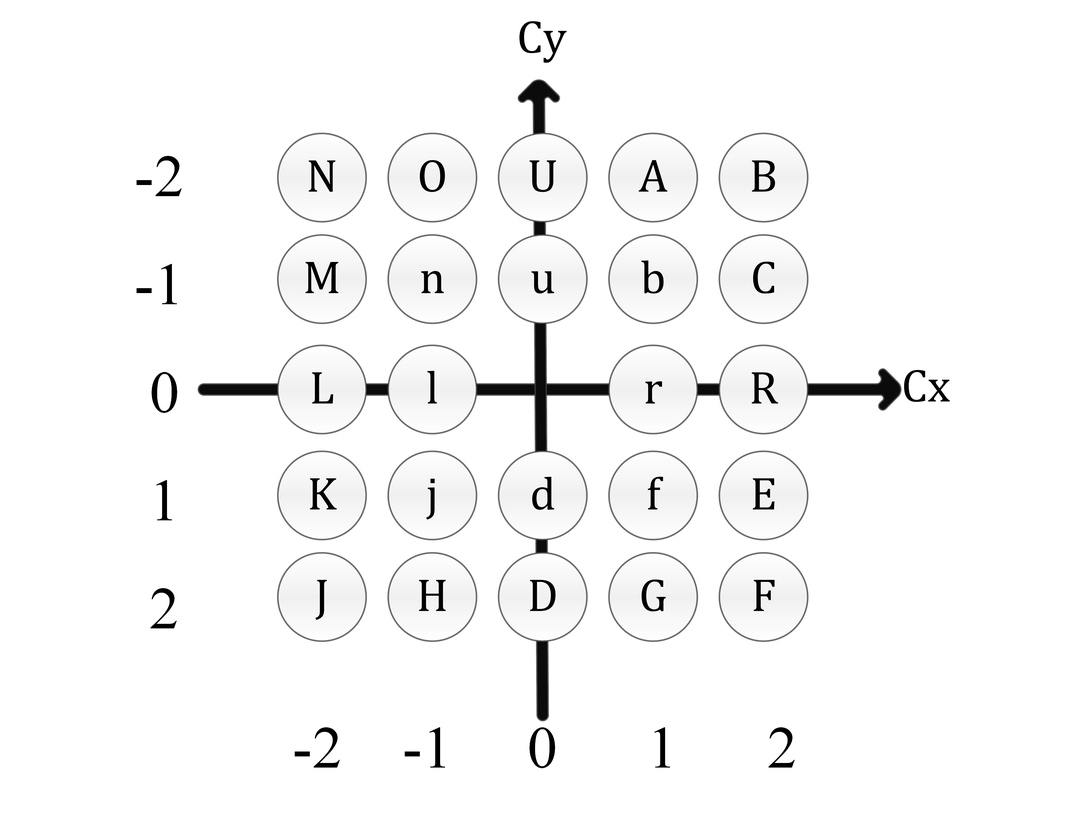}
\end{center}
\caption{Motion encoding scheme.}
\label{fig:motionencode}
\end{figure}

$Cy_b^a$ is also quantized to $\hat Cy_b^a$ in a similar manner.

Based on the joint sequence $(\hat Cx_b^a,\hat Cy_b^a)$, a string sequence is generated as in Fig. \ref{fig:motionencode}.

The normalized histogram of the string sequence over a sliding temporal window is used as the feature for classification.

\subsection{Feature extraction from motion}

Motion features are extracted from the optical flow between subsequent frames.
Let the $i^{th}$ frame be denoted as $F_i$. For each frame, corner detection is performed to obtain the candidate points 
to track. The points are tracked using Lucas-Kanade optical flow. Successfully tracked points are found out using forward-backward error \cite{kalal2010forward}. The median flow between the frames can be computed as
\begin{equation}
\Delta x = median(\delta {x_j}),\,j \in [1,K]
\end{equation}
\begin{equation}
\Delta y = median(\delta {y_j}),\,j \in [1,K]
\end{equation}

Where $K$ is the number of sparse points tracked between $F_i$ and $F_{i+1}$, and $\delta {x_j}$ and $\delta {y_j}$ denote the optical flow of $j^{th}$ point in $x$ and $y$ direction respectively.

Once the global optical flow is obtained, we use a similar encoding scheme as used for eye gaze data. The histogram of the encoded sequence obtained over a temporal window is used as the feature for the classification task.

\subsection{Fusion and classification framework}

Features obtained from the three independent modalities namely ego-motion, eye gaze features and visual features are combined in the proposed approach. Feature level fusion \cite{ross2005feature} is adopted where three modalities are concatenated to form the final feature vector. We have extracted all the features using a temporal sliding window of 25 seconds with a stride of one second. Histogram of each independent feature is computed and concatenated for training the classifier model.

The classification model chosen should be able to handle different types of data as inputs. We have chosen Random Forest (RF) Classifier for this task. Random forest algorithm is an ensemble of decision trees initially proposed by Breiman \cite{breiman2001random}. It can intrinsically handle multi-class classification problems.  Instead of using a single tree for classification, predictions from a large number of trees are integrated to form the final prediction. Different trees in the forest are trained from bootstrap samples. The original data is sampled with replacement and trees are trained using these bootstrap samples. For each tree, a subset of predictors are randomly selected at each node and an optimal split is found \cite{ma2005classification}. The tree is grown without pruning.  In the testing phase, the test sample is fed to $N$ trees in the forest. Each tree makes a prediction by evaluating the decision tree. The final prediction is obtained using voting strategy among the outputs of $N$ decision trees. Random forest is robust to noise and faster to train. RF gives better predictions without overfitting due to the out of bag error cross-validation used during the training.  

\section{Experiments and results}

 Activities performed in office environments are considered in the experiments as they are difficult to classify by other methods. We have evaluated the accuracy of individual features as well as joint representation in a multi-class scenario to assess their performance.

\subsection{Database used}
We have used UTokyo First-Person Activity Recognition Dataset \cite{ogaki2012coupling} for the evaluation task.  The dataset contain the data recording of five subjects performing five different actions in an office environment. The classes available were reading a book, watching a video, copying text, writing on paper, and internet browsing.  Each of these activities was performed for two minutes. There was a time gap of thirty seconds (`Void' class) between each activity where subjects were allowed to converse, sing and move freely. Each subject performed the activities twice. The data from these two sessions were used as the training and test sets.  The recordings obtained from EMR-9 eye tracking device was also available with the dataset. For analysis purpose, we have used the eye tracking data and the low-resolution video ($640 \times 480$ resolution) from the dataset.

\subsection{Experiment protocol}
For each subject in the dataset, the features corresponding to visual, eye movement and ego-motion were extracted from the dataset.  For each subject, two separate instances of the same activity class are available. We have used these two folds for the evaluations. Initially, the first fold was used for training and the second one for testing. In the second fold, training and testing sets were interchanged and the average accuracy computed across these two sets are reported. The evaluations were performed for a multi-class scenario, data across all subjects were used for training and testing.

\subsection{Multi-class classification}
 We have analyzed the performance with two different scenarios namely five class and six class classification. In the latter, `Void' class is also used as a valid label.

\subsubsection{Experiments with five activity classes}

\begin{figure}[!htb]
\begin{center}
\includegraphics[width=1\linewidth]{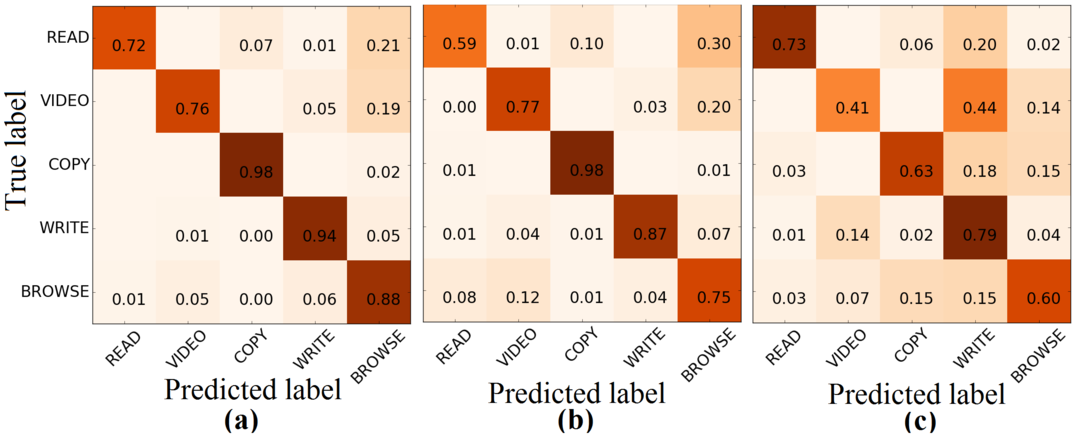}
\end{center}
\caption{Normalized confusion matrix for five classes, a) Combined features, b) Joint Ego-Eye motion feature, c) Visual features }
\label{fig:conf_fiveclass}
\end{figure}

We have used five activity classes in this trial. Experiments were performed in multiclass classification scenario to evaluate the generalization capability of the features. Training and testing were done across all the individuals. The first session data from all the subjects were used for training. A Random Forest model was trained using the joint feature vector obtained from ego-motion, eye motion, and CNN features. The individual accuracy of the modalities was also tested by training separate models for CNN as well as joint eye-ego motion features.  The experiment was also performed by interchanging the training and testing sets. The average results among these two folds were found. The normalized confusion matrix obtained is shown in Fig. \ref{fig:conf_fiveclass}. The individual confusion matrices for visual and motion features alone are also shown in Fig. \ref{fig:conf_fiveclass}. The average accuracy over multiple runs is shown in Table. \ref{tab:avg_accuracy}.

\subsubsection{Experiments with six activity classes}

\begin{figure}[!htb]
\begin{center}
\includegraphics[width=1\linewidth]{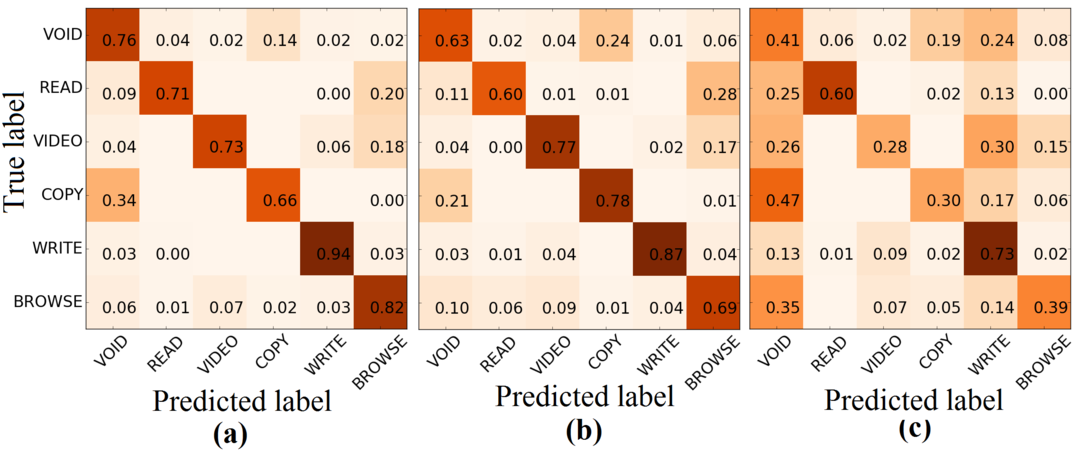}
\end{center}
\caption{Normalized confusion matrix for six classes, a) Combined features, b) Joint Ego-Eye motion feature, c) Visual features }
\label{fig:conf_sixclass}
\end{figure}

In this experiment, we have considered all six classes including the 'Void' class. We have followed similar testing methodology as described for five class scenario. The results obtained are shown in Fig. \ref{fig:conf_sixclass} and Table \ref{tab:avg_accuracy}.

\subsubsection{Accuracy across different subjects}
\begin{figure}[H]
\begin{center}
\includegraphics[width=0.7\linewidth]{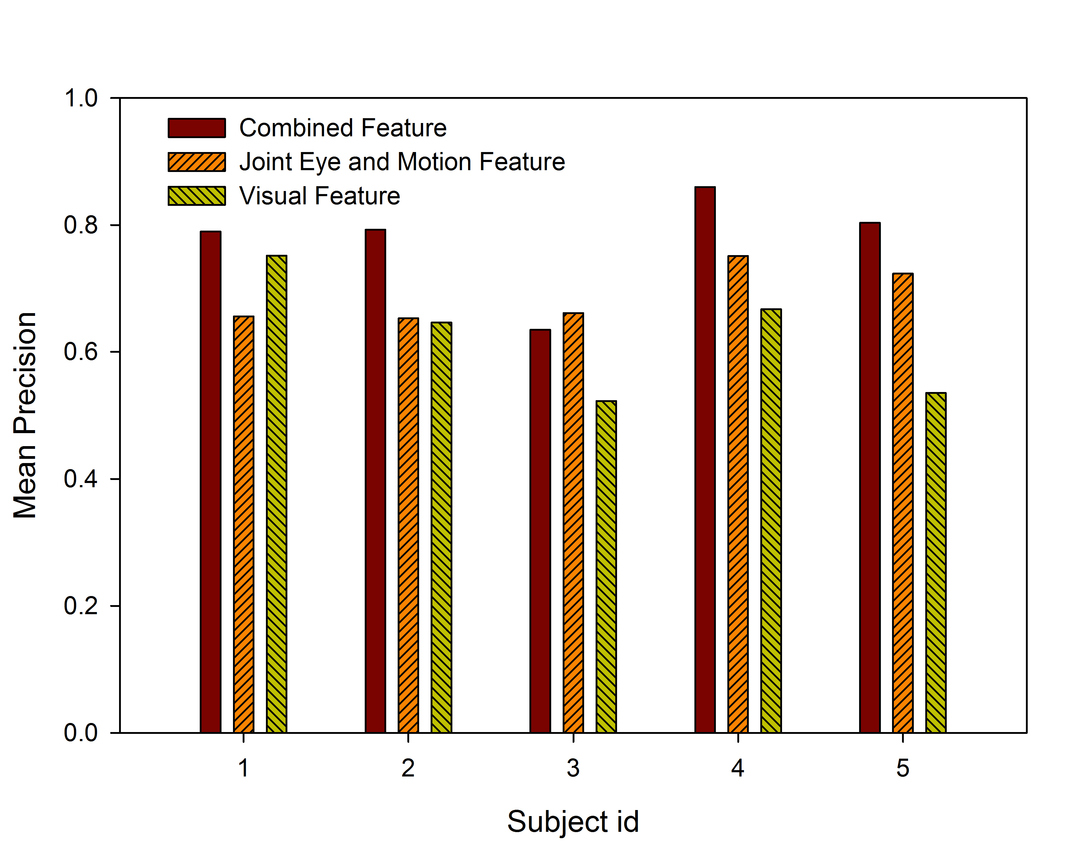}
\end{center}
\caption{Variation of accuracy across different subjects}
\label{fig:acc_subjects}
\end{figure}

The variations in accuracy across different subjects are shown in Fig. \ref{fig:acc_subjects}.
The combined feature gives better results for most of the subjects. 
\begin{table}[H]
\centering
\caption{Average accuracy for all three for both 5 and 6 class scenarios}
\label{tab:avg_accuracy}
\begin{tabularx}{1\linewidth}{@{\extracolsep{\fill}}llll@{}}
\toprule
Classes & \begin{tabular}[c]{@{}l@{}}Combined\\  Feature\end{tabular} & \begin{tabular}[c]{@{}l@{}}Eye and Ego Motion\\ Feature\end{tabular} & \begin{tabular}[c]{@{}l@{}}Visual  (CNN)\\ Feature\end{tabular} \\ \midrule
6 class & \textbf{77.09\%  }                                                   & 72.49\%                                                              & 45.03\%                                                         \\
5 class & \textbf{85.65\% }                                                    & 79.38\%                                                              & 62.97\%                                                         \\ \bottomrule
\end{tabularx}
\end{table}

\subsubsection{Accuracy across classes}

\begin{figure}[H]
\begin{center}
\includegraphics[width=0.7\linewidth]{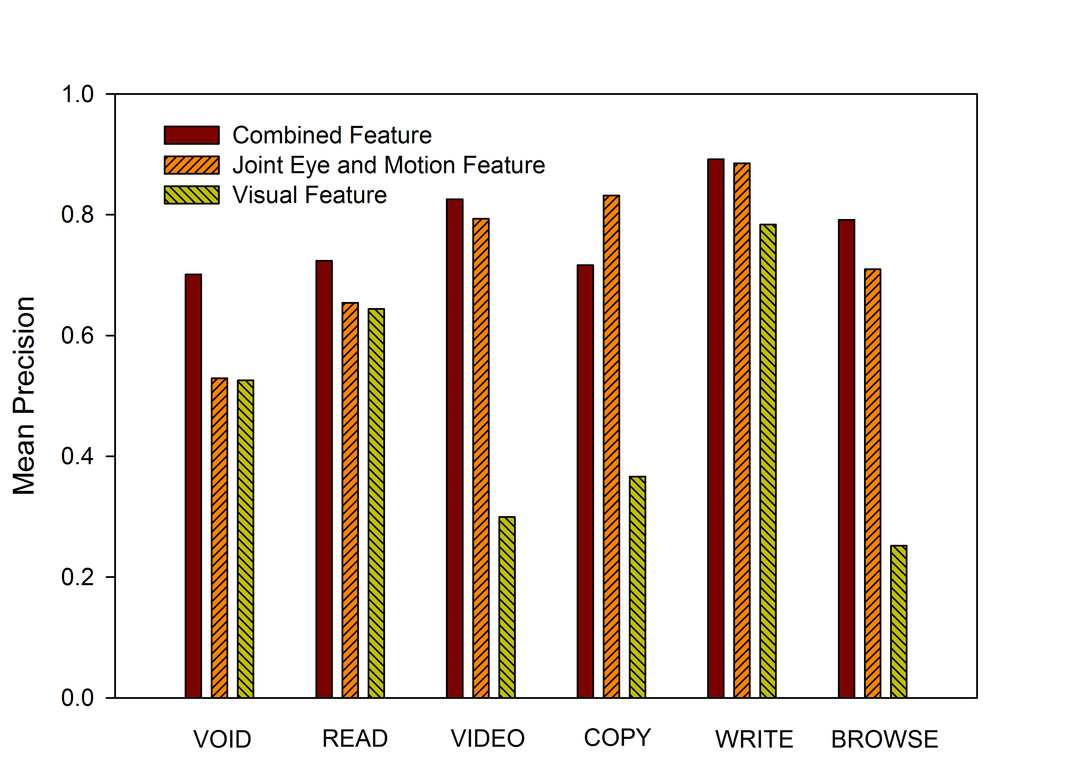}
\end{center}
\caption{Variation of accuracy across different classes}
\label{fig:acc_classes}
\end{figure}

The accuracy of different classes for different feature combinations are shown in Fig. \ref{fig:acc_classes}. The joint representation achieves better results as compared to the individual features. The joint eye-ego motion feature obtains the best accuracy among the features. The `Void' class shares similar visual features and motion features as the subjects were allowed to interact freely during those periods. This could explain the low accuracy of the `Void' class. Visual features give good results in activities like `Write' and `Read' since the field of view is different from other activities. 

\subsection{Comparison with other methods}
\begin{figure}[H]
\begin{center}
\includegraphics[width=0.8\linewidth]{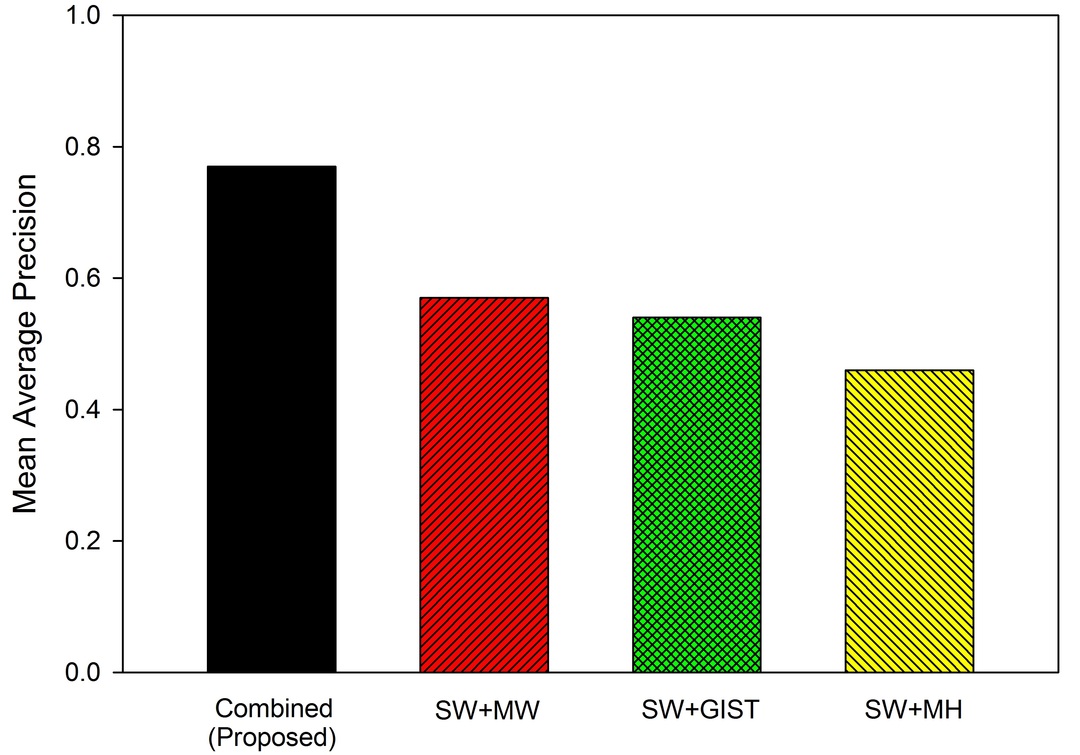}
\end{center}
\caption{Comparison with state of the art methods \cite{ogaki2012coupling}, SW+MW (Saccade Word+ Motion Word) \cite{ogaki2012coupling}, MH (Motion Histogram) \cite{kitani2011fast}, GIST \cite{oliva2001modeling}}
\label{fig:comparison}
\end{figure}

We have compared the results obtained with different methods. Saccade word and motion word (SW + MW) \cite{ogaki2012coupling}, which is a combination of eye movement and egomotion `N-gram' features, obtains the second best result.
GIST features \cite{oliva2001modeling} extracts the visual content of the scene can be used for activity recognition in egocentric video \cite{spriggs2009temporal}. A combination of saccade word (SW) and GIST effectively combines motion and visual features. 
Motion histogram (MH) proposed by Kitani \textit{et al}.\cite{kitani2011fast} encodes the instantaneous as well as period motion using Fourier analysis. The accuracy of saccade word and motion histogram is also taken for comparison. The mean average precisions of the methods are compared in Fig. \ref{fig:comparison}. 
 The proposed method outperforms all the other methods. The addition of visual features along with the motion and eye gaze features improved the accuracy significantly. Compared to other methods, the higher representation power of the CNN based feature and the combination of ego-eye motion features makes the algorithm more accurate.

\subsection{Discussions}

 From the results obtained, it can be seen that the addition of three modalities improves the accuracy. In six class scenario, highest accuracy is achieved for class `Write'.  This can be attributed to both distinct gaze patterns as well as visual features during the activity. Especially, the high accuracy of visual features during this activity may be due to the appearance of paper and pen which are unique to this activity.  Even though the addition of visual features increases the overall accuracy, the individual performance of visual features in many cases are poor. The activities used in this experiment were performed in an office environment, which does not have much diversity in visual information. The addition of the ‘Void’ class introduces more errors as the same visual features appear in multiple activities.

In the five class scenario, `Void' class was not present. The accuracy of visual features is much better than the six class case. The overall accuracy of classification is also much better in this scenario.  The random forest based classifier tries to identify the important features for activity classification from the joint feature representation.

Some of the advantages of the proposed system are described here. Three distinct channels of information are fused in the proposed approach. This improves the generalizability of the approach for a larger number of classes. Representation of one particular activity might not require the features from all three channels. For example, reading has a characteristic pattern as observed from eye tracking data (sequence of small fixations and saccades), It may be possible to identify reading activity from eye tracking data alone. Classifying browsing activity from watching movies might require all three channels of information. The high-level CNN descriptors used are suitable for giving a context to the actions. The random forest algorithm is capable of identifying important features which are relevant for the identification of a particular action. The framework can determine the important features required for classifying the activities accurately.  Even though the activity classes used in this work are small, the framework is capable of handling a large number of classes. The Random Forest based classifier can compute the features relevant for identifying each action.

\section{Summary}

In this chapter, we have proposed an approach for combining different modalities such as ego-motion features, eye movement features and visual features for classification of activities. A joint feature vector is formed from the individual feature extractors, and a random forest classifier was used to classify the activities using this joint representation.  Joint eye-ego motion feature gave the best individual accuracy among the features. However, the addition of visual feature resulted in a higher accuracy in activity classification. Additional channels of information can be easily added to the framework. The addition of activity-dependent object detectors and a weighted fusion of these three modalities might improve the results.

\end{onehalfspacing}

%% file: Conclusion/conclusion_v10.tex
\chapter{Conclusion and Future Scope}{Conclusion and Future Scope}
\begin{onehalfspacing}
\section{Conclusions}

This thesis presents the development of gaze tracking algorithms and their applications in specific areas.  The first part of the thesis deals with the development low-cost eye gaze tracking algorithms for desktop as well as head mounted cameras. In the second part, two applications which leverage the eye tracking data is developed.

 A webcam-based system was developed to bring down the cost of gaze tracking while maintaining reasonable speed and accuracy. The challenging problem of iris center localization is solved using an efficient two-stage algorithm.  The main contribution is the simplification of the ellipse fitting problem with a rather simple two stage scheme using appropriate constraints obtained from the face detection stage. Another advantage here is that small errors in the first stage can be refined in the second stage.  The algorithm is implemented as a convolutional kernel which improved its real-time performance.  The iris center estimation stage has been extended to a gaze tracking framework using eye corner detection and tracking. Pose variations due to in-plane rotations are handled using an affine transformation of the estimated gaze in the calibration plane. The approach developed achieves real-time performance in desktop environments.

Further, a pupil center localization algorithm is developed for head-mounted eye trackers.  The performance of most of the available algorithms deteriorates with uncontrolled lighting conditions.  We have developed a robust algorithm for pupil localization in NIR images which works even in challenging conditions. The algorithm uses either edge based method or grayscale intensity based on the quality of the image. One significant advantage here is that, even if the first edge based stage of the algorithm fails due to some reflections, the second stage can identify the pupil (though more computation is required). Further, the tracking approach reduces false detection rate at the same time reduces computational load as the search space is considerably less. Most of the algorithms are designed to maximize the per frame based detection rates. Here, we have added a simple tracking framework which directly extends the algorithm for video.  The algorithm had been evaluated in labeled pupils in the wild (LPW) dataset and found to outperform state of the art methods while achieving real-time performance. The eye gaze position obtained from such a gaze tracker can be used for various HCI applications in real-world scenarios.

 Classification of gaze direction is useful in various HCI tasks. Further, it can be used to identify eye accessing cues thereby allowing us to infer various cognitive processes. The proposed algorithm leverages the information obtained from the facial landmark detection stage to limit the computation to eye regions.  The alignment using eye corners reduces the intra-class variability. Score level fusion from the two separately trained CNN's further improve the accuracy.  The proposed approach achieved superior performance compared to the classical gaze direction classification methods while obtaining real-time performance.

We have developed algorithms for two applications where eye tracking data is useful. In the first application, we use eye movement pattern as a biometric modality. A framework for biometric identification based on eye movements is developed in this work.  A score level fusion approach using a novel set of features extracted from fixations and saccades are used for biometric authentication.  Most of the features used were extracted from the position, velocity and acceleration profiles of eye movements. Eye movements are generated from a complex oculomotor plant, however estimating the parameters of the model directly is difficult. In this work, we tried to characterize each individual based on the statistical properties of saccades and fixations of different amplitudes and direction. Important features were identified using a backward selection framework, and Radial basis function network was used for classification.
The developed framework is evaluated in BioEye 2015 dataset and was found to outperform state of the art methods. The proposed approach obtained an average EER of 2.59\%. Even though the method can be used as an independent modality, augmenting eye movement biometrics with conventional iris recognition technology may lead to a counterfeit resistant biometric modality with inbuilt liveliness detection and continuous authentication capabilities. The proposed eye movement based biometrics can complement the iris based authentication.  Iris based authentication is one of the most feasible and accurate biometric modality available today (with EER close to 0.0011). However, it is susceptible to spoofing. A high-quality printout of an NIR iris pattern printed on a contact lens worn by an impostor can spoof the system.  Incorporating eye movement features along with iris recognition systems might make spoofing attacks impractical. The dynamics of eye movements are very fast and complex, which makes it very difficult to replicate with any mechanical systems.

The second application is the identification of human activities from a head-mounted eye tracker. Head mounted eye trackers can provide the gaze locations along with the ego-centric video.  The information from various eye movements and ego-motion are encoded by quantizing the motion. Image features are computed from an image patch centered around the gaze point. A convolutional neural network based descriptor is used as the feature. Three distinct channels of information are fused in the proposed approach. This improves the generalizability of the approach for a larger number of classes. Representation of one particular activity might not require the features from all three channels. The high-level CNN descriptors used are suitable for giving a context to the actions. The random forest algorithm is capable of identifying important features which are relevant for the identification of a particular action. The proposed approach obtained better accuracy as compared to state of the art methods.

\section{Limitations and Future scopes}

The proposed webcam based gaze estimation framework uses eye corners as the reference point. This method could fail with large head-pose variations after the calibration stage. Since the computational overhead from the IC localization stage is small, complex 3D model based tracking can be employed for pose invariant gaze estimation.
The algorithm proposed for pupil detection in the head mounted trackers returns the parameters of the ellipse fitted to the pupil boundary. This ellipse can be back-projected to determine the variations in pupil diameter. Further, pupil diameter estimation and its analysis can be useful in identifying affective and alertness states. The proposed algorithm currently uses a simple tracking mechanism, however, model-based tracking with explicit eye movement type identification could improve the performance.

The performance of the proposed eye movement biometrics framework could be enhanced with score normalizations from different samples. Feature selection with a larger dataset collected in various sessions could alleviate the template aging problem. Estimation of the features accurately from noisy, low sampling rate eye tracking data is another path to be explored.

The activity recognition framework works well for indoor environments. More robust image-based recognition algorithms can be added for improving the recognition from the video. Additional channels of information can be easily included in the framework. Further, it is possible to tune the contribution of each feature modality for a particular task. For example, for the classification of activities in the outdoor sports, ego-motion and visual features might be more helpful. Whereas in an office environment, ego-motion and eye movement might be more discriminative.

Understanding driver behavior with eye tracking and first person video could be a useful future direction in this respect. The eye movements of the driver in response to various real world situations, traffic signs, and pedestrians can be helpful in gauging the alertness level of the driver. The addition of the visual information and the visual understanding obtained from the CNN based descriptor (context) could make it possible anticipate and understand the driver's actions. Eye movements during driving conditions can also be used to gauge the expertise level of a driver.

Development of systems / software which can be used for the particular task might help eye tracking technology become ubiquitous. End to end software packages for e-learning, assistive systems, fatigue detection, biometric identification, stress detection, disease diagnosis, and  advanced user interfaces are few areas which might be benefited from the eye tracking technology.

Some of the possible extensions of the current work are summarized below

\begin{itemize}
\setlength\itemsep{0em}
\item {Extension of the gaze estimation framework using full 3d model}
\item {Addition of pupil diameter estimation along with pupil center localization}
\item {Addition of score normalization and feature selection in eye movement biometrics framework}
\item {Extension of eye movement based activity recognition with more detailed visual descriptors}
\item {Implementation of end to end systems for eye tracking applications}
\end{itemize}

Eye tracking and information available from eye movements could play a major role in improving human-computer interaction. Eye gaze provides a natural interaction channel for the immersive 3D environment in virtual and augmented reality devices. The gaze tracking framework developed can be further extended to the analysis of pupil diameter variations and saccadic velocity, which can be used for the estimation of affective and cognitive states. Eye movement biometrics can be utilized as an independent biometric modality or can be used along with conventional iris recognition systems for a counterfeit resistant authentication system. 

Activity recognition from eye gaze tracking holds potential in applications where classification using other modalities fails. Especially classifying actions performed in office environments where visual or head motion cannot help. However, more involved methodologies for including gaze as an active cue in the use of visual feature extraction could be employed.  In addition to these, parameters related to eye movements could be helpful in finding out alertness level, fatigue, emotional states, disease diagnostics, etc. The combination of all these features might lead to a system which can have large implications for human-computer interaction, lifelogging, context-aware HCI, etc.

Eye tracking provides a rich amount of information regarding users identity, stress levels, some class of diseases, user actions, alertness level, and several other parameters.  It can also be used as an HCI channel. In this context, eye tracking technology holds the potential to become a universal tool. With the advancements in the development of low-cost eye trackers as well as innovative applications, the opportunities are endless. We hope that in near future eye tracking technology can be used as a channel for intelligent HCI, where the machine can understand the intentions, actions, and identity of the user and the interact intelligently.

\end{onehalfspacing}

%% file: Appendix/AppendixI_v3.tex
\chapter{Appendix}{Face Detection and Tracking Framework}
\graphicspath{{Appendix/}}

\begin{onehalfspacing}

\section{Introduction}
 Face detection is an important stage in many computer vision applications like face recognition, facial expression analysis, and gaze tracking. Several methods have been proposed in the literature for face detection. Among these methods, the use of Haar-like features is found to be quite robust. The direct application of Viola-Jones approach has certain disadvantages such as 1) Computational overload while using for real-time applications, 2) detects only frontal faces and 3) Does not use temporal information in video sequences. We have used a simple but effective approach to solving these issues.

\section{The Algorithm}

\textbf{Haar-like Features}\\
Haar-like features \cite{viola2001rapid} are features similar to Haar wavelets used in image processing applications for fast computation of features. An algorithm for face detection using Haar-like features was first developed by Viola \textit{et. al. }\cite{jones2001rapid} and was later extended by Lienhart \textit{et. al.}\cite{lienhart2002extended}. The implementation of the algorithm was fast since the features could be obtained easily once the integral image is computed. The algorithm uses a cascade of classifiers to identify the face location.  We have made three additions to make it suitable for our real-time applications. We assume that the field of view of the camera contains a face, and the location of the face does not change abruptly.
Based on these assumptions we use a tracking framework using Kalman Filter (KF) to constrain the area of search. The three modifications are discussed below.

\subsection{Speeding up operation with downsampling and ROI remapping }
In the original algorithm, the input image at full resolution is used for face detection. Integral images \cite{viola2001rapid} are computed from the full resolution images. Once the integral images are computed, any one of these Haar-like features can be computed at any scale or location in constant time. The classifier then searches over multiple scales in a sliding window fashion. To speed up the detection, the size of the image as well as the size of the face to be searched can be limited. Here we assume that the face is close to the camera (maximum 1 meter from the camera), based on this constraint we downsample the image by a scale factor of four. The downsampling approach reduces the number of steps for face size as well as the spatial area for search which reduces the computational time considerably. Even though face detection is carried out in the downsampled image, the detected ROI is remapped to the original image which retains high-resolution images for further applications like face recognition or eye detection. 
It was empirically observed that the accuracy of face detection was not affected much by scale factors up to four (as long as the face was near to the camera) \cite{george2015framework}, \cite{dasgupta2016improved}.

\subsection{Tilted face detection using an affine tranformation}
\begin{figure}[h]
\begin{center}
   \includegraphics[width=1\linewidth]{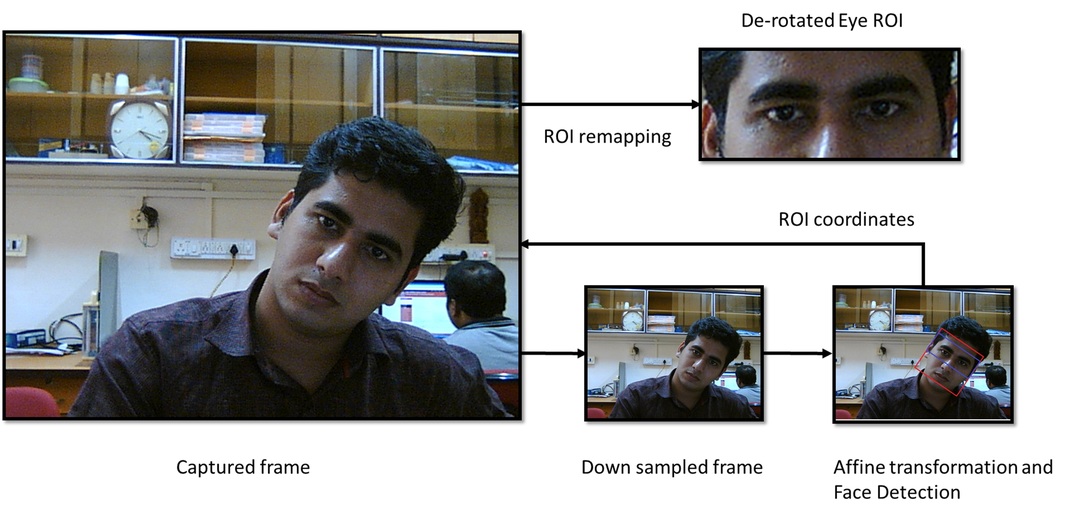}
\end{center}
\vspace{-1.5em}
   \caption{ Face detection schematic}
\label{fig:tilted_face}
\end{figure}

Direct application of Viola-Jones algorithm detects upright frontal images only.  Face detection along with subsequent stages fail if there is a moderate amount of tilt.  Most of the applications require the detection of faces in tilted conditions as well. An affine transformation based method is adopted for the detection of tilted (in-plane rotated) faces. The rotation matrix can be found for an $n$ dimensional image once its size, center, and angle of rotation needed are known. The affine transformation is added along with down sampling to make a robust face detection algorithm. The algorithm starts with applying the affine transformation based on the face detection result from the previous frame.  A detailed description of this algorithm is provided in our earlier works \cite{dasgupta2013board}, \cite{dasgupta2013vision}.  A schematic of the combined affine transformation along with the downsampling and ROI-remapping is shown in Fig. \ref{fig:tilted_face}.

\subsection{Face tracking using Kalman Filter}

Detection of the face using the modified Viola-Jones approach fails to detect a face in some frames.  Kalman Filter based tracking is used to avoid this issue. There are two advantages wit this tracking approach, 1) The predictions from Kalman filter can be used to constrain the search space for face detection, 2) The prediction from Kalman filter can be used as the tentative face location if the face detection stage fails.  We have used a uniform velocity model for face tracking. In every frame, the model is updated if the face detector returns the face location. In case the face detection fails, the predicted location of the Kalman filter is used as the tentative location of the face.

\subsection{Optical flow based face tracking}

\begin{figure}[h]
\begin{center}
   \includegraphics[width=1\linewidth]{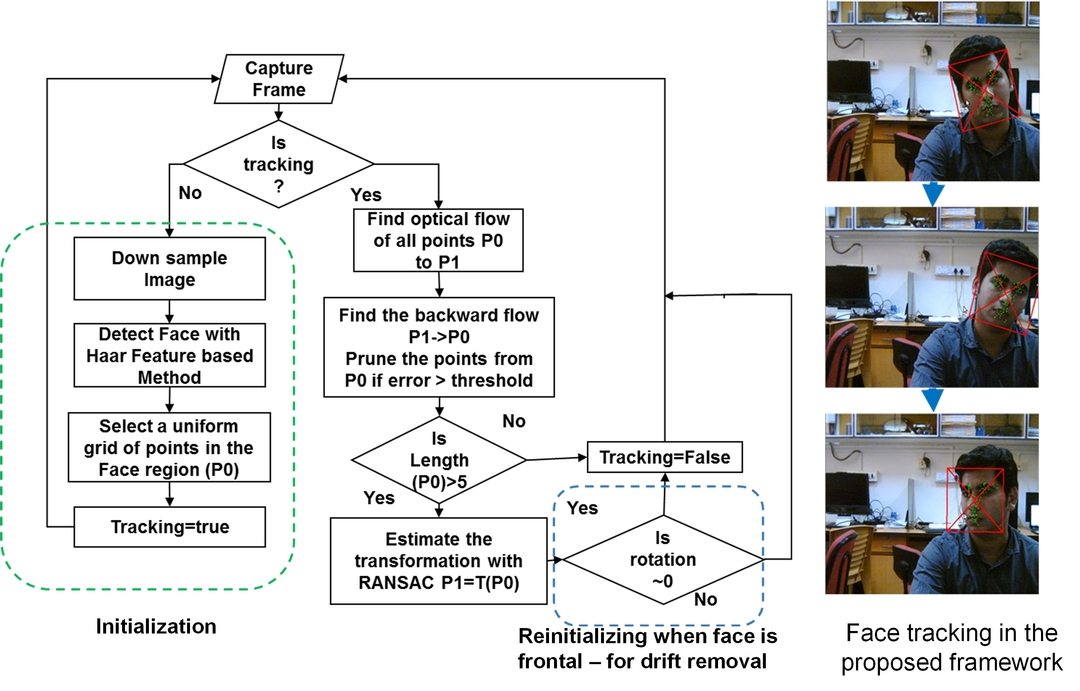}
\end{center}
\vspace{-1.5em}
   \caption{Lucas-Kanade based face tracking}
\label{fig:lktrack_v2}
\end{figure}
The Haar-like feature based method fails in detecting the face with off-plane rotations. To avoid this, we have used an optical flow based tracking scheme. Tracking stage is initialized using detected face region. A uniform grid of points are selected in the detected face region and are tracked in the subsequent frames using Lucas-Kanade optical flow. The successfully tracked points are found out using forward-backward error \cite{kalal2010forward}. Face position in the next frame is found out using the transformation between the point sets with RANSAC algorithm. Optical flow based tracking approach may drift in the long term. Optical flow tracking is reinitialized when the angle of the face become horizontal to avoid drift. The schematic of this approach is shown in Fig. \ref{fig:lktrack_v2}. This approach is suitable for video based applications where continuous face position is required.

\end{onehalfspacing}

%% file: HeadTail/publication_v2.tex
\chapter*{Publications from this Thesis} 
\vspace*{-0.15cm}

\subsection*{Journals}

\begin{itemize}
\setlength\itemsep{-0.25em}

\item [1.] \textbf{A. George}, A. Routray, ``Fast and Accurate Eye Localisation Algorithm for Gaze Tracking in Low Resolution Images'', \textit{IET Computer Vision}, vol. 10, no. 7, pp.660-669, 2016.

\item [2.] \textbf{A. George}, A. Routray, ``A score level fusion method for eye movement biometrics'',
 \textit{Pattern Recognition Letters}, vol. 82, no. 2, pp. 207-215, Elsevier, 2015.

\item [3.] A. Sengupta, A. Dasgupta, A. Chaudhuri, \textbf{A. George}, A. Routray, R. Guha,``A Multimodal System for Assessing Alertness Levels due to Cognitive Loading'', in \textit{IEEE in Transactions on Neural Systems \& Rehabilitation Engineering}, 2017.
\end{itemize} 

\subsection*{Book Chapters}
\begin{itemize}
\setlength\itemsep{-0.25em}
\item [1.] A. Sengupta, \textbf{A. George}, A. Dasgupta, A. Chaudhuri, B. Kabi, A. Routray , ``Alertness monitoring system for vehicle drivers using physiological signals''. \textit{Handbook of Research on Emerging Innovations in Rail Transportation Engineering}, pp. 273-311, IGI Global, 2016.

\end{itemize}

\subsection*{Conferences}
\begin{itemize}
\setlength\itemsep{-0.25em}
\item [1.] \textbf{A. George}, A. Routray, ``Real-time Eye Gaze Direction Classification Using Convolutional Neural Network'',  \textit{SPCOM, International Conference on Signal Processing and Communications, IEEE}, pp. 1-5, 2016.

\item [2.] A. Dasgupta, A. Mandloi, \textbf{A. George}, A. Routray, ``An Improved Algorithm for Eye Corner Detection'', \textit{SPCOM, International Conference on Signal Processing and Communications, IEEE }, pp. 1-4, 2016.

\item [3.] G. Banik, P. Patnaik, \textbf{A. George}, A. Routray, ``Contextual Priming and Perception Manipulation: An Exploration through Eye-Tracking and Audience Response'', \textit{ NAOP Convention 2016, Allahabad}.

\end{itemize}

%% file: HeadTail/biodata_v6.tex
\thispagestyle{empty}

\par{\centering{\Huge \bf Curriculum Vita}\bigskip\par} 
 \subsection*{Contact Information}
 \hrule
 
\begin{table}[H]
  
  \begin{tabular}{rl}
  
\textsc{Name:} & Anjith George\\
    
\textsc{Permanent Address:} & Thannikkapara House, Kolikkadavu,\\& Payam P.O, Kannur DT, Kerala\\ &PIN--670704, India.\\
\textsc{Email:} & anjith2006@gmail.com\\
\textsc{Mobile:} & +91-7501-549613 \\
  \end{tabular}
\end{table}

\subsection*{Research Interests} 

Human Computer Interaction, Gaze Tracking and applications, Biometrics, Computer Vision, Pattern Recognition and Machine Learning.

\subsection*{Education} 

\begin{table}[H]
\begin{tabular}{r|p{11cm}}
\textsc{2012-2017} &  \textbf{Ph.D}\\
& Image based eye gaze tracking and its applications\\
& Indian Institute of Technology Kharagpur, India\\
\multicolumn{2}{c}{} \\
\textsc{2010-2012} &  \textbf{M.Tech}\\
& Instrumentation Engineering\\
& Indian Institute of Technology Kharagpur, India\\
\multicolumn{2}{c}{} \\
\textsc{2006-2010} &  \textbf{B.Tech}\\
& Electrical and Electronics Engineering\\
& University of Calicut, Kerala, India\\
\end{tabular}
\end{table}

\subsection*{Awards \& Honours}

    \begin{itemize}  \itemsep -2pt 

                   \item \textbf{Winner}, BioEye 2015, International Eye Movement based biometrics competition, Organized by IEEE BTAS.
                  \item \textbf {Finalist}: Intel India Embedded Challenge 2012. 
                  \item \textbf{Secured 99.2 and 98.1 percentile} in Electrical engineering GATE 2010 and GATE 2012.
                  \item \textbf{MHRD Fellowship} during M Tech and PhD
                 \end{itemize}

\cleardoublepage                
\subsection*{Publications}
\small
\begin{itemize}
  \setlength\itemsep{-0.25em}

\item \textbf{A. George}, A. Routray, ``Fast and Accurate Eye Localisation Algorithm for Gaze Tracking in Low Resolution Images'', \textit{IET Computer Vision}, vol. 10, no. 7, pp.660-669, 2016.

\item \textbf{A. George}, A. Routray, ``A score level fusion method for eye movement biometrics'',
 \textit{Pattern Recognition Letters}, vol. 82, no. 2, pp. 207-215, Elsevier, 2015.
\item A. Dasgupta, \textbf{A. George}, S. L. Happy, A. Routray, ``A Vision Based System for Monitoring the Loss of
Attention in Automotive Drivers'', in \textit{IEEE Transactions on Intelligent Transportation Systems}, vol. 14, no.
4, pp.1825-1838, 2013
\item A. Dasgupta, \textbf{A. George}, S. L. Happy, A. Routray, Tara Shanker, ``An on-board vision based system for
drowsiness detection in automotive drivers'', in \textit{International Journal of Advances in Engineering Sciences
and Applied Mathematics}, Springer, vol. 5,
no. 2-3, pp. 94-103, 2013.
\item A. Sengupta, A. Dasgupta, A. Chaudhuri, \textbf{A. George}, A. Routray, R. Guha,``A Multimodal System for Assessing Alertness Levels due to Cognitive Loading'', in \textit{IEEE in Transactions on Neural Systems \& Rehabilitation Engineering}, 2017.
\item A. Sengupta, \textbf{A. George}, A. Dasgupta, A. Chaudhuri, B. Kabi, A. Routray , ``Alertness monitoring system for vehicle drivers using physiological signals''. \textit{Handbook of Research on Emerging Innovations in Rail Transportation Engineering}, pp. 273-311, IGI Global, 2016.
\item \textbf{A. George}, A. Routray, ``Real-time Eye Gaze Direction Classification Using Convolutional Neural Network'',  \textit{SPCOM, International Conference on Signal Processing and Communications, IEEE}, pp. 1-5, 2016.

\item A. Dasgupta, A. Mandloi, \textbf{A. George}, A. Routray, ``An Improved Algorithm for Eye Corner Detection'', \textit{SPCOM, International Conference on Signal Processing and Communications, IEEE }, pp. 1-4, 2016.

\item SL Happy, \textbf{A. George}, A. Routray, ``A real time facial expression classification system using Local Binary Patterns'',
 \textit{ International Conference on Intelligent Human Computer Interaction (IHCI)}, IEEE, 2012.

\item SL Happy, A. Dasgupta, \textbf{A. George}, A. Routray, ``A video database of human faces under near Infra-Red illumination for human computer interaction applications'',
 \textit{ International Conference on Intelligent Human Computer Interaction (IHCI)}, IEEE, 2012.

\end{itemize}     

\subsection*{Patents} \begin{itemize}
\setlength\itemsep{-.25em}

\item ``A SYSTEM FOR REAL-TIME ASSESSMENT OF ALERTNESS LEVEL OF HUMAN BEINGS'', A. Routray, A. Dasgupta, \textbf{A. George}, SL Happy. IN, 2013,
 \textit{ 634/KOL/2013}.

\end{itemize}     
\rule{410pt}{2pt}

%% file: main.bbl
\begin{thebibliography}{100}
\providecommand{\url}[1]{#1}
\csname url@samestyle\endcsname
\providecommand{\newblock}{\relax}
\providecommand{\bibinfo}[2]{#2}
\providecommand{\BIBentrySTDinterwordspacing}{\spaceskip=0pt\relax}
\providecommand{\BIBentryALTinterwordstretchfactor}{4}
\providecommand{\BIBentryALTinterwordspacing}{\spaceskip=\fontdimen2\font plus
\BIBentryALTinterwordstretchfactor\fontdimen3\font minus
  \fontdimen4\font\relax}
\providecommand{\BIBforeignlanguage}[2]{{%
\expandafter\ifx\csname l@#1\endcsname\relax
\typeout{** WARNING: IEEEtran.bst: No hyphenation pattern has been}%
\typeout{** loaded for the language `#1'. Using the pattern for}%
\typeout{** the default language instead.}%
\else
\language=\csname l@#1\endcsname
\fi
#2}}
\providecommand{\BIBdecl}{\relax}
\BIBdecl

\bibitem{ogaki2012coupling}
K.~Ogaki, K.~M. Kitani, Y.~Sugano, and Y.~Sato, ``Coupling eye-motion and
  ego-motion features for first-person activity recognition,'' in
  \emph{Conference on Computer Vision and Pattern Recognition Workshops}.\hskip
  1em plus 0.5em minus 0.4em\relax IEEE, 2012, pp. 1--7.

\bibitem{kitani2011fast}
K.~M. Kitani, T.~Okabe, Y.~Sato, and A.~Sugimoto, ``Fast unsupervised
  ego-action learning for first-person sports videos,'' in \emph{Computer
  Vision and Pattern Recognition, IEEE Conference on}, 2011, pp. 3241--3248.

\bibitem{oliva2001modeling}
A.~Oliva and A.~Torralba, ``Modeling the shape of the scene: A holistic
  representation of the spatial envelope,'' \emph{International journal of
  computer vision}, vol.~42, no.~3, pp. 145--175, 2001.

\bibitem{george2015design}
A.~George and A.~Routray, ``Design and implementation of real-time algorithms
  for eye tracking and perclos measurement for on board estimation of alertness
  of drivers,'' \emph{arXiv preprint arXiv:1505.06162}, 2015.

\bibitem{duchowski2007eye}
A.~Duchowski, \emph{Eye tracking methodology: Theory and practice}.\hskip 1em
  plus 0.5em minus 0.4em\relax Springer Science \& Business Media, 2007, vol.
  373.

\bibitem{happy2012video}
S.~Happy, A.~Dasgupta, A.~George, and A.~Routray, ``A video database of human
  faces under near infra-red illumination for human computer interaction
  applications,'' in \emph{2012 4th International Conference on Intelligent
  Human Computer Interaction (IHCI)}.\hskip 1em plus 0.5em minus 0.4em\relax
  IEEE, 2012, pp. 1--4.

\bibitem{young1975survey}
L.~R. Young and D.~Sheena, ``Survey of eye movement recording methods,''
  \emph{Behavior research methods \& instrumentation}, vol.~7, no.~5, pp.
  397--429, 1975.

\bibitem{westheimer1954mechanism}
G.~Westheimer, ``Mechanism of saccadic eye movements,'' \emph{AMA Archives of
  Ophthalmology}, vol.~52, no.~5, pp. 710--724, 1954.

\bibitem{morimoto2005eye}
C.~H. Morimoto and M.~R. Mimica, ``Eye gaze tracking techniques for interactive
  applications,'' \emph{Computer Vision and Image Understanding}, vol.~98,
  no.~1, pp. 4--24, 2005.

\bibitem{liu2002real}
X.~Liu, F.~Xu, and K.~Fujimura, ``Real-time eye detection and tracking for
  driver observation under various light conditions,'' in \emph{Intelligent
  Vehicle Symposium, IEEE}, vol.~2, 2002, pp. 344--351.

\bibitem{guang2008real}
Z.~Guang-yuan, C.~Bo, J.~Zhe, and L.~Jia-wen, ``A real-time eye detection
  system based on the active ir illumination,'' in \emph{Chinese Control and
  Decision Conference}.\hskip 1em plus 0.5em minus 0.4em\relax IEEE, 2008, pp.
  1255--1260.

\bibitem{ferhat2016low}
O.~Ferhat and F.~Vilari{\~n}o, ``Low cost eye tracking: The current panorama,''
  \emph{Computational intelligence and neuroscience}, 2016.

\bibitem{sengupta2017multimodal}
A.~Sengupta, A.~Dasgupta, A.~Chaudhuri, A.~George, A.~Routray, and R.~Guha, ``A
  multimodal system for assessing alertness levels due to cognitive loading,''
  \emph{IEEE Transactions on Neural Systems and Rehabilitation Engineering},
  vol.~25, no.~7, pp. 1037--1046, 2017.

\bibitem{sengupta2016alertness}
A.~Sengupta, A.~George, A.~Dasgupta, A.~Chaudhuri, B.~Kabi, and A.~Routray,
  ``Alertness monitoring system for vehicle drivers using physiological
  signals,'' in \emph{Handbook of Research on Emerging Innovations in Rail
  Transportation Engineering}.\hskip 1em plus 0.5em minus 0.4em\relax IGI
  Global, 2016, pp. 273--311.

\bibitem{dasgupta2015drowsiness}
A.~Dasgupta, B.~Kabi, A.~George, S.~Happy, and A.~Routray, ``A drowsiness
  detection scheme based on fusion of voice and vision cues,'' \emph{arXiv
  preprint arXiv:1509.04887}, 2015.

\bibitem{kasprowski2004eye}
P.~Kasprowski and J.~Ober, ``Eye movements in biometrics,'' in
  \emph{International Workshop on Biometric Authentication}.\hskip 1em plus
  0.5em minus 0.4em\relax Springer, 2004, pp. 248--258.

\bibitem{pai2016gazesim}
Y.~S. Pai, B.~Tag, B.~Outram, N.~Vontin, K.~Sugiura, and K.~Kunze, ``Gazesim:
  simulating foveated rendering using depth in eye gaze for vr,'' in \emph{ACM
  SIGGRAPH Posters}, 2016, p.~75.

\bibitem{duchowski2004gaze}
A.~T. Duchowski, N.~Cournia, and H.~Murphy, ``Gaze-contingent displays: A
  review,'' \emph{CyberPsychology \& Behavior}, vol.~7, no.~6, pp. 621--634,
  2004.

\bibitem{duchowski2002breadth}
A.~T. Duchowski, ``A breadth-first survey of eye-tracking applications,''
  \emph{Behavior Research Methods, Instruments, \& Computers}, vol.~34, no.~4,
  pp. 455--470, 2002.

\bibitem{jacob2003eye}
R.~Jacob and K.~S. Karn, ``Eye tracking in human-computer interaction and
  usability research: Ready to deliver the promises,'' \emph{Mind}, vol.~2,
  no.~3, p.~4, 2003.

\bibitem{zhai1999manual}
S.~Zhai, C.~Morimoto, and S.~Ihde, ``Manual and gaze input cascaded (magic)
  pointing,'' pp. 246--253, 1999.

\bibitem{patney2016perceptually}
A.~Patney, J.~Kim, M.~Salvi, A.~Kaplanyan, C.~Wyman, N.~Benty, A.~Lefohn, and
  D.~Luebke, ``Perceptually-based foveated virtual reality,'' in \emph{ACM
  SIGGRAPH Emerging Technologies}, 2016, p.~17.

\bibitem{majaranta2014eye}
P.~Majaranta and A.~Bulling, ``Eye tracking and eye-based human--computer
  interaction,'' in \emph{Advances in physiological computing}.\hskip 1em plus
  0.5em minus 0.4em\relax Springer, 2014, pp. 39--65.

\bibitem{majaranta2002twenty}
P.~Majaranta and K.-J. R{\"a}ih{\"a}, ``Twenty years of eye typing: systems and
  design issues,'' in \emph{Symposium on Eye tracking research \&
  applications}.\hskip 1em plus 0.5em minus 0.4em\relax ACM, 2002, pp. 15--22.

\bibitem{ueno2002dynamics}
A.~Ueno, T.~Tateyama, M.~Takase, and H.~Minamitani, ``Dynamics of saccadic eye
  movement depending on diurnal variation in human alertness,'' \emph{Systems
  and Computers in Japan}, vol.~33, no.~7, pp. 95--103, 2002.

\bibitem{diamantopoulos2009critical}
G.~Diamantopoulos, S.~I. Woolley, and M.~Spann, ``A critical review of past
  research into the neuro-linguistic programming eye-accessing cues model,''
  \emph{Current Research in NLP}, p.~8, 2009.

\bibitem{leigh2015neurology}
R.~J. Leigh and D.~S. Zee, \emph{The neurology of eye movements}.\hskip 1em
  plus 0.5em minus 0.4em\relax Oxford University Press, USA, 2015, vol.~90.

\bibitem{levy2010eye}
D.~L. Levy, A.~B. Sereno, D.~C. Gooding, and G.~A. O’Driscoll, ``Eye tracking
  dysfunction in schizophrenia: characterization and pathophysiology,'' in
  \emph{Behavioral Neurobiology of Schizophrenia and Its Treatment}.\hskip 1em
  plus 0.5em minus 0.4em\relax Springer, 2010, pp. 311--347.

\bibitem{majaranta2011gaze}
P.~Majaranta, \emph{Gaze Interaction and Applications of Eye Tracking: Advances
  in Assistive Technologies: Advances in Assistive Technologies}.\hskip 1em
  plus 0.5em minus 0.4em\relax IGI Global, 2011.

\bibitem{hansen2010eye}
D.~W. Hansen and Q.~Ji, ``In the eye of the beholder: A survey of models for
  eyes and gaze,'' \emph{Pattern Analysis and Machine Intelligence, IEEE
  Transactions on}, vol.~32, no.~3, pp. 478--500, 2010.

\bibitem{zhang2015appearance}
X.~Zhang, Y.~Sugano, M.~Fritz, and A.~Bulling, ``Appearance-based gaze
  estimation in the wild,'' in \emph{Proceedings of the IEEE Conference on
  Computer Vision and Pattern Recognition}, 2015, pp. 4511--4520.

\bibitem{illingworth1988survey}
J.~Illingworth and J.~Kittler, ``A survey of the hough transform,''
  \emph{Computer vision, graphics, and image processing}, vol.~44, no.~1, pp.
  87--116, 1988.

\bibitem{young1995specialised}
D.~Young, H.~Tunley, and R.~Samuels, \emph{Specialised hough transform and
  active contour methods for real-time eye tracking}.\hskip 1em plus 0.5em
  minus 0.4em\relax University of Sussex, Cognitive \& Computing Science, 1995.

\bibitem{smereka2008circular}
M.~Smereka and I.~Duleba, ``Circular object detection using a modified hough
  transform,'' \emph{International Journal of Applied Mathematics and Computer
  Science}, vol.~18, no.~1, pp. 85--91, 2008.

\bibitem{atherton1999size}
T.~J. Atherton and D.~J. Kerbyson, ``Size invariant circle detection,''
  \emph{Image and Vision computing}, vol.~17, no.~11, pp. 795--803, 1999.

\bibitem{yang2004novel}
P.~Yang, B.~Du, S.~Shan, and W.~Gao, ``A novel pupil localization method based
  on gaboreye model and radial symmetry operator,'' in \emph{Image Processing,
  International Conference on}, vol.~1.\hskip 1em plus 0.5em minus 0.4em\relax
  IEEE, 2004, pp. 67--70.

\bibitem{valenti2012accurate}
R.~Valenti and T.~Gevers, ``Accurate eye center location through invariant
  isocentric patterns,'' \emph{Pattern Analysis and Machine Intelligence, IEEE
  Transactions on}, vol.~34, no.~9, pp. 1785--1798, 2012.

\bibitem{valenti2012combining}
R.~Valenti, N.~Sebe, and T.~Gevers, ``Combining head pose and eye location
  information for gaze estimation,'' \emph{Image Processing, IEEE Transactions
  on}, vol.~21, no.~2, pp. 802--815, 2012.

\bibitem{timm2011accurate}
F.~Timm and E.~Barth, ``Accurate eye centre localisation by means of
  gradients.'' \emph{VISAPP}, vol.~11, pp. 125--130, 2011.

\bibitem{d2002ball}
T.~D'Orazio, N.~Ancona, G.~Cicirelli, and M.~Nitti, ``A ball detection
  algorithm for real soccer image sequences,'' in \emph{Pattern Recognition,
  16th International Conference on}, vol.~1.\hskip 1em plus 0.5em minus
  0.4em\relax IEEE, 2002, pp. 210--213.

\bibitem{daugman2004iris}
J.~Daugman, ``How iris recognition works,'' \emph{Circuits and Systems for
  Video Technology, IEEE Transactions on}, vol.~14, no.~1, pp. 21--30, 2004.

\bibitem{baek2013eyeball}
S.-J. Baek, K.-A. Choi, C.~Ma, Y.-H. Kim, and S.-J. Ko, ``Eyeball model-based
  iris center localization for visible image-based eye-gaze tracking systems,''
  \emph{Consumer Electronics, IEEE Transactions on}, vol.~59, no.~2, pp.
  415--421, 2013.

\bibitem{sewell2010real}
W.~Sewell and O.~Komogortsev, ``Real-time eye gaze tracking with an unmodified
  commodity webcam employing a neural network,'' in \emph{CHI'10 Extended
  Abstracts on Human Factors in Computing Systems}.\hskip 1em plus 0.5em minus
  0.4em\relax ACM, 2010, pp. 3739--3744.

\bibitem{zhou2004projection}
Z.-H. Zhou and X.~Geng, ``Projection functions for eye detection,''
  \emph{Pattern recognition}, vol.~37, no.~5, pp. 1049--1056, 2004.

\bibitem{bhaskar2003blink}
T.~Bhaskar, F.~T. Keat, S.~Ranganath, and Y.~Venkatesh, ``Blink detection and
  eye tracking for eye localization,'' in \emph{Conference on Convergent
  Technologies for the Asia-Pacific Region}, vol.~2.\hskip 1em plus 0.5em minus
  0.4em\relax IEEE, 2003, pp. 821--824.

\bibitem{wang2003eye}
J.~Wang, E.~Sung, and R.~Venkateswarlu, ``Eye gaze estimation from a single
  image of one eye,'' in \emph{Computer Vision, Ninth IEEE International
  Conference on}, 2003, pp. 136--143.

\bibitem{markuvs2014eye}
N.~Marku{\v{s}}, M.~Frljak, I.~S. Pand{\v{z}}i{\'c}, J.~Ahlberg, and
  R.~Forchheimer, ``Eye pupil localization with an ensemble of randomized
  trees,'' \emph{Pattern recognition}, vol.~47, no.~2, pp. 578--587, 2014.

\bibitem{schneider2014manifold}
T.~Schneider, B.~Schauerte, and R.~Stiefelhagen, ``Manifold alignment for
  person independent appearance-based gaze estimation,'' in \emph{22nd
  International Conference on Pattern Recognition}.\hskip 1em plus 0.5em minus
  0.4em\relax IEEE, 2014, pp. 1167--1172.

\bibitem{sugano2014learning}
Y.~Sugano, Y.~Matsushita, and Y.~Sato, ``Learning-by-synthesis for
  appearance-based 3d gaze estimation,'' in \emph{Proceedings of the IEEE
  Conference on Computer Vision and Pattern Recognition}, 2014, pp. 1821--1828.

\bibitem{viola2001rapid}
P.~Viola and M.~Jones, ``Rapid object detection using a boosted cascade of
  simple features,'' in \emph{Computer Vision and Pattern Recognition.
  Proceedings of the IEEE Computer Society Conference on}, vol.~1, 2001, pp.
  I--511.

\bibitem{dasgupta2013vision}
A.~Dasgupta, A.~George, S.~Happy, and A.~Routray, ``A vision-based system for
  monitoring the loss of attention in automotive drivers,'' \emph{IEEE
  Transactions on Intelligent Transportation Systems}, vol.~14, no.~4, pp.
  1825--1838, 2013.

\bibitem{li2005starburst}
D.~Li, D.~Winfield, and D.~J. Parkhurst, ``Starburst: A hybrid algorithm for
  video-based eye tracking combining feature-based and model-based
  approaches,'' in \emph{Computer Vision and Pattern Recognition-Workshops,
  IEEE Computer Society Conference on}, 2005, pp. 79--79.

\bibitem{fitzgibbon1999direct}
A.~Fitzgibbon, M.~Pilu, and R.~B. Fisher, ``Direct least square fitting of
  ellipses,'' \emph{Pattern Analysis and Machine Intelligence, IEEE
  Transactions on}, vol.~21, no.~5, pp. 476--480, 1999.

\bibitem{fischler1981random}
M.~A. Fischler and R.~C. Bolles, ``Random sample consensus: a paradigm for
  model fitting with applications to image analysis and automated
  cartography,'' \emph{Communications of the ACM}, vol.~24, no.~6, pp.
  381--395, 1981.

\bibitem{swirski2012robust}
L.~{\'S}wirski, A.~Bulling, and N.~Dodgson, ``Robust real-time pupil tracking
  in highly off-axis images,'' in \emph{Proceedings of the Symposium on Eye
  Tracking Research and Applications}.\hskip 1em plus 0.5em minus 0.4em\relax
  ACM, 2012, pp. 173--176.

\bibitem{yoon2008new}
Y.~Yoon, A.~Kosaka, and A.~C. Kak, ``A new kalman-filter-based framework for
  fast and accurate visual tracking of rigid objects,'' \emph{Robotics, IEEE
  Transactions on}, vol.~24, no.~5, pp. 1238--1251, 2008.

\bibitem{kiruluta1997predictive}
A.~Kiruluta, M.~Eizenman, and S.~Pasupathy, ``Predictive head movement tracking
  using a kalman filter,'' \emph{Systems, Man, and Cybernetics, Part B:
  Cybernetics, IEEE Transactions on}, vol.~27, no.~2, pp. 326--331, 1997.

\bibitem{dalal2005histograms}
N.~Dalal and B.~Triggs, ``Histograms of oriented gradients for human
  detection,'' in \emph{Computer Vision and Pattern Recognition, IEEE Computer
  Society Conference on}, vol.~1, 2005, pp. 886--893.

\bibitem{cristinacce2006facial}
D.~Cristinacce and T.~F. Cootes, ``Facial feature detection and tracking with
  automatic template selection,'' in \emph{Automatic Face and Gesture
  Recognition, 7th International Conference on}.\hskip 1em plus 0.5em minus
  0.4em\relax IEEE, 2006, pp. 429--434.

\bibitem{vukadinovic2005fully}
D.~Vukadinovic and M.~Pantic, ``Fully automatic facial feature point detection
  using gabor feature based boosted classifiers,'' in \emph{Systems, Man and
  Cybernetics, IEEE International Conference on}, vol.~2, 2005, pp. 1692--1698.

\bibitem{lewis1995fast}
J.~Lewis, ``Fast normalized cross-correlation,'' in \emph{Vision interface},
  vol.~10, no.~1, 1995, pp. 120--123.

\bibitem{tomasi1991detection}
C.~Tomasi and T.~Kanade, \emph{Detection and tracking of point features}.\hskip
  1em plus 0.5em minus 0.4em\relax School of Computer Science, Carnegie Mellon
  Univ. Pittsburgh, 1991.

\bibitem{pires2013visible}
B.~Pires, M.~Hwangbo, M.~Devyver, and T.~Kanade, ``Visible-spectrum gaze
  tracking for sports,'' in \emph{Proceedings of the IEEE Conference on
  Computer Vision and Pattern Recognition Workshops}, 2013, pp. 1005--1010.

\bibitem{sigut2011iris}
J.~Sigut and S.-A. Sidha, ``Iris center corneal reflection method for gaze
  tracking using visible light,'' \emph{Biomedical Engineering, IEEE
  Transactions on}, vol.~58, no.~2, pp. 411--419, 2011.

\bibitem{nadaraya1964estimating}
E.~A. Nadaraya, ``On estimating regression,'' \emph{Theory of Probability \&
  Its Applications}, vol.~9, no.~1, pp. 141--142, 1964.

\bibitem{kohn2001nonparametric}
R.~Kohn, M.~Smith, and D.~Chan, ``Nonparametric regression using linear
  combinations of basis functions,'' \emph{Statistics and Computing}, vol.~11,
  no.~4, pp. 313--322, 2001.

\bibitem{cootes2001active}
T.~F. Cootes, G.~J. Edwards, and C.~J. Taylor, ``Active appearance models,''
  \emph{IEEE Transactions on Pattern Analysis \& Machine Intelligence},
  vol.~23, no.~6, pp. 681--685, 2001.

\bibitem{cristinacce2006feature}
D.~Cristinacce and T.~F. Cootes, ``Feature detection and tracking with
  constrained local models.'' in \emph{Proceedings of the British Machine
  Vision Conference}, vol.~2, no.~5, 2006, p.~6.

\bibitem{viola2004robust}
P.~Viola and M.~J. Jones, ``Robust real-time face detection,''
  \emph{International journal of computer vision}, vol.~57, no.~2, pp.
  137--154, 2004.

\bibitem{jesorsky2001robust}
O.~Jesorsky, K.~J. Kirchberg, and R.~W. Frischholz, ``Robust face detection
  using the hausdorff distance,'' in \emph{Audio-and video-based biometric
  person authentication}.\hskip 1em plus 0.5em minus 0.4em\relax Springer,
  2001, pp. 90--95.

\bibitem{BioID}
``Bioid database,'' \url{https://www.bioid.com/About/BioID-Face-Database/},
  accessed: 2015-04-09.

\bibitem{ponz2012dataset}
V.~Ponz, A.~Villanueva, and R.~Cabeza, ``Dataset for the evaluation of eye
  detector for gaze estimation,'' in \emph{Proceedings of the ACM Conference on
  Ubiquitous Computing}, 2012, pp. 681--684.

\bibitem{burrus1991dft}
C.~Burrus and T.~W. Parks, \emph{DFT/FFT and Convolution Algorithms: theory and
  Implementation}.\hskip 1em plus 0.5em minus 0.4em\relax John Wiley \& Sons,
  Inc., 1991.

\bibitem{bradski2000opencv}
G.~Bradski \emph{et~al.}, ``The opencv library,'' \emph{Doctor Dobbs Journal},
  vol.~25, no.~11, pp. 120--126, 2000.

\bibitem{george2018escaf}
A.~George and A.~Routray, ``Escaf: Pupil centre localization algorithm with
  candidate filtering,'' \emph{arXiv preprint arXiv:1807.10520}, 2018.

\bibitem{starner2013project}
T.~Starner, ``Project glass: An extension of the self,'' \emph{Pervasive
  Computing, IEEE}, vol.~12, no.~2, pp. 14--16, 2013.

\bibitem{benko2015fovear}
H.~Benko, E.~Ofek, F.~Zheng, and A.~D. Wilson, ``Fovear: Combining an optically
  see-through near-eye display with projector-based spatial augmented
  reality,'' in \emph{Proceedings of the 28th Annual ACM Symposium on User
  Interface Software \& Technology}.\hskip 1em plus 0.5em minus 0.4em\relax
  ACM, 2015, pp. 129--135.

\bibitem{guenter2012foveated}
B.~Guenter, M.~Finch, S.~Drucker, D.~Tan, and J.~Snyder, ``Foveated 3d
  graphics,'' \emph{ACM Transactions on Graphics}, vol.~31, no.~6, p. 164,
  2012.

\bibitem{san2010evaluation}
J.~San~Agustin, H.~Skovsgaard, E.~Mollenbach, M.~Barret, M.~Tall, D.~W. Hansen,
  and J.~P. Hansen, ``Evaluation of a low-cost open-source gaze tracker,'' in
  \emph{Symposium on Eye-Tracking Research \& Applications}.\hskip 1em plus
  0.5em minus 0.4em\relax ACM, 2010, pp. 77--80.

\bibitem{kassner2014pupil}
M.~Kassner, W.~Patera, and A.~Bulling, ``Pupil: an open source platform for
  pervasive eye tracking and mobile gaze-based interaction,'' in
  \emph{Proceedings of the 2014 ACM International Joint Conference on Pervasive
  and Ubiquitous Computing: Adjunct Publication}.\hskip 1em plus 0.5em minus
  0.4em\relax ACM, 2014, pp. 1151--1160.

\bibitem{javadi2015set}
A.-H. Javadi, Z.~Hakimi, M.~Barati, V.~Walsh, and L.~Tcheang, ``Set: a pupil
  detection method using sinusoidal approximation,'' \emph{Frontiers in
  neuroengineering}, vol.~8, 2015.

\bibitem{fuhl2015excuse}
W.~Fuhl, T.~K{\"u}bler, K.~Sippel, W.~Rosenstiel, and E.~Kasneci, ``Excuse:
  Robust pupil detection in real-world scenarios,'' in \emph{Computer analysis
  of images and patterns}.\hskip 1em plus 0.5em minus 0.4em\relax Springer,
  2015, pp. 39--51.

\bibitem{fuhl2015else}
W.~Fuhl, T.~C. Santini, T.~Kuebler, and E.~Kasneci, ``Else: Ellipse selection
  for robust pupil detection in real-world environments,'' pp. 123--130, 2016.

\bibitem{suzuki1985topological}
S.~Suzuki \emph{et~al.}, ``Topological structural analysis of digitized binary
  images by border following,'' \emph{Computer Vision, Graphics, and Image
  Processing}, vol.~30, no.~1, pp. 32--46, 1985.

\bibitem{douglas1973algorithms}
D.~H. Douglas and T.~K. Peucker, ``Algorithms for the reduction of the number
  of points required to represent a digitized line or its caricature,''
  \emph{Cartographica: The International Journal for Geographic Information and
  Geovisualization}, vol.~10, no.~2, pp. 112--122, 1973.

\bibitem{Fitzgibbon95abuyer's}
A.~Fitzgibbon and R.~B. Fisher, ``A buyer's guide to conic fitting,'' in
  \emph{British Machine Vision Conference}, 1995, pp. 513--522.

\bibitem{matas2004robust}
J.~Matas, O.~Chum, M.~Urban, and T.~Pajdla, ``Robust wide-baseline stereo from
  maximally stable extremal regions,'' \emph{Image and vision computing},
  vol.~22, no.~10, pp. 761--767, 2004.

\bibitem{nister2008linear}
D.~Nist{\'e}r and H.~Stew{\'e}nius, ``Linear time maximally stable extremal
  regions,'' in \emph{European Conference on Computer Vision}.\hskip 1em plus
  0.5em minus 0.4em\relax Springer, 2008, pp. 183--196.

\bibitem{forssen2007shape}
P.-E. Forss{\'e}n and D.~G. Lowe, ``Shape descriptors for maximally stable
  extremal regions,'' in \emph{11th International Conference on Computer
  Vision}.\hskip 1em plus 0.5em minus 0.4em\relax IEEE, 2007, pp. 1--8.

\bibitem{tonsen2016labelled}
M.~Tonsen, X.~Zhang, Y.~Sugano, and A.~Bulling, ``Labelled pupils in the wild:
  a dataset for studying pupil detection in unconstrained environments,'' in
  \emph{Proceedings of the Ninth Biennial ACM Symposium on Eye Tracking
  Research \& Applications}.\hskip 1em plus 0.5em minus 0.4em\relax ACM, 2016,
  pp. 139--142.

\bibitem{fuhl2016pupil}
W.~Fuhl, M.~Tonsen, A.~Bulling, and E.~Kasneci, ``Pupil detection in the wild:
  An evaluation of the state of the art in mobile head-mounted eye tracking,''
  \emph{Machine Vision and Applications}, 2016.

\bibitem{di2012towards}
L.~L. Di~Stasi, R.~Renner, A.~Catena, J.~J. Ca{\~n}as, B.~M. Velichkovsky, and
  S.~Pannasch, ``Towards a driver fatigue test based on the saccadic main
  sequence: A partial validation by subjective report data,''
  \emph{Transportation research part C: emerging technologies}, vol.~21, no.~1,
  pp. 122--133, 2012.

\bibitem{terao2011initiation}
Y.~Terao, H.~Fukuda, A.~Yugeta, O.~Hikosaka, Y.~Nomura, M.~Segawa, R.~Hanajima,
  S.~Tsuji, and Y.~Ugawa, ``Initiation and inhibitory control of saccades with
  the progression of parkinson's disease--changes in three major drives
  converging on the superior colliculus,'' \emph{Neuropsychologia}, vol.~49,
  no.~7, pp. 1794--1806, 2011.

\bibitem{bandlerfrogs}
R.~Bandler and J.~Grinder, ``Frogs into princes: Neuro linguistic
  programming,'' 2012.

\bibitem{sturt2012neurolinguistic}
J.~Sturt, S.~Ali, W.~Robertson, D.~Metcalfe, A.~Grove, C.~Bourne, and
  C.~Bridle, ``Neurolinguistic programming: a systematic review of the effects
  on health outcomes,'' \emph{British Journal of General Practice}, vol.~62,
  no. 604, pp. e757--e764, 2012.

\bibitem{vranceanu2013computer}
R.~Vranceanu, L.~Florea, and C.~Florea, ``A computer vision approach for the
  eye accesing cue model used in neuro-linguistic programming,'' \emph{Sci.
  Bull. Univ. Politehnica Bucharest Ser. C}, vol.~75, no.~4, pp. 79--90, 2013.

\bibitem{vranceanu2011fast}
R.~Vr{\^a}nceanu, C.~Vertan, R.~Condorovici, L.~Florea, and C.~Florea, ``A fast
  method for detecting eye accessing cues used in neuro-linguistic
  programming,'' in \emph{Intelligent Computer Communication and Processing,
  IEEE International Conference on}, 2011, pp. 225--229.

\bibitem{vranceanu2013automatic}
R.~Vranceanu, C.~Florea, L.~Florea, and C.~Vertan, ``Automatic detection of
  gaze direction for nlp applications,'' in \emph{Signals, Circuits and
  Systems, International Symposium on}.\hskip 1em plus 0.5em minus 0.4em\relax
  IEEE, 2013, pp. 1--4.

\bibitem{radlak2014gaze}
K.~Radlak, M.~Kawulok, B.~Smolka, and N.~Radlak, ``Gaze direction estimation
  from static images,'' in \emph{Multimedia Signal Processing, 16th
  International Workshop on}.\hskip 1em plus 0.5em minus 0.4em\relax IEEE,
  2014, pp. 1--4.

\bibitem{song2013literature}
F.~Song, X.~Tan, S.~Chen, and Z.-H. Zhou, ``A literature survey on robust and
  efficient eye localization in real-life scenarios,'' \emph{Pattern
  Recognition}, vol.~46, no.~12, pp. 3157--3173, 2013.

\bibitem{vranceanu2015gaze}
R.~Vr{\^a}nceanu, C.~Florea, L.~Florea, and C.~Vertan, ``Gaze direction
  estimation by component separation for recognition of eye accessing cues,''
  \emph{Machine Vision and Applications}, vol.~26, no. 2-3, pp. 267--278, 2015.

\bibitem{asteriadis2009natural}
S.~Asteriadis, D.~Soufleros, K.~Karpouzis, and S.~Kollias, ``A natural head
  pose and eye gaze dataset,'' in \emph{Proceedings of the International
  Workshop on Affective-Aware Virtual Agents and Social Robots}.\hskip 1em plus
  0.5em minus 0.4em\relax ACM, 2009, p.~1.

\bibitem{kazemi2014one}
V.~Kazemi and J.~Sullivan, ``One millisecond face alignment with an ensemble of
  regression trees,'' in \emph{Computer Vision and Pattern Recognition, IEEE
  Conference on}, 2014, pp. 1867--1874.

\bibitem{krizhevsky2012imagenet}
A.~Krizhevsky, I.~Sutskever, and G.~E. Hinton, ``Imagenet classification with
  deep convolutional neural networks,'' in \emph{Advances in neural information
  processing systems}, 2012, pp. 1097--1105.

\bibitem{lecun1998gradient}
Y.~LeCun, L.~Bottou, Y.~Bengio, and P.~Haffner, ``Gradient-based learning
  applied to document recognition,'' \emph{Proceedings of the IEEE}, vol.~86,
  no.~11, pp. 2278--2324, 1998.

\bibitem{bottou2010large}
L.~Bottou, ``Large-scale machine learning with stochastic gradient descent,''
  in \emph{Proceedings of COMPSTAT}.\hskip 1em plus 0.5em minus 0.4em\relax
  Springer, 2010, pp. 177--186.

\bibitem{florea2013can}
L.~Florea, C.~Florea, R.~Vr{\^a}nceanu, and C.~Vertan, ``Can your eyes tell me
  how you think? a gaze directed estimation of the mental activity,'' in
  \emph{Proceedings of the British Machine Vision Conference}, 2013, pp. 60--1.

\bibitem{vranceanu2013nlp}
R.~Vr{\^a}nceanu, C.~Florea, L.~Florea, and C.~Vertan, ``Nlp eac recognition by
  component separation in the eye region,'' in \emph{Computer Analysis of
  Images and Patterns}.\hskip 1em plus 0.5em minus 0.4em\relax Springer, 2013,
  pp. 225--232.

\bibitem{valstar2010facial}
M.~Valstar, B.~Martinez, X.~Binefa, and M.~Pantic, ``Facial point detection
  using boosted regression and graph models,'' in \emph{Computer Vision and
  Pattern Recognition, IEEE Conference on}, 2010, pp. 2729--2736.

\bibitem{valenti2008accurate}
R.~Valenti and T.~Gevers, ``Accurate eye center location and tracking using
  isophote curvature,'' in \emph{Computer Vision and Pattern Recognition, IEEE
  Conference on}, 2008, pp. 1--8.

\bibitem{zhu2012face}
X.~Zhu and D.~Ramanan, ``Face detection, pose estimation, and landmark
  localization in the wild,'' in \emph{Computer Vision and Pattern Recognition,
  IEEE Conference on}, 2012, pp. 2879--2886.

\bibitem{jain2007handbook}
A.~K. Jain, P.~Flynn, and A.~A. Ross, \emph{Handbook of biometrics}.\hskip 1em
  plus 0.5em minus 0.4em\relax Springer Science \& Business Media, 2007.

\bibitem{jain2004introduction}
A.~K. Jain, A.~Ross, and S.~Prabhakar, ``An introduction to biometric
  recognition,'' \emph{Circuits and Systems for Video Technology, IEEE
  Transactions on}, vol.~14, no.~1, pp. 4--20, 2004.

\bibitem{wang2009behavioral}
L.~Wang, X.~Geng, L.~Wang, and X.~Geng, \emph{Behavioral Biometrics For Human
  Identification: Intelligent Applications}.\hskip 1em plus 0.5em minus
  0.4em\relax IGI Global, 2009.

\bibitem{marcel2007person}
S.~Marcel and J.~d.~R. Mill{\'a}n, ``Person authentication using brainwaves
  (eeg) and maximum a posteriori model adaptation,'' \emph{Pattern Analysis and
  Machine Intelligence, IEEE Transactions on}, vol.~29, no.~4, pp. 743--752,
  2007.

\bibitem{plataniotis2006ecg}
K.~N. Plataniotis, D.~Hatzinakos, and J.~K. Lee, ``Ecg biometric recognition
  without fiducial detection,'' in \emph{Biometric Consortium
  Conference}.\hskip 1em plus 0.5em minus 0.4em\relax IEEE, 2006, pp. 1--6.

\bibitem{ib2005independent}
{I.B. Group}, ``Independent testing of iris recognition technology,''
  \emph{Final Report, NBCHC030114/0002}, 2005.

\bibitem{roberts2007biometric}
C.~Roberts, ``Biometric attack vectors and defences,'' \emph{Computers \&
  Security}, vol.~26, no.~1, pp. 14--25, 2007.

\bibitem{schuckers2002issues}
S.~Schuckers, L.~Hornak, T.~Norman, R.~Derakhshani, and S.~Parthasaradhi,
  ``Issues for liveness detection in biometrics,'' in \emph{Proceedings of
  Biometric Consortium Conference. IEEE}, 2002.

\bibitem{leigh1999neurology}
R.~J. Leigh and D.~S. Zee, \emph{The neurology of eye movements}.\hskip 1em
  plus 0.5em minus 0.4em\relax Oxford university press New York, 1999, vol.~90.

\bibitem{kinnunen2010towards}
T.~Kinnunen, F.~Sedlak, and R.~Bednarik, ``Towards task-independent person
  authentication using eye movement signals,'' in \emph{Symposium on
  Eye-Tracking Research \& Applications}.\hskip 1em plus 0.5em minus
  0.4em\relax ACM, 2010, pp. 187--190.

\bibitem{bednarik2005eye}
R.~Bednarik, T.~Kinnunen, A.~Mihaila, and P.~Fr{\"a}nti, ``Eye-movements as a
  biometric,'' in \emph{Image analysis}.\hskip 1em plus 0.5em minus 0.4em\relax
  Springer, 2005, pp. 780--789.

\bibitem{komogortsev2010biometric}
O.~V. Komogortsev, S.~Jayarathna, C.~R. Aragon, and M.~Mahmoud, ``Biometric
  identification via an oculomotor plant mathematical model,'' in
  \emph{Symposium on Eye-Tracking Research \& Applications}.\hskip 1em plus
  0.5em minus 0.4em\relax ACM, 2010, pp. 57--60.

\bibitem{komogortsev2012biometric}
O.~V. Komogortsev, A.~Karpov, L.~R. Price, and C.~Aragon, ``Biometric
  authentication via oculomotor plant characteristics,'' in \emph{Biometrics,
  5th IAPR International Conference on}.\hskip 1em plus 0.5em minus 0.4em\relax
  IEEE, 2012, pp. 413--420.

\bibitem{holland2013complexb}
C.~D. Holland and O.~V. Komogortsev, ``Complex eye movement pattern biometrics:
  the effects of environment and stimulus,'' \emph{Information Forensics and
  Security, IEEE Transactions on}, vol.~8, no.~12, pp. 2115--2126, 2013.

\bibitem{rigas2012biometric}
I.~Rigas, G.~Economou, and S.~Fotopoulos, ``Biometric identification based on
  the eye movements and graph matching techniques,'' \emph{Pattern Recognition
  Letters}, vol.~33, no.~6, pp. 786--792, 2012.

\bibitem{rigas2012human}
I.~Rigas, G.~Economou, and S.~Fotopoulos, ``Human eye movements as a trait for
  biometrical identification,'' in \emph{Biometrics: Theory, Applications and
  Systems, Fifth International Conference on}.\hskip 1em plus 0.5em minus
  0.4em\relax IEEE, 2012, pp. 217--222.

\bibitem{zhang2012biometric}
Y.~Zhang and M.~Juhola, ``On biometric verification of a user by means of eye
  movement data mining,'' in \emph{The Second International Conference on
  Advances in Information Mining and Management}, 2012, pp. 85--90.

\bibitem{cantoni2015gant}
V.~Cantoni, C.~Galdi, M.~Nappi, M.~Porta, and D.~Riccio, ``Gant: Gaze analysis
  technique for human identification,'' \emph{Pattern Recognition}, vol.~48,
  no.~4, pp. 1027--1038, 2015.

\bibitem{holland2011biometric}
C.~Holland and O.~V. Komogortsev, ``Biometric identification via eye movement
  scanpaths in reading,'' in \emph{Biometrics, International Joint Conference
  on}.\hskip 1em plus 0.5em minus 0.4em\relax IEEE, 2011, pp. 1--8.

\bibitem{holland2013complexa}
C.~D. Holland and O.~V. Komogortsev, ``Complex eye movement pattern biometrics:
  Analyzing fixations and saccades,'' in \emph{Biometrics, International
  Conference on}.\hskip 1em plus 0.5em minus 0.4em\relax IEEE, 2013, pp. 1--8.

\bibitem{bioeye}
``Bioeye2015,competition on biometrics via eye movements,''
  \url{http://bioeye.cs.txstate.edu/}, accessed: 2015-04-09.

\bibitem{collewijn1988binocular}
H.~Collewijn, C.~J. Erkelens, and R.~Steinman, ``Binocular co-ordination of
  human horizontal saccadic eye movements.'' \emph{The Journal of Physiology},
  vol. 404, no.~1, pp. 157--182, 1988.

\bibitem{harris1984instrument}
C.~M. Harris, I.~Abramov, and L.~Hainl, ``Instrument considerations in
  measuring fast eye movements,'' \emph{Behavior Research Methods, Instruments,
  \& Computers}, vol.~16, no.~4, pp. 341--350, 1984.

\bibitem{krishnan2013selection}
S.~R. Krishnan and C.~S. Seelamantula, ``On the selection of optimum
  savitzky-golay filters,'' \emph{Signal Processing, IEEE Transactions on},
  vol.~61, no.~2, pp. 380--391, 2013.

\bibitem{savitzky1964smoothing}
A.~Savitzky and M.~J. Golay, ``Smoothing and differentiation of data by
  simplified least squares procedures.'' \emph{Analytical chemistry}, vol.~36,
  no.~8, pp. 1627--1639, 1964.

\bibitem{holland2012biometric}
C.~D. Holland and O.~V. Komogortsev, ``Biometric verification via complex eye
  movements: The effects of environment and stimulus,'' in \emph{Biometrics:
  Theory, Applications and Systems, Fifth International Conference on}.\hskip
  1em plus 0.5em minus 0.4em\relax IEEE, 2012, pp. 39--46.

\bibitem{salvucci2000identifying}
D.~D. Salvucci and J.~H. Goldberg, ``Identifying fixations and saccades in
  eye-tracking protocols,'' in \emph{Symposium on Eye tracking research \&
  applications}.\hskip 1em plus 0.5em minus 0.4em\relax ACM, 2000, pp. 71--78.

\bibitem{harwood2008optimally}
M.~R. Harwood and J.~P. Herman, ``Optimally straight and optimally curved
  saccades,'' \emph{The Journal of Neuroscience}, vol.~28, no.~30, pp.
  7455--7457, 2008.

\bibitem{kohavi1997wrappers}
R.~Kohavi and G.~H. John, ``Wrappers for feature subset selection,''
  \emph{Artificial intelligence}, vol.~97, no.~1, pp. 273--324, 1997.

\bibitem{goossens1997human}
H.~H. Goossens and A.~Van~Opstal, ``Human eye-head coordination in two
  dimensions under different sensorimotor conditions,'' \emph{Experimental
  Brain Research}, vol. 114, no.~3, pp. 542--560, 1997.

\bibitem{broomhead1988radial}
D.~S. Broomhead and D.~Lowe, ``Radial basis functions, multi-variable
  functional interpolation and adaptive networks,'' DTIC Document, Tech. Rep.,
  1988.

\bibitem{schwenker2001three}
F.~Schwenker, H.~A. Kestler, and G.~Palm, ``Three learning phases for
  radial-basis-function networks,'' \emph{Neural networks}, vol.~14, no.~4, pp.
  439--458, 2001.

\bibitem{maio2004fvc2004}
D.~Maio, D.~Maltoni, R.~Cappelli, J.~L. Wayman, and A.~K. Jain, ``Fvc2004:
  Third fingerprint verification competition,'' in \emph{Biometric
  Authentication}.\hskip 1em plus 0.5em minus 0.4em\relax Springer, 2004, pp.
  1--7.

\bibitem{phillips2010frvt}
P.~J. Phillips, W.~T. Scruggs, A.~J. O'Toole, P.~J. Flynn, K.~W. Bowyer, C.~L.
  Schott, and M.~Sharpe, ``Frvt 2006 and ice 2006 large-scale experimental
  results,'' \emph{Pattern Analysis and Machine Intelligence, IEEE Transactions
  on}, vol.~32, no.~5, pp. 831--846, 2010.

\bibitem{komogortsev2014template}
O.~V. Komogortsev, C.~D. Holland, and A.~Karpov, ``Template aging in eye
  movement-driven biometrics,'' in \emph{SPIE Defense+ Security}.\hskip 1em
  plus 0.5em minus 0.4em\relax International Society for Optics and Photonics,
  2014, pp. 90\,750A--90\,750A.

\bibitem{kasprowski2013impact}
P.~Kasprowski, ``The impact of temporal proximity between samples on eye
  movement biometric identification,'' in \emph{Computer Information Systems
  and Industrial Management}.\hskip 1em plus 0.5em minus 0.4em\relax Springer,
  2013, pp. 77--87.

\bibitem{george2018recognition}
A.~George and A.~Routray, ``Recognition of activities from eye gaze and
  egocentric video,'' \emph{arXiv preprint arXiv:1805.07253}, 2018.

\bibitem{poppe2010survey}
R.~Poppe, ``A survey on vision-based human action recognition,'' \emph{Image
  and vision computing}, vol.~28, no.~6, pp. 976--990, 2010.

\bibitem{weinland2011survey}
D.~Weinland, R.~Ronfard, and E.~Boyer, ``A survey of vision-based methods for
  action representation, segmentation and recognition,'' \emph{Computer Vision
  and Image Understanding}, vol. 115, no.~2, pp. 224--241, 2011.

\bibitem{turaga2008machine}
P.~Turaga, R.~Chellappa, V.~S. Subrahmanian, and O.~Udrea, ``Machine
  recognition of human activities: A survey,'' \emph{Circuits and Systems for
  Video Technology, IEEE Transactions on}, vol.~18, no.~11, pp. 1473--1488,
  2008.

\bibitem{fathi2011understanding}
A.~Fathi, A.~Farhadi, and J.~M. Rehg, ``Understanding egocentric activities,''
  in \emph{Computer Vision, IEEE International Conference on}, 2011, pp.
  407--414.

\bibitem{pirsiavash2012detecting}
H.~Pirsiavash and D.~Ramanan, ``Detecting activities of daily living in
  first-person camera views,'' in \emph{Computer Vision and Pattern
  Recognition, IEEE Conference on}, 2012, pp. 2847--2854.

\bibitem{yan2015egocentric}
Y.~Yan, E.~Ricci, G.~Liu, and N.~Sebe, ``Egocentric daily activity recognition
  via multitask clustering,'' \emph{Image Processing, IEEE Transactions on},
  vol.~24, no.~10, pp. 2984--2995, 2015.

\bibitem{nguyen2016recognition}
T.-H.-C. Nguyen, J.-C. Nebel, and F.~Florez-Revuelta, ``Recognition of
  activities of daily living with egocentric vision: A review.'' \emph{Sensors
  (Basel, Switzerland)}, vol.~16, no.~72, 2016.

\bibitem{bulling2011eye}
A.~Bulling, J.~A. Ward, H.~Gellersen, and G.~Troster, ``Eye movement analysis
  for activity recognition using electrooculography,'' \emph{IEEE transactions
  on pattern analysis and machine intelligence}, vol.~33, no.~4, pp. 741--753,
  2011.

\bibitem{hipiny2012recognising}
I.~M. Hipiny and W.~Mayol-Cuevas, ``Recognising egocentric activities from gaze
  regions with multiple-voting bag of words,'' University of Bristol, Tech.
  Rep., 2012.

\bibitem{li2015delving}
Y.~Li, Z.~Ye, and J.~M. Rehg, ``Delving into egocentric actions,'' in
  \emph{Proceedings of the IEEE Conference on Computer Vision and Pattern
  Recognition}, 2015, pp. 287--295.

\bibitem{fathi2012learning}
A.~Fathi, Y.~Li, and J.~M. Rehg, ``Learning to recognize daily actions using
  gaze,'' in \emph{European Conference on Computer Vision}.\hskip 1em plus
  0.5em minus 0.4em\relax Springer, 2012, pp. 314--327.

\bibitem{shiga2014daily}
Y.~Shiga, T.~Toyama, Y.~Utsumi, K.~Kise, and A.~Dengel, ``Daily activity
  recognition combining gaze motion and visual features,'' in \emph{Proceedings
  of the International Joint Conference on Pervasive and Ubiquitous Computing:
  Adjunct Publication}.\hskip 1em plus 0.5em minus 0.4em\relax ACM, 2014, pp.
  1103--1111.

\bibitem{kunze2013activity}
K.~Kunze, M.~Iwamura, K.~Kise, S.~Uchida, and S.~Omachi, ``Activity recognition
  for the mind: Toward a cognitive" quantified self",'' \emph{Computer},
  vol.~46, no.~10, pp. 105--108, 2013.

\bibitem{sharif2014cnn}
A.~Sharif~Razavian, H.~Azizpour, J.~Sullivan, and S.~Carlsson, ``Cnn features
  off-the-shelf: an astounding baseline for recognition,'' in \emph{Proceedings
  of the IEEE Conference on Computer Vision and Pattern Recognition Workshops},
  2014, pp. 806--813.

\bibitem{gong2014multi}
Y.~Gong, L.~Wang, R.~Guo, and S.~Lazebnik, ``Multi-scale orderless pooling of
  deep convolutional activation features,'' in \emph{European Conference on
  Computer Vision}.\hskip 1em plus 0.5em minus 0.4em\relax Springer, 2014, pp.
  392--407.

\bibitem{kalal2010forward}
Z.~Kalal, K.~Mikolajczyk, and J.~Matas, ``Forward-backward error: Automatic
  detection of tracking failures,'' in \emph{Pattern recognition, IEEE 20th
  international conference on}, 2010, pp. 2756--2759.

\bibitem{ross2005feature}
A.~A. Ross and R.~Govindarajan, ``Feature level fusion of hand and face
  biometrics,'' in \emph{Defense and Security}.\hskip 1em plus 0.5em minus
  0.4em\relax International Society for Optics and Photonics, 2005, pp.
  196--204.

\bibitem{breiman2001random}
L.~Breiman, ``Random forests,'' \emph{Machine learning}, vol.~45, no.~1, pp.
  5--32, 2001.

\bibitem{ma2005classification}
Y.~Ma, B.~Cukic, and H.~Singh, ``A classification approach to multi-biometric
  score fusion,'' in \emph{International Conference on Audio-and Video-Based
  Biometric Person Authentication}.\hskip 1em plus 0.5em minus 0.4em\relax
  Springer, 2005, pp. 484--493.

\bibitem{spriggs2009temporal}
E.~H. Spriggs, F.~De~La~Torre, and M.~Hebert, ``Temporal segmentation and
  activity classification from first-person sensing,'' in \emph{IEEE Computer
  Society Conference on Computer Vision and Pattern Recognition Workshops},
  2009, pp. 17--24.

\bibitem{jones2001rapid}
P.~Jones, P.~Viola, and M.~Jones, ``Rapid object detection using a boosted
  cascade of simple features,'' in \emph{University of Rochester. Charles
  Rich}, 2001.

\bibitem{lienhart2002extended}
R.~Lienhart and J.~Maydt, ``An extended set of haar-like features for rapid
  object detection,'' in \emph{Image Processing, International Conference on},
  vol.~1, 2002, pp. I--900.

\bibitem{george2015framework}
A.~George, A.~Dasgupta, and A.~Routray, ``A framework for fast face and eye
  detection,'' \emph{arXiv preprint arXiv:1505.03344}, 2015.

\bibitem{dasgupta2016improved}
A.~Dasgupta, A.~Mandloi, A.~George, and A.~Routray, ``An improved algorithm for
  eye corner detection,'' in \emph{2016 International Conference on Signal
  Processing and Communications (SPCOM)}.\hskip 1em plus 0.5em minus
  0.4em\relax IEEE, 2016, pp. 1--4.

\bibitem{dasgupta2013board}
A.~Dasgupta, A.~George, S.~Happy, A.~Routray, and T.~Shanker, ``An on-board
  vision based system for drowsiness detection in automotive drivers,''
  \emph{International Journal of Advances in Engineering Sciences and Applied
  Mathematics}, vol.~5, no. 2-3, pp. 94--103, 2013.

\end{thebibliography}
